%% file: main.tex
\documentclass{aifrontiers}
\usepackage{aifrontiers}
\usepackage{listings}
\usepackage{algorithm}
\usepackage{algpseudocode}
\usepackage{pifont}
\usepackage{xcolor}
\usepackage{graphicx}
\usepackage{lipsum}
\usepackage{array}
\usepackage{colortbl}
\usepackage{booktabs}
\usepackage{subcaption}
\usepackage{wrapfig}
\usepackage{makecell}
\usepackage{caption}
\usepackage{svg}
\usepackage{amssymb}
\usepackage{array}
\input{preamble}

\begin{document}
\vspace*{-30pt}

\title{\texorpdfstring{%
  \raisebox{-0.6ex}{\includegraphics[height=3.5ex]{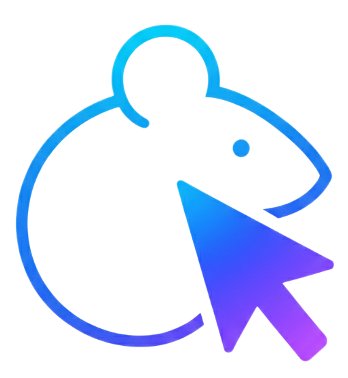}}%
  \hspace{0.5em}%
  Fara-1.5: Scalable Learning Environments for \\
  Computer Use Agents%
}{Fara-1.5: Scalable Synthetic Data Generation for Computer Use Agents}}
\shorttitle{Fara-1.5 CUA model}
\author{
    Ahmed Awadallah, Sahil Gupta, Yash Lara, Yadong Lu, Hussein Mozannar, Akshay Nambi, \\
    Zach Nussbaum, Yash Pandya, Aravind Rajeswaran, Corby Rosset, Alexey Taymanov, \\
    Luiz do Valle, Vibhav Vineet, Spencer Whitehead, Andrew Zhao
}
\blfootnote{}
\date{\today}

\renewcommand{\ghlink}{https://github.com/microsoft/fara}
\renewcommand{\foundrylink}{https://aka.ms/fara1.5-9B-foundry}
\renewcommand{\weblink}{https://aka.ms/fara1.5}
\renewcommand{\hflink}{https://aka.ms/fara1.5-hf}

\begin{abstract}
\input{sections/abstract}
\end{abstract}

\maketitle

\begin{figure}[H]
    \centering
    \includegraphics[width=0.95\linewidth]{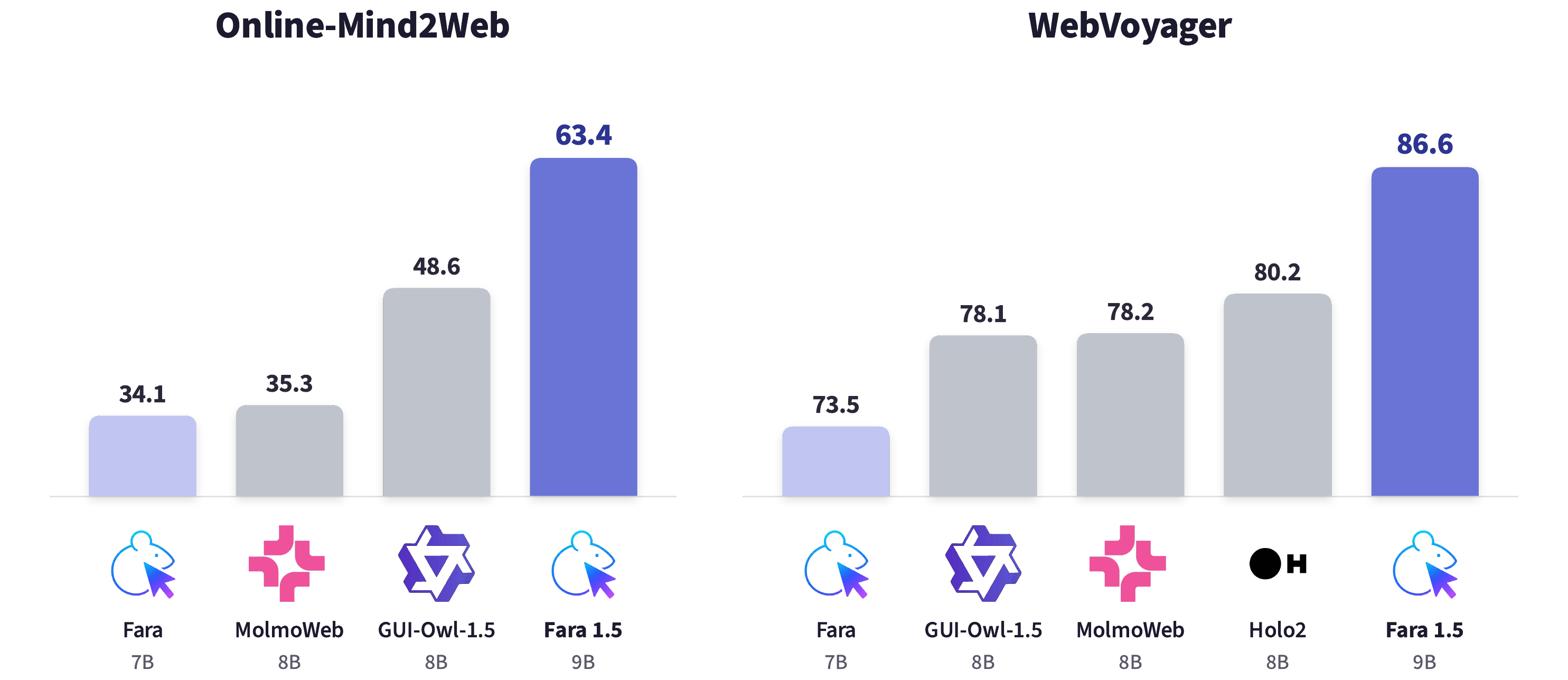}
    \caption{\small Task success rate on Online-Mind2Web and WebVoyager for similarly sized CUA models. \faranine\ reaches 63.4\% on Online-Mind2Web (a +29.3 point improvement over \model~\citep{fara7b2025} and a +14.8 point improvement over the prior best at this scale, GUI-Owl-1.5-8B) and 86.6\% on WebVoyager, setting a new state of the art for the 8--9B size class on both benchmarks.}
    \label{fig:hero-figure}
\end{figure}

\input{sections/introduction}

\input{sections/data_generation}
\input{sections/model}

\input{sections/experiments}
\input{sections/related_work}
\input{sections/discussion}
\input{sections/author_contributions}
\input{sections/acks}
\clearpage
\bibliographystyle{ACM-Reference-Format}
\bibliography{main}

\input{sections/appendix}

\end{document}

%% file: preamble.tex
\usepackage[utf8]{inputenc} %
\usepackage[T1]{fontenc}    %
\usepackage{url}            %
\usepackage{booktabs}       %
\usepackage{amsfonts}       %
\usepackage{nicefrac}       %
\usepackage{microtype}      %
\usepackage[table]{xcolor}  %
\usepackage{mathtools, nccmath}
\usepackage{minitoc}
\usepackage{xcolor}
\usepackage[inline, nomargin, draft]{fixme}
\fxsetup{theme=color}
\definecolor{fxnote}{rgb}{0.858, 0.188, 0.478}

\usepackage{microtype}
\usepackage{graphicx}
\usepackage{booktabs} %

\usepackage{comment}
\usepackage{pgfplots}
\pgfplotsset{compat=1.18}
\usepackage{url}
\usepackage{longtable}

\usepackage{amsfonts}       %
\usepackage{amsmath}
\usepackage{amsfonts}
\usepackage{amssymb}
\usepackage{comment}
\usepackage{pythonhighlight}

\usepackage{amsthm}
\usepackage{amscd}
\usepackage{t1enc}
\usepackage{indentfirst}
\usepackage{listings}
\usepackage{graphicx}
\usepackage{pict2e}
\usepackage{epic}
\usepackage{epstopdf} 
\usepackage{listings}
\usepackage{minitoc}

\usepackage{multirow}

\usepackage{amsmath}

\usepackage{empheq}

\renewcommand{\nu}{\vartheta}

\usepackage{float}
\usepackage{bm}
\usepackage{subcaption}
\usepackage{wrapfig}

\newtheorem{corollary }{Corollary }
\newtheorem*{theorem*}{Theorem}

\usepackage{apptools}
\AtAppendix{\counterwithin{lemma}{section}}
\AtAppendix{\counterwithin{theorem}{section}}
\AtAppendix{\counterwithin{corollary}{section}}

\usepackage{environ}
\usepackage{tikz}
\usetikzlibrary{arrows.meta,positioning}
\usepackage{float}

\usepackage{enumitem}

\usepackage{color-edits}
\addauthor{vc}{blue}

\usepackage{xspace}

\newcommand{\project}{FaraGen\xspace}
\newcommand{\model}{Fara-7B\xspace}
\newcommand{\fara}{Fara1.5\xspace}
\newcommand{\farafour}{Fara1.5-4B\xspace}
\newcommand{\faranine}{Fara1.5-9B\xspace}
\newcommand{\faratwentyseven}{Fara1.5-27B\xspace}
\newcommand{\faragen}{FaraGen1.5\xspace}

\newcommand{\qwennew}{Qwen3.5\xspace}

\newcommand{\fivefour}{GPT-5.4\xspace}

\newcommand{\farabench}{WebTailBench\xspace}

\newcommand{\tabref}[1]{Table\xspace~\ref{#1}}
\newcommand{\figref}[1]{Figure\xspace~\ref{#1}}
\newcommand{\secref}[1]{Section\xspace~\ref{#1}}

\newcommand{\eg}[1]{(\textit{e.g.,} #1)\xspace}
\newcommand{\ie}[1]{(\textit{i.e.,} #1)\xspace}

\newcommand{\myparagraph}[1]{\noindent\textbf{#1}}

\newlength\savewidth\newcommand\shline{\noalign{\global\savewidth\arrayrulewidth
  \global\arrayrulewidth 1pt}\hline\noalign{\global\arrayrulewidth\savewidth}}

\newcolumntype{x}[1]{>{\centering\arraybackslash}p{#1pt}}
\newcolumntype{y}[1]{>{\raggedright\arraybackslash}p{#1pt}}
\newcolumntype{z}[1]{>{\raggedleft\arraybackslash}p{#1pt}}

\definecolor{baselinecolor}{gray}{0.93}

\definecolor{arylideyellow}{rgb}{0.91, 0.84, 0.42}
\definecolor{babyblue}{rgb}{0.54, 0.81, 0.94}
\definecolor{babypink}{rgb}{0.96, 0.76, 0.76}
\definecolor{pastelgreen}{rgb}{0.47, 0.87, 0.47}
\definecolor{pastelpurple}{rgb}{0.86, 0.82, 1.0}
\definecolor{LightGray}{gray}{0.9}

\newcommand*{\rowstyle}[1]{%
  \gdef\@rowstyle{#1}%
  \@rowstyle\ignorespaces%
}
\newcolumntype{=}{%
  >{\gdef\@rowstyle{}}%
}
\newcolumntype{+}{%
  >{\@rowstyle}%
}
\definecolor{Gray}{gray}{0.9}
\newcolumntype{g}{>{\columncolor{Gray}}c}

\newcommand\blfootnote[1]{%
  \begingroup
  \renewcommand\thefootnote{}\footnote{#1}%
  \addtocounter{footnote}{-1}%
  \endgroup
}

\usepackage[most]{tcolorbox}
\usepackage{xcolor}

\definecolor{blue}{RGB}{0,120,140}
\definecolor{lightblue}{RGB}{232,245,250}

\newtcolorbox{findings}{
    enhanced,
    breakable,
    colback=lightblue,      %
    colframe=blue,          %
    boxrule=1.5pt,
    arc=0.25em,
    left=1em,
    right=1em,
    top=1em,
    bottom=0.75em,
    before=\vspace{1em},
    overlay unbroken and first={
        \node[
            fill=blue,
            text=white,
            font=\bfseries,
            anchor=west,
            inner xsep=0.75em,
            inner ysep=0.5em,
            rounded corners=0.25em
        ] 
        at ([xshift=0.75em]frame.north west) {Finding};
    }
}

\newtcolorbox{highlight}{
    enhanced,
    breakable,
    colback=lightblue,      %
    colframe=blue,          %
    boxrule=1.5pt,
    arc=0.25em,
    left=1em,
    right=1em,
    top=1em,
    bottom=0.75em,
    before=\vspace{1em},
    overlay unbroken and first={
        \node[
            fill=blue,
            text=white,
            font=\bfseries,
            anchor=west,
            inner xsep=0.75em,
            inner ysep=0.5em,
            rounded corners=0.25em
        ] 
        at ([xshift=0.75em]frame.north west) {Highlight};
    }
}

\newtcolorbox{stepbox}[1][]{
    colback=lightblue,      %
    colframe=blue,          %
    boxrule=1.5pt,
    sharp corners,
    enhanced,
    title=#1
}

\newtcolorbox{promptbox}[1]{
    enhanced, breakable,
    colback=lightblue, colframe=blue,
    boxrule=1pt, arc=0.25em,
    left=0.75em, right=0.75em, top=0.5em, bottom=0.5em,
    fonttitle=\bfseries\small,
    title=#1
}

%% file: sections/abstract.tex
Collecting computer use data from human demonstrations is expensive and slow, motivating the need for scalable generation strategies.
This requires two key ingredients: \emph{environments} in which agents can act and \emph{verifiers} that can judge whether their demonstrations succeeded.
We introduce \faragen, a scalable data pipeline for computer use agents composed of three modular components: environments, solvers, and verifiers.
\faragen\ uses both live websites and synthetic environments that faithfully simulate domains gated by authentication or that require irreversible actions. It employs a solver harness that can be powered by multiple models, including strong frontier models such as \fivefour, and also incorporates a user simulator to enable multi-turn rollouts. Finally, \faragen scores the resulting trajectories with three complementary verifiers covering task correctness, efficiency, and critical-point adherence.  
Using data produced by this pipeline, we train \fara, a family of native computer use agents (CUAs) at three scales built on \qwennew (4B, 9B, and 27B). To train these models, we employ a supervised finetuning (SFT) recipe that carefully balances data from \faragen for broad coverage, specific high-value tasks, and target model deficiencies in an iterative approach. Each model sets a new state of the art (SoTA) for its size class on browser-use benchmarks: \faranine\ reaches \textbf{63.4\%} on Online-Mind2Web and \textbf{86.6\%} on WebVoyager, while \faratwentyseven\ achieves \textbf{72.3\%} on Online-Mind2Web, which is competitive with much larger proprietary systems. We also release weights for the \fara models under MIT license, making SoTA computer use accessible for all beyond closed API-only systems.

%% file: sections/introduction.tex
\section{Introduction}
\label{sec:intro}

Large Language Models (LLMs) are increasingly being deployed as agents that act on a user's behalf, with computer use agents (CUAs) for web browsers emerging as an immediate and impactful instantiation~\citep{claudecomputeruse,fara7b2025,geminicomputeruse,openaioperator}.
Users provide tasks in natural language and the agent carries out the necessary steps like filling out forms on their computer.
Training such agents requires computer-use demonstrations of tasks on real world digital environments and verifiers to ensure the demonstrations accomplish the task. This type of data does not exist freely on the internet and human demonstrations are too expensive and slow to collect, motivating the use of automated recipes to generate such data \cite{fara7b2025}.

Live websites provide rich, real-world environments for CUAs to operate in, but they also carry real-world constraints and risks.
A wide range of user tasks are time-sensitive in nature, meaning they quickly become invalid with natural progression of time or website changes.
Another challenge is that many tasks require credentialed login for completion or task verification, which is dangerous for agents to attempt without oversight, especially for under-trained models without guardrails.
Finally, live websites may also have policies that limit the movement of agents or bots through them.
These practical constraints put limitations on the ability to scale training data from live websites.

Meanwhile, recent advancements in CUA research have produced agents that achieve non-trivial success rates on web tasks~\cite{anthropic2026opus47systemcard, gpt54}.
These models are often proprietary systems with complex scaffolds that allow an LLM to act as a web agent.
Nevertheless, they can serve as useful teacher models for web data collection, to be ultimately distilled into simpler \textit{``native'' CUA agents} that consume pixels and produce actions~\citep{fara7b2025}.
For these trajectories to act as useful training data, a high-quality \emph{verifier} is necessary to ensure quality control. This is a difficult task in itself as fully programmatic verification may not be feasible for websites with inaccessible backends, and LLMs as judges may hallucinate.

In this work, we approach these challenges in two ways.
First, building on our previous efforts for Fara-7B~\cite{fara7b2025}, we improve our data generation pipeline by redesigning our verifiers so they are more reliable as well as streamlining our task solving system and incorporating stronger teacher models.
Second, we generate \emph{synthetic environments}: fully functional and sandboxed websites, complete with frontend and backend infrastructure.
These synthetic environments allow us to create specific scenarios and tasks that go beyond the open internet, such as checking emails or managing experiments from a cloud provider.
To support these new kinds of tasks, we refine our approach for generating synthetic user interaction data, allowing us to create trajectories that involve realistic exchanges with simulated users.
Additionally, in these synthetic environments we have full control over the website and we are able to verify trajectories based the actual state of the website, rather than only screenshots or other artifacts.
These advancements make up the next iteration of our data generation pipeline, \faragen.

\myparagraph{Results.} We train \fara, a new family of CUA models for web browsing, using our improved \faragen pipeline. 
This family of models spans three sizes (\farafour, \faranine, and \faratwentyseven) and are the most capable web agents for their respective sizes while remaining practical to deploy on modest hardware (\figref{fig:hero-figure}).
Compared to our prior \model generation~\cite{fara7b2025}, we see clear improvements both quantitatively across all benchmarks and qualitatively through user experiences. Across key benchmarks, \fara outperforms other models of similar sizes. For example, on the \textbf{Online-Mind2Web benchmark} consisting of 300 tasks across 136 popular sites, \faranine achieves a task success rate of \textbf{63.4\%} which nearly doubles the performance of \model and significantly improves over the performance of GUI-Owl-1.5-8B (48.6\%), the prior best performing model at this scale. The larger \faratwentyseven model achieves \textbf{72.3\%} on the same benchmark, which is competitive or outperforms much larger models and proprietary systems like Gemini 2.5 Computer Use~\citep{geminicomputeruse} and Yutori n1~\citep{yutori2025navigator}. 

We also train \fara with a keen eye on improving user experience. This includes the choice of task distribution, synthetic domains we designed, the user-interaction pattern, adherence to user preferences and constraints. Unlike the predecessor \model, which was trained to never go past critical points -- irreversible actions like final steps in making payments or submitting forms -- \fara is trained to complete those steps after explicitly seeking out user approval. We complement our training data with synthetic domains that simulate popular online websites/apps to allow our model to act beyond gated domains and, e.g., send the email or book the flight rather than just searching for it. As a result, on internal user studies, we found that users experience smoother interactions and better control over their tasks.

\myparagraph{Open release.} The \fara\ family is released as open-weight models on HuggingFace, along with a model card and usage guidance. We hope this enables widespread adoption, reproducibility, and community research on computer use and web agents.

%% file: sections/data_generation.tex
\begin{figure}[t!]
    \centering
    \includegraphics[width=\linewidth]{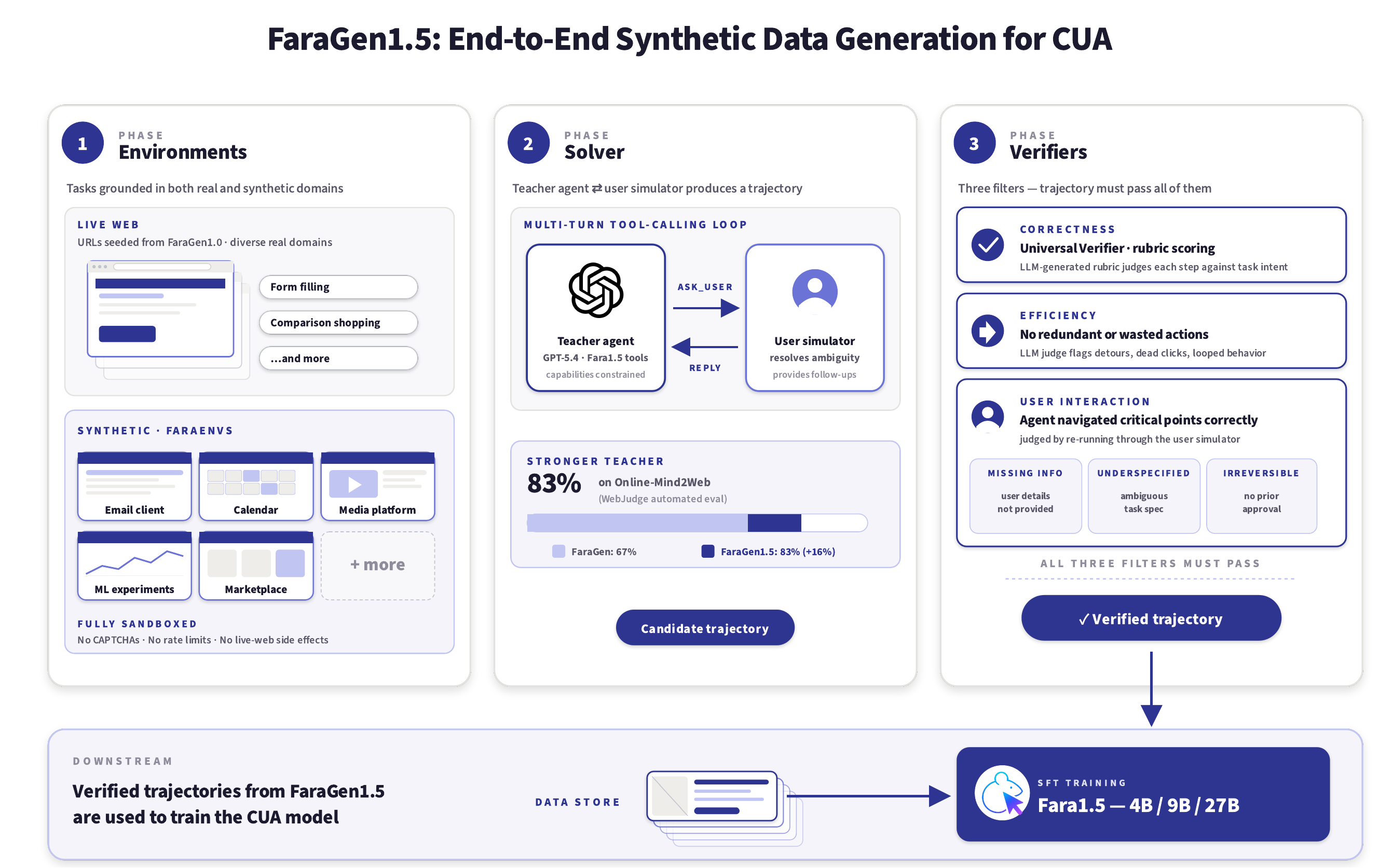}
    \caption{\small \faragen\ pipeline. Phase 1 instantiates tasks in two kinds of environments: \emph{online}, seeded by URLs from the \project~\citep{fara7b2025} index, and \emph{synthetic}, a set of six sandboxed replicas (\emph{FaraEnvs}) covering domains gated by authentication. Phase 2 has a \fivefour-based solver attempt each task in cooperation with a user simulator; the new solver reaches 83\% on Online-Mind2Web under the WebJudge of~\citet{xue2025om2w}, a +16-point absolute gain over the multi-agent \project\ solver. Phase 3 admits a trajectory into training only if it passes all three independent verifiers: correctness, efficiency, and critical-point adherence.}
    \label{fig:faragen_pipeline}
\end{figure}

\section{\faragen\ -- A Data Engine for Computer Use Agents}
\label{sec:data_generation}
Training performant CUAs is bottlenecked by the lack of high-quality interaction data. Human demonstration data is slow and expensive to collect and there is no naturally occurring dataset capturing realistic multi-step tasks. Additionally, existing synthetic pipelines are limited to what can be accomplished on the open internet, without authentication or irreversible side effects. \project~\citep{fara7b2025} generates verified trajectories from live websites at roughly \$1 per task, but the open web limits the activities a downstream agent can learn.
The agent never observes what successful login flows look like, how purchases are confirmed, or how messages get sent.
\faragen is the next evolution of this pipeline designed to address these challenges.
As shown in \figref{fig:faragen_pipeline}, we retain the three-stage organization of \project: environments that instantiate tasks, a solver that produces trajectories to solve tasks, and verifiers that filter the trajectories.
However, we rebuild each stage to address specific challenges: the open-web restriction on environments, the plateau in solver quality, and the limited reliability of single-perspective verification.

\subsection{Environments}
\subsubsection{Live Web Environments}
\label{subsec:online_environments}
Live web environments cover tasks that can be completed on the open internet safely and without authentication.
We source a wide range of websites to use as seeds for task generation from a large index of publicly accessible URLs~\cite{overwijk2022clueweb2210billionweb}.
To generate tasks for these sites, we first create summaries of each seed website's content, features, and structure through an initial exploration of the site which is then cached and re-used.
We provide this summary as part of a prompt to an LLM, which uses the summary to generate plausible tasks for the website.

We propose tasks by sampling along several dimensions of variation: target website, task complexity, phrasing style, user interaction type and task feasibility.
Given this sampling, we feed the values and the website summary to a task proposal LLM.
This proposal LLM is instructed to select a user intent \eg{form completion} and persona \eg{busy professional, student} that is plausible for the target site from a pre-defined set, and generate tasks based on the combination of the intent, persona, sampled dimensions, and website summary.
This hybrid scheme balances controlled coverage along the policy-relevant dimensions with site-specific plausibility along the user-side ones.
Since these dimensions are categorical, we use a coverage tracker to monitor the empirical distribution along every dimension. In Table \ref{tab:task_proposal_dimensions} we show the different sampling dimensions for tasks, and Table~\ref{tab:task_proposal_examples} shows a sample of random tasks from this pipeline. 

We put the proposed tasks through a two-stage filter.
A deterministic check rejects surface-level failures \eg{LLM refused, uses bare URLs}, and an LLM judge then rates each remaining task as good or bad on whether it describes something real with a clear goal.
Intentionally ambiguous or under-specified tasks are kept if they are judged as realistic by the LLM, since this matches the user-instruction distribution we want our solver to handle.

This process yields a large, rich set of tasks for a broad set of URLs and types of tasks.
However, as previously mentioned, the distribution of tasks covers those that can be completed on the open internet.
We build synthetic environments to target tasks beyond these.

\begin{figure}
    \centering
    \includegraphics[width=1\linewidth]{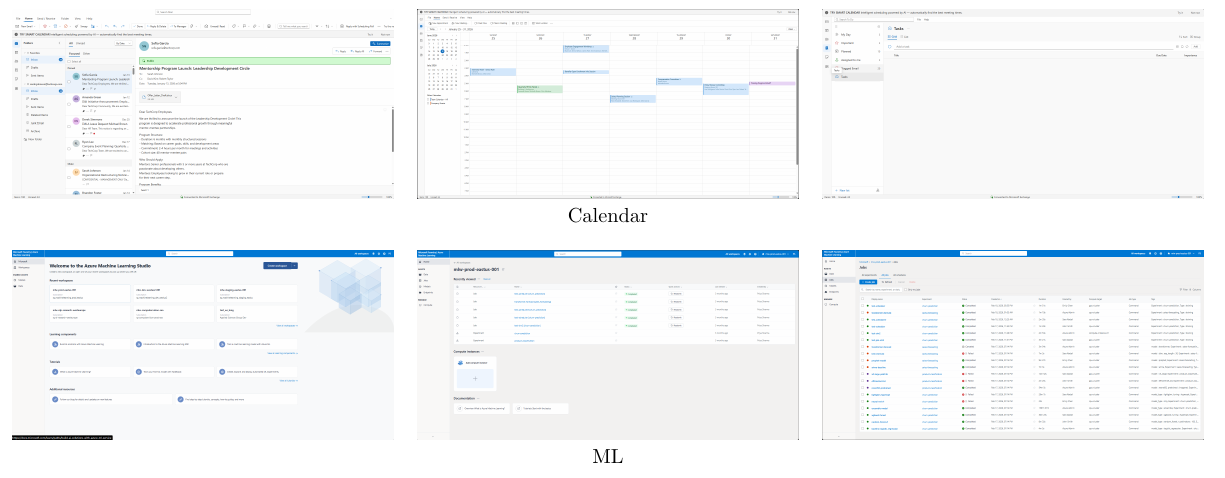}
    \caption{Sample FaraEnvs for Calendar and ML Management environments populated with realistic data generated by our pipeline, illustrating rich, production-like workflows and interactions across both environments.}
    \label{fig:synth-env}
\end{figure}

\subsubsection{Synthetic Environments}
\label{subsec:synthetic_environments}

Synthetic environments cover tasks that online environments cannot: tasks behind authentication, tasks that mutate persistent state, and tasks where the only success criterion is that an irreversible action was taken correctly.
The defining property of a synthetic environment is that we control the full stack - frontend, API, database, and seed data - so every task has a verifiable ground-truth success criterion expressible as a state predicate over the backend. This yields two key properties. First, tasks with irreversible side effects can be attempted safely: the websites are sandboxed and the database is reset between trajectories. Second, we can use the internal website state for execution-based verification of trajectories, rather than relying solely on screenshots or other surface metadata.

Hand-authoring high-fidelity replicas of complex web apps does not scale to many domains, so we instead use a semi-automated recipe driven by a coding agent (GitHub Copilot~\cite{github_copilot_docs}).
We first record human interaction trajectories on the target domain.
These trajectories are given to the coding agent, which generate a specification for a sandboxed clone consisting of a React frontend, a FastAPI backend, a SQLite schema, and a seed-data script to populate the database from a persona narrative.
The coding agent iteratively refines the clone through several rounds of human review: the first iteration is typically incomplete (non-functional buttons, missing edge cases, inconsistent state transitions), but each round of feedback continually improves the website.
After a small number of iterations, typically 3-5, we obtain a replica whose surface behavior matches the recorded human trajectories. This human-agent loop accelerates the development process while ensuring that the resulting environments are functionally correct.

Once an environment has been created, we generate tasks via a task proposal LLM that has direct access to the hosted synthetic environment through Model Context Protocol (MCP) tool servers: a SQLite MCP server that introspects the schema and queries entities and a Playwright MCP server that browses the live UI.
The LLM is also given a small bank of human-written exemplar tasks per category to anchor style as well as a user persona, similar to our live web environments.
For example, in our email environment, the database is populated as the inbox of an employee at a small IT firm, with internally consistent calendar invites, project threads, and recurring collaborators.
Tasks are proposed in two separate phases for the training split and a held-out test split, which are deduplicated against one another so no tasks/intents are overlapping.
Since the backend state is fully observable, every task carries a precomputed success criterion.
For tasks that mutate state, we record a database snapshot before and after the trajectory and use \texttt{sqldiff} to compute the row-level difference, which an LLM judge confirms is the intended mutation based on the task.
For read-only tasks, we precompute a reference answer (or a SQL query that produces it) at generation time and score the trajectory's final response against it.

Environment diversity matters more than data volume.
Recent work ~\cite{cuagym,gymanything} that expand to more trajectories within the same environments cannot substitute for it, and shallow environments add little value regardless of how many trajectories they yield.
Using our pipeline, we produce six synthetic environments (\figref{fig:synth-env}) that we collectively call \emph{FaraEnvs} -- email, calendar, media streaming, ML experiment management, marketplaces, and scheduling -- that prioritize \textbf{depth}: environments populated with coherent, realistic data; environments whose tasks require understanding the application's actual data model and UI patterns; and environments that feel like real user workflows rather than simple tasks.

\subsection{Solver}
\label{subsec:solver}

The solver attempts each proposed task and returns a trajectory that, if accepted by the verifiers, becomes a training example.
In the original \project, the solver is an orchestrator-worker multi-agent system built on Magentic-One and Magentic-UI~\citep{fourney2024magenticonegeneralistmultiagentsolving,mozannar2025magenticuihumanintheloopagenticsystems}.
Here we collapse it to a single agent built on \fivefour that runs a multi-turn tool-calling loop over a tool set mirroring the action space the student will eventually emit.
Our motivation for this is twofold: 1) A single-policy solver better matches the inference-time setting of the trained \fara\ models, reducing distribution shift between teacher and student; 2) it lets us inherit improvements in frontier model capability without re-engineering the orchestration layer for each generation of model.
On the same Online-Mind2Web tasks~\cite{xue2025om2w}, replacing the multi-agent system with a \fivefour-based solver raises end-to-end success from 67\% to 83\%, a +16-point absolute gain that translates directly to a higher yield of accepted training trajectories.
To keep this data \emph{learnable} by the smaller \fara\ students, we explicitly disable solver capabilities the student cannot replicate.
In particular, within our harness, we forbid the solver agent from issuing complex URL queries that would bypass site interaction entirely, since these queries are harder for smaller models to replicate and the resulting trajectories skip the UI interaction behavior we want the student to learn. The solver is also not allowed to do any dangerous actions or irreversible actions on the web such as creating accounts or posting on forums.

\myparagraph{User simulator.}
As the solver is attempting a task, a user simulator is invoked in two distinct circumstances.
First, when there is missing information for the task or the task is ambiguous, the solver agent issues an \texttt{ask\_user} and the simulator responds with the necessary information to continue.
Second, once the solver believes a task is complete, the simulator can either accept the result or extend it with a follow-up request, turning a single-task demonstration into a multi-turn dialogue; this second mode is the principal source of multi-turn training data for \fara. The exact prompts used at both callsites are given in Appendix~\ref{sec:user_simulator_prompts}.

Because the solver runs against real, live third-party websites, an unrestricted user simulator could push it into actions with real-world consequences --- creating accounts, completing checkouts, sending messages, or submitting real personal information --- all of which mutate third-party state and create legal exposure.
We therefore guard the \texttt{ask\_user} callsite with a lightweight LLM gate that fires before the simulator drafts its reply.
The gate inspects the agent's pending question against a list of forbidden side-effect categories (account registration, real login, real checkout, real messaging, public posting, subscription changes, real-data mutation, identity-tier PII, and final booking steps) and only lets the simulator answer when the question is in the harmless filtering, sorting, or preference-selection regime.
Otherwise the trajectory is halted and dropped.
Gate decisions, including the reasoning trace, are persisted alongside the trajectory so we can audit and tune the boundary over time.

\input{Tables/critical_point_types_classification}

\subsection{Verifiers}
\label{subsec:verifiers}

A high-quality solver still produces a long tail of low-quality trajectories: tasks completed in the wrong way, tasks completed inefficiently, or tasks where the agent should have deferred to the user but did not.
We use three independent verifiers, each targeting a distinct failure mode, and admit a trajectory into training only if it passes all three.
The verifiers are complementary: correctness alone cannot distinguish efficient from meandering successes, efficiency alone cannot catch trajectories that hallucinate a result, and neither catches trajectories that fabricate user information they should have asked for.

\myparagraph{Task correctness.}
We instantiate task correctness differently depending on whether ground truth is available. On live web environments, we use the process-reward score from the Universal Verifier~\citep{rosset2026art}, an ensemble LLM judge that generates a task-specific rubric and scores the trajectory step-by-step against it.
We accept trajectories whose rubric process score is at least 0.8.
On synthetic environments, we use two different criteria depending on the tasks: 1) state-mutating tasks are scored by computing the row-level \texttt{sqldiff} between the seed database and the post-trajectory database and asking an LLM judge whether the observed diff matches the task's intended mutation (and nothing else); 2) read-only tasks are scored by comparing the trajectory's final answer to a precomputed reference.
The judge for state-mutating tasks has lower label noise than the Universal Verifier, which is one reason we treat synthetic environments as a strict generalization of online ones; when a task \emph{can} be expressed synthetically, we prefer it.

\myparagraph{Task efficiency.}
A correct trajectory is not automatically a good training example.
Agents frequently complete tasks that contain redundant clicks, unnecessary navigations, or repeated state-checks.
Imitating these behaviors yields a student that is correct but slow, with all the resulting cost and latency consequences for inference.
The task efficiency verifier is a dedicated LLM judge that explicitly identifies looped action sequences as well as individual unnecessary actions, and assigns each trajectory an integer efficiency rating on a 1-5 scale.
We accept trajectories whose rating is at least 4.

\myparagraph{Critical-point adherence.}
The third verifier targets behavior at \emph{critical points}: steps at which the agent should pause and consult the user before acting. We first classify each task along three dimensions:

\begin{enumerate}
    \item \emph{Permission to perform irreversible action}: Does the task require an irreversible action and did the user explicitly provide authorization to perform it? Examples include finalizing a purchase or submitting a contact form.
    \item \emph{Task fully specified}: Was all the task-specific information needed to complete the task provided? For example, booking a flight would require the departure and return dates.
    \item \emph{Required user information provided}: Did the user provide all required Personally Identifiable Information (PII) needed to complete the task? For example, signing up for a newsletter would require the user's email address.
\end{enumerate}
These three dimensions combine to produce 8 distinct types of critical point classifications (Table \ref{tab:critical_point_types}).
The classification is done by an LLM judge and is based on the initial task prompt and the screenshots of the pages that the agent interacted with (to identify required fields, for instance).
During verification, the classification is injected into the rubric that scores the trajectory, and a dedicated compliance check determines whether the solver actually exhibited the expected behavior at each critical point.
Trajectories that cross a critical point without prior approval are rejected.
This verifier enables our model to know when to appropriately defer to the user.

\subsection{Data Distribution}

\begin{figure}[t!]
    \centering
    \hspace*{-20pt} \includegraphics[width=1.1\linewidth]{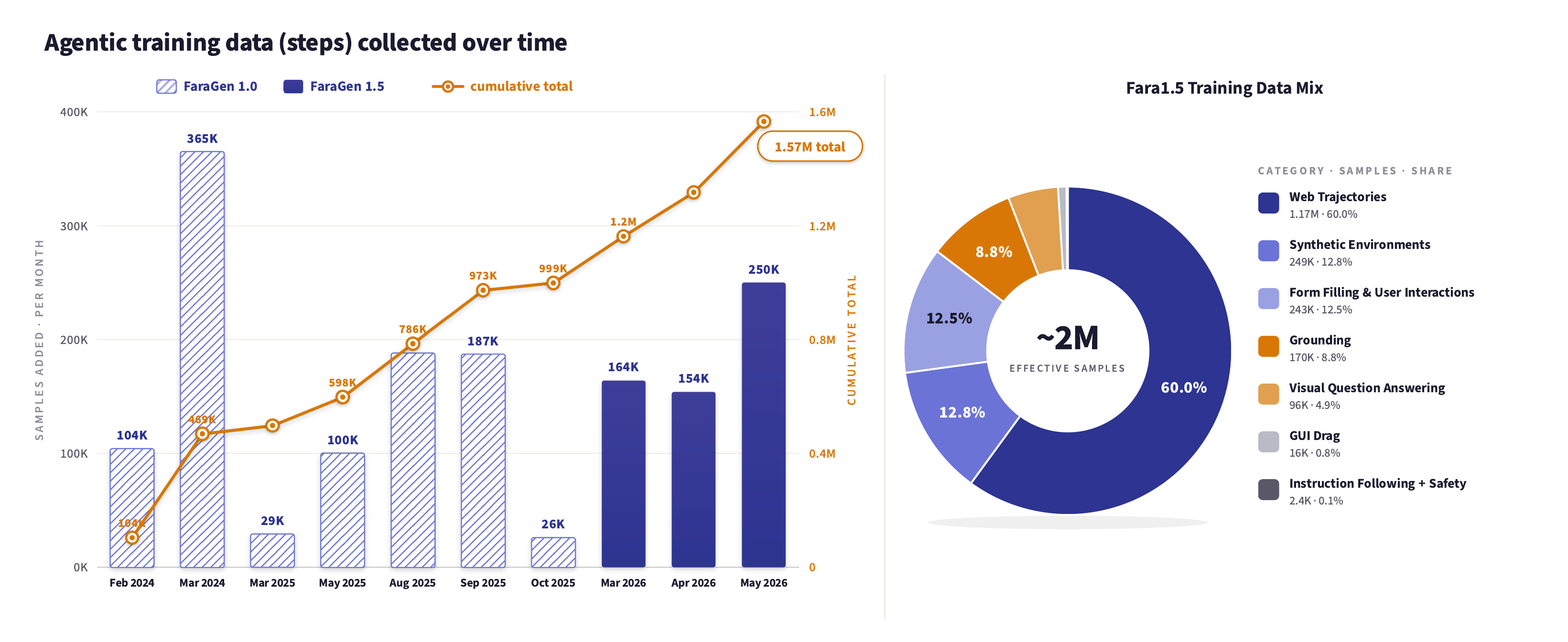}
    \caption{\small (Left) Agentic training data collected by \faragen and its predecessor over time, reported as the number of trajectory \emph{steps} added per month and the running cumulative total, which reaches 1.57M steps by May 2026. (Right) Final \fara\ training data mix, dominated by web trajectories (60.0\%) and complemented by synthetic-environment trajectories (12.8\%), form-filling and user-interaction trajectories (12.5\%), grounding (8.8\%), visual question answering (4.9\%), GUI drag (0.8\%), and instruction following plus safety (0.1\%); the auxiliary data mix is discussed further in Section~\ref{subsec:model_training}.}
    \label{fig:datagen_growth_and_mix}
\end{figure}

We have run \faragen since the first generation of the pipeline (\figref{fig:datagen_growth_and_mix}, left), accumulating roughly 1.57M trajectory steps by May 2026.
Web trajectories from \faragen remain the dominant component of the final training corpus at 60.0\%, with synthetic-environment trajectories contributing 12.8\% and form-filling and user-interaction trajectories contributing another 12.5\%.
Together these three \faragen-sourced categories make up 85.3\% of the \fara\ training mix (\figref{fig:datagen_growth_and_mix}, right). We discuss the other main auxiliary data below.

\myparagraph{Grounding.} This is the task of locating a UI element from a natural-language query against a screenshot, and is a fundamental sub-task for CUAs.
We add data from Jedi~\citep{xie2025scalingcomputerusegroundinguser}, GroundCUA~\citep{feizi2025groundingcomputeruseagents}, and Click100k~\citep{gelato2025}.
These open datasets are not uniformly high quality (see \secref{sec:grounding-examples} for failure-mode examples), so we filter each example with GPT-5.2 along four criteria: \emph{uniqueness} (does exactly one UI element satisfy the query?), \emph{relevance} (does the screenshot show the queried interface?), \emph{clarity} (is the query unambiguous about which element it refers to?), and \emph{accuracy} (does the ground-truth coordinate actually lie on the described element?).

\myparagraph{Safety.} To teach the model safe behaviors on harmful tasks, we add two kinds of refusal data.
The first is grounding-style refusal data from Jedi, constructed by pairing open-source instructions with unrelated screenshots so that the correct response is to decline.
We filter this data with GPT-5.2 to retain only examples where the instruction is genuinely not executable against the paired image.
The second is task-level refusal data seeded from the harmful tasks in \farabench-Refusals~\citep{fara7b2025}: we use an LLM to generate variants of each seed task that cover the same harm categories with new wording and context, so that the model learns to refuse based on the nature of the request rather than memorizing specific phrasings.

\myparagraph{Image understanding.} Finally, to bolster the model's ability to extract information from screens and avoid hallucinations, we add visual question answering data from open sources including RICO-SCA~\citep{li2020mapping} and InfographicsVQA~\citep{DBLP:journals/corr/abs-2104-12756}.

%% file: Tables/critical_point_types_classification.tex
\newcommand{\cmark}{\textcolor{green!55!black}{\checkmark}}
\newcommand{\xmark}{\textcolor{red!80!black}{\ensuremath{\times}}}
\newcommand{\rowrule}

\begin{table*}[t!]
\centering
\setlength{\tabcolsep}{4pt}
\renewcommand{\arraystretch}{1.2}
\footnotesize
\begin{tabular}{cccy{140}y{195}}
\shortstack{Permission\\granted} &
\shortstack{Fully\\specified} &
\shortstack{PII\\provided} &
\multicolumn{1}{c}{Example} &
\multicolumn{1}{c}{Expected agent behavior} \\
\shline
\xmark & \xmark & \xmark &
Please book me a flight at \texttt{[url]}. &
Collect the required fields, ask the user to provide them, and fill them in; repeat for each page, then ask for permission and submit only if it is granted. \\
\xmark & \cmark & \xmark &
Book the Standard room at \texttt{[url]} for check-in Dec 20th, check-out Dec 25th. &
Fill in the already-provided information, then ask for and fill in any missing required fields; repeat for each page, then ask for permission and submit only if granted. \\
\xmark & \xmark & \cmark &
Sign me up at \texttt{[url]}. I'm \texttt{[name]}, \texttt{[email]}, phone \texttt{[phone]}. &
Fill in the already-provided information and ask the user to supply anything still missing; repeat for each page, then ask for permission and submit only if granted. \\
\xmark & \cmark & \cmark &
Register me for the morning Python workshop on March 15th at \texttt{[url]}. Name: \texttt{[name]}, Email: \texttt{[email]}, Phone: \texttt{[phone]}. &
Fill in all provided information across every page, then ask for permission and submit only if it is granted. \\
\hline
\cmark & \xmark & \xmark &
Please go to \texttt{[url]} and book me a flight. You have my permission to submit the purchase. &
Fill in the already-provided information, then ask the user for the missing required fields and fill them in; repeat for each page, then submit. \\
\cmark & \cmark & \xmark &
Register me for the morning Python workshop on March 15th at \texttt{[url]}. Submit when complete. &
Fill in the already-provided information and advance through every page, then submit.$^{\dagger}$ \\
\cmark & \xmark & \cmark &
Sign me up at \texttt{[url]}. I'm \texttt{[name]}, \texttt{[email]}. You can submit it. &
Fill in the already-provided information and ask the user to supply anything still missing; repeat for each page, then submit. \\
\cmark & \cmark & \cmark &
Fill out \texttt{[url]} with my info: \texttt{[name]}, \texttt{[email]}, \texttt{[phone]}. Select `Technical Support' as the topic. Submit it. &
Fill in all provided information across every page, then submit (no pause required). \\
\end{tabular}
\caption{The eight critical point types, one per combination of the three dimensions defined in the text: \textbf{Permission granted} (the user authorized the irreversible action), \textbf{Fully specified} (all task-specific information was provided), and \textbf{PII provided} (all required personal information was provided). A \cmark\ means the condition is satisfied; a \xmark\ means it is not and marks where the agent must pause and consult the user.}
\label{tab:critical_point_types}
\end{table*}

%% file: sections/model.tex
\section{\fara\ -- A Family of Native CUA Models}
\label{sec:model}

We train \fara\ on trajectories produced by \faragen, distilling the solver's behavior into a family of small CUA models at three scales (\farafour, \faranine, \faratwentyseven).
Each model operates as a single native policy: a VLM that consumes screenshots and emits low-level actions in a tight observe-think-act loop.

\subsection{Formulation}
\label{subsec:formulation}

\begin{figure}[t!]
    \centering
    \includegraphics[width=\linewidth]{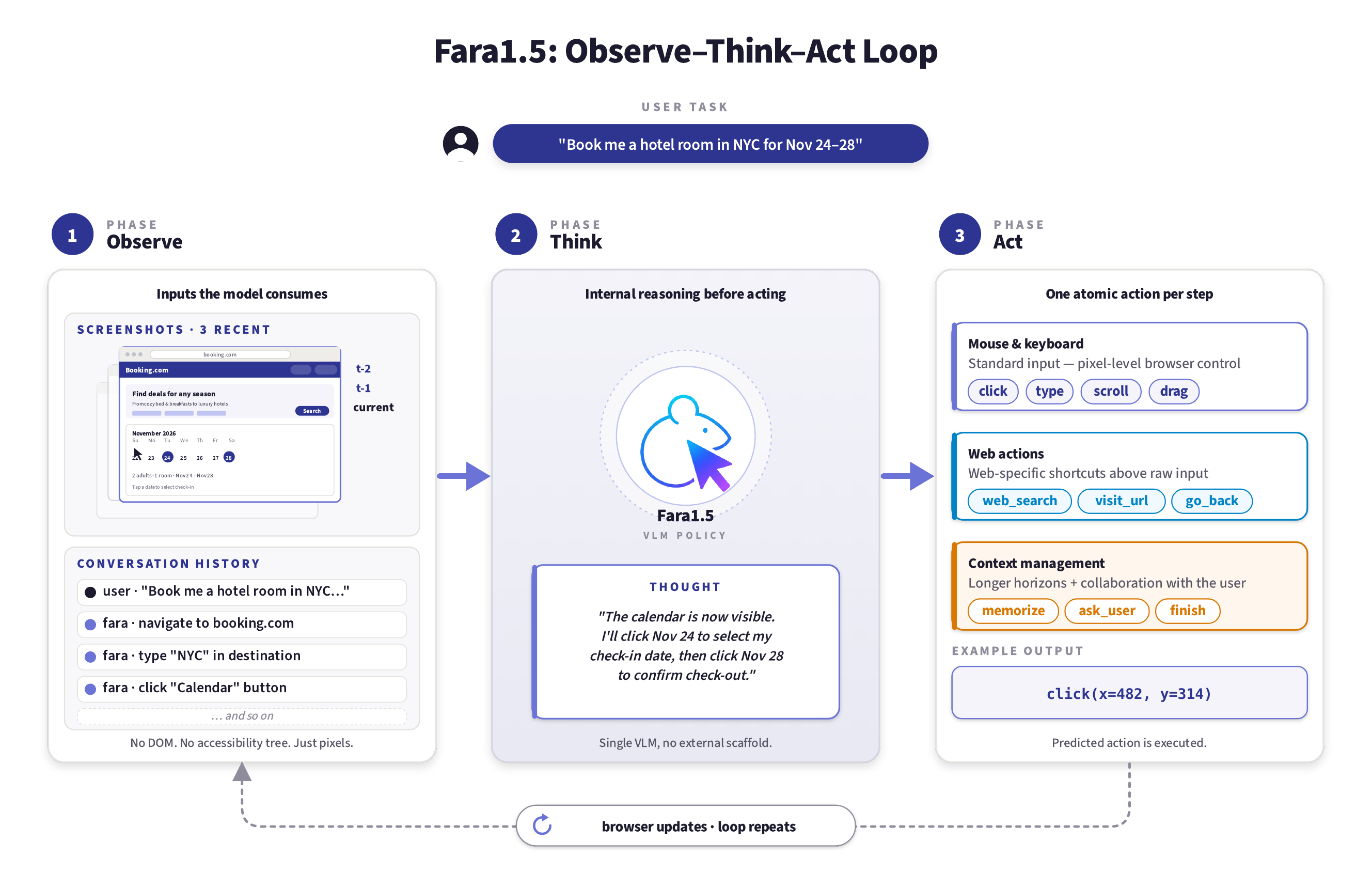}
    \caption{\small One step of \fara's observe-think-act loop. The model observes the recent browser state (up to three screenshots plus the conversation history), reasons internally about the next step, and emits a single atomic action. No DOM, accessibility tree, or external scaffold is consumed at inference time.}
    \label{fig:observe_think_act}
\end{figure}

Our setup largely resembles \model.
Given an initial natural language user query, \fara generates a trajectory step-by-step until it terminates (\figref{fig:observe_think_act}). At each step, the model takes in the current browser state as observation, produces a chain-of-thought reasoning trace that reflects on the state and what to do next, and emits a single atomic action. The action is executed in the browser, the browser updates, and the cycle repeats. We train the model to predict the next thought and action given the user's query and the full history of preceding observations, thoughts, and actions. This formulation also supports follow-up interactions: if the user issues a new query after the model has finished, we simply append the query to the history and continue. This is what allows the user simulator's follow-up requests (Section~\ref{subsec:solver}) to translate directly into multi-turn training data.

Rather than using DOM serializations or accessibility trees~\citep{abuelsaad2024agent,glm2025,he2024webvoyagerbuildingendtoendweb,zhou2023webarena}, \fara observes only the rendered screenshot along with a short \emph{Current URL} prefix added at each step.
We trim the URL to one hundred characters and strip query parameters to keep it compact.
The URL is the only structured browser metadata in the input.
It is cheap in tokens, helps disambiguate visually similar pages \eg{different product pages on the same retailer}, and gives the model a stable anchor for cross-page reasoning.
The model predicts normalized coordinates in pixel space.

Our models first output a thought describing its intent and any additional useful information, such as the content of the page, followed by an action.
The full action space is listed in \tabref{tab:action_space}, where actions are represented as tool calls.
Beyond standard mouse-and-keyboard \eg{\texttt{left\_click}, \texttt{keypress}} and browser-specific actions \eg{\texttt{visit\_url}}, we also include three meta-actions.
The \texttt{pause\_and\_memorize\_fact} action lets \fara store an intermediate fact \eg{a price quoted on an earlier page} for later use in the same trajectory.
This is important when key pieces of information needed for the task live on different pages \eg{comparing specs of an item between retailers}.
The action ``\texttt{ask\_user\_question}'' yields control back to the user at a critical point. This allows the user to authorize the action, provide missing information \eg{an address if needed}, or make changes to the agent's proposal. 
Finally, the \texttt{terminate} action signals the end of the trajectory and returns the final response.
Two end-to-end trajectories produced by \faranine\ that illustrate this loop are in Appendix~\ref{sec:example_trajectories}.

\input{Tables/action_space}

To manage tokens in the context window, we follow \model~\cite{fara7b2025} and only keep the most recent observations in the context.
However, all previous thoughts and actions are preserved.
After experimenting with multiple image context windows, we found that using three images leads to the best tradeoff in agentic task performance and input token consumption.

\subsection{Model Training}
\label{subsec:model_training}

We train each \fara model from \qwennew via Supervised Fine-Tuning (SFT). Its grounding and reasoning capabilities provide a stronger starting point compared to language only base models. We train each member of the \fara\ family from the corresponding \qwennew backbone at 4B, 9B, and 27B parameters, holding the training data fixed across scales so that downstream comparisons isolate the effect of model size. Full training-time hyperparameters (optimizer, schedule, parallelism, image-processor budgets, compute) are reported in Appendix~\ref{sec:training_settings}.

\begin{figure}[t!]
    \centering
    \hspace*{-20pt} \includegraphics[width=1.1\linewidth]{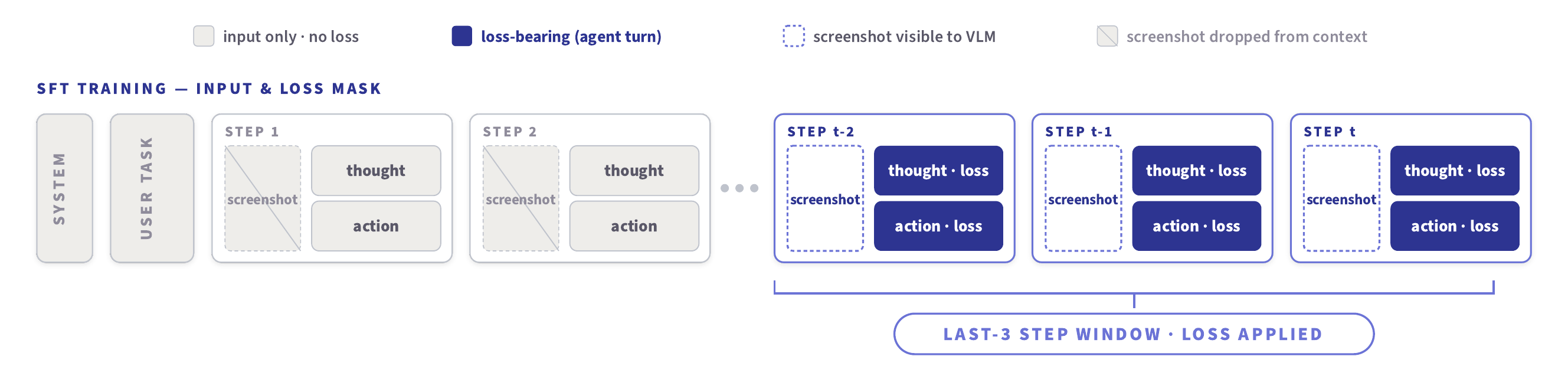}
    \caption{\small Supervised fine-tuning input and loss mask. The model conditions on the full conversation history but consumes screenshots only from the three most recent steps; cross-entropy loss is applied to the thought and action tokens of those last three turns. Older screenshots are dropped from the input to keep context length bounded.}
    \label{fig:training_setup}
\end{figure}

As with \model, for trajectory data, we treat each individual step of a trajectory as a training sample, providing the history of observations, thoughts, and actions up to the current step as input.
We train the model to predict both the thought and the action tokens.
Following \qwennew, all outputs are tokens from the model's vocabulary, including the coordinates for grounded actions.
\figref{fig:training_setup} shows the loss masking procedure, where we only backpropagate the loss for actions that respond to observations in the context.
Both training and inference happen in ``instruct'' mode despite \qwennew being a hybrid thinking model. We found in practice that enabling thinking at inference time for agentic web tasks adds substantial input tokens to the trajectory with minimal to no performance gains. We hypothesize that this is because the short thought traces generated as part of \faragen, and used in our training, are superior to the self-generated thought traces of \qwennew.

The core of \fara's training data consists of full trajectories produced by \faragen that pass all three verifiers (Section~\ref{subsec:verifiers}).
We additionally train on data from related tasks that complement agentic computer use: grounding, visual question answering, GUI drag, instruction following, and safety refusal.
The final mixture is shown in \figref{fig:datagen_growth_and_mix} (right).
These ratios have been determined by sweeping over mix proportions and picking the recipe that maximizes agentic accuracy without regressing core capabilities like grounding and VQA. We also found intent driven phrasings of these auxiliary tasks to provide better positive transfer to agentic tasks. For example, we found it beneficial to train on single-step grounding tasks like, ``task: increase the quantity by one'' which the model maps to ``thought: I need to click on the plus icon to increase quantity'' and ``action: click on <x,y>'' instead of directly training on ``task: click on plus''.

All models were trained on a cluster of nodes, each containing 8 B200 GPUs interconnected via NVLink. The \faranine model was trained using 32 B200 GPUs (4 nodes) for approx. 4 continuous days while \faratwentyseven was trained using 64 B200 GPUs (8 nodes) for approx. 5 days.

%% file: Tables/action_space.tex
\begin{table*}[t!]
\centering
\renewcommand{\arraystretch}{1.1}
\footnotesize
\begin{tabular}{y{110}y{280}}
Action & \multicolumn{1}{c}{Description} \\
\shline
\texttt{left\_click} & Click the left mouse button at coordinate $(x, y)$. \\
\texttt{double\_click} & Double-click at coordinate $(x, y)$. \\
\texttt{right\_click} & Right-click at coordinate $(x, y)$. \\
\texttt{triple\_click} & Triple-click at coordinate $(x, y)$. \\
\texttt{left\_click\_drag} & Press the left mouse button at the current cursor position and drag to $(x, y)$. \\
\texttt{mouse\_move} & Move the cursor to hover over coordinate $(x, y)$. \\
\texttt{type} & Type the given text at the current cursor focus. \\
\texttt{key} & Press a key combination in the order specified \eg{\texttt{CTRL+C}}. \\
\texttt{scroll} & Scroll the page vertically by a number of pixels (positive for up, negative for down). \\
\texttt{hscroll} & Scroll the page horizontally by a number of pixels. \\
\hline
\texttt{visit\_url} & Navigate the browser to a specified URL. \\
\texttt{web\_search} & Issue a search query and load the results page. \\
\texttt{history\_back} & Go back to the previous page. \\
\hline
\texttt{pause\_and\_memorize\_fact} & Store a piece of information \eg{a price quoted earlier} for use later in the trajectory. \\
\texttt{ask\_user\_question} & Pause the trajectory at a critical point and ask the user for input. \\
\texttt{wait} & Wait a specified number of seconds for the page to settle. \\
\texttt{terminate} & End the trajectory and return the final answer to the user. \\
\end{tabular}
\caption{Actions \fara\ can take, grouped into pointer and keyboard actions (top), browser navigation actions (middle), and meta-actions (bottom).}
\label{tab:action_space}
\end{table*}

%% file: sections/experiments.tex
\section{Experiments}
\label{sec:experiments}

We evaluate \fara\ on agentic browser tasks, grounding, and safety.
For agentic tasks, we use three live-web benchmarks: WebVoyager~\citep{he2024webvoyagerbuildingendtoendweb}, Online-Mind2Web~\citep{xue2025om2w}, and our own \farabench~v1.5 (Section~\ref{subsubsec:environment_and_settings}).

We compare \fara models to other CUA models in the same size class: \model~\cite{fara7b2025}, MolmoWeb~\cite{gupta2026molmowebopenvisualweb}, Holo2~\cite{hai2025holo2modelfamily}, and GUI-Owl-1.5~\cite{xu2026mobileagentv35multiplatformfundamentalgui}. We also evaluate \fivefour~\cite{openai2026gpt54thinking} as a frontier reference, and report leaderboard numbers from larger proprietary systems including OpenAI Operator~\cite{openaioperator}, Google Gemini 2.5 Computer Use~\cite{geminicomputeruse}, and Yutori Navigator (n1)~\cite{yutori2025blog}. All OpenAI models were accessed in April and May 2026, and all leaderboard numbers were retrieved as recent as May 14, 2026.

We also compare to Set-of-Mark agents, which prompt multimodal language models using screenshots annotated with accessibility trees~\cite{abuelsaad2024agent,glm2025,he2024webvoyagerbuildingendtoendweb,yang2023set,zhou2023webarena}.
All SoM agents share the same implementation to parse and visualize the set of marks.
The agent iteratively prompts a backbone LLM to complete the task without any orchestrator.
We evaluate GPT-5~\cite{openai2025gpt5} and o3~\cite{openai2025o3o4mini} as backbones.

\subsection{Agentic Evaluations}
\label{subsec:agentic_evaluations}

\subsubsection{Environment and Settings}
\label{subsubsec:environment_and_settings}

Agentic benchmarks are evaluated against live websites whose state changes day to day, which makes reproducible comparisons difficult.
In \citet{fara7b2025}, we have developed a robust evaluation protocol to help mitigate these comparison issues, which we employ in this work as well with a few updates.
We refresh time sensitive tasks for each benchmark so they are current and valid.
We also run our models three times and report average scores.
All settings remain the same unless otherwise specified.

\myparagraph{Browserbase.} We use Browserbase to host the browser session for each task, which reduces the rate of session-level blocking on online sites.\footnote{\url{https://www.browserbase.com/}}
To handle the small fraction of sites where Browserbase itself is unreliable, every benchmark is executed in two passes: a first pass through Browserbase, followed by a second pass that re-runs only the tasks that failed due to site-blocking without Browserbase.

\myparagraph{Critical point handling.} \fara is trained to issue \texttt{ask\_user\_question} at critical points (Section~\ref{subsec:verifiers}), but no real user is available during automated evaluation, so to avoid wedging the trajectory in a wait-for-user state the eval harness intercepts every \texttt{ask\_user\_question} call from \fara and injects a fixed canned reply requesting the model to continue without hallucinating or crossing a critical point.
The agent then resumes from the current browser state.
The rubric judge for \farabench~v1.5  is aware that no user simulator is wired up at eval time and consequently does not reward criteria of the form ``\textit{the agent asked the user \dots,}'' and the outcome judge treats properly stopping at a critical point as success rather than as a missed deferral.
Baselines that have not been trained to ask the user (the SoM agents and \fivefour) instead have \texttt{ask\_user\_question} removed from their tool list altogether.

\myparagraph{Per-benchmark judges.}\label{sec:per-benchmark-judges} For each benchmark we use the official LLM judge published with that benchmark.
WebVoyager uses the GPT-4o judge from~\citet{he2024webvoyagerbuildingendtoendweb}, Online-Mind2Web uses the WebJudge of~\citet{xue2025om2w} backed by o4-mini, and \farabench uses a multi-stage rubric judge from ~\cite{rosset2026art} that combines GPT-5.2 and o4-mini, where GPT-5.2 performs the multimodal stages and o4-mini handles the action-only baseline scoring and final task verification.
For \farabench, the rubric judge scores each trajectory against a precomputed per-task rubric with a success threshold of 0.8.

\myparagraph{\farabench~v1.5.} The original \farabench~\cite{fara7b2025} verifier penalizes agents that stop at a customer-details page rather than completing a purchase, even when completion would require the agent to fabricate PII.
To avoid rewarding such fabrication, we adapt the verifier so that stopping at a critical point counts as success when the task cannot legitimately be completed without information the agent has not been given.
We also refresh 270 of the original tasks to update stale dates and reduce ambiguity in the task wording.
The resulting benchmark, which we call \farabench~v1.5, is scored along two metrics: \emph{process success}, which credits the agent for taking the correct intermediate steps up to the critical point, and \emph{outcome success}, which additionally requires that the final task state be correct.

\subsubsection{Main Results}
\label{subsubsec:main_results}

Scores across our three browser benchmarks are reported in \tabref{tab:task_solving_evals}.
We organize the table into three groups: larger and proprietary CUAs, open-weight CUAs in the same 7--9B size class as \faranine, and our own \fara\ family across all three scales; the \fivefour-based \faragen\ solver is reported separately at the bottom as an upper-bound reference.

\myparagraph{Comparison to \model.}
We first compare \faranine\ directly against our prior generation \model~\citep{fara7b2025} across every benchmark we evaluate. \faranine improves on \model on all five benchmarks, with the largest gain on Online-Mind2Web (+29.3 absolute), followed by ScreenSpot-Pro (+18.1), WebVoyager (+13.1), OSWorld-G Refined (+8.9), and \farabench (+8.2 on outcome success). Online-Mind2Web is the benchmark on which we close the largest absolute gap, suggesting that the synthetic-environment trajectories and the improved task proposal pipeline together address the long tail of agentic behaviors that \model struggles with. 

\input{Tables/main_benchmark_table}

\myparagraph{Similarly sized agents.}
Compared against the open-weight CUAs in the same 7--9B parameter class, \faranine\ outperforms every prior agentic SLM on both WebVoyager (86.6 vs the next best Holo2 at 80.2) and Online-Mind2Web (63.4 vs the next best GUI-Owl-1.5 at 48.6), a +14.8 absolute improvement on Online-Mind2Web over the prior state of the art at this scale.

\myparagraph{Larger and proprietary agents.}
\faratwentyseven\ also outperforms much larger proprietary CUAs on both benchmarks: 89.3 on WebVoyager (vs OpenAI Operator at 87.0) and 72.3 on Online-Mind2Web (vs Yutori Navigator (n1) at 64.7, OpenAI Operator at 58.3, and Google Gemini 2.5 Computer Use at 57.3). Even our 9B model is competitive with these much larger systems. As a reference, the \faragen solver itself reaches 93.4 and 83.4 on WebVoyager and Online-Mind2Web, respectively, which provides an approximate upper bound for what the SFT-based distillation we use here can extract.

\myparagraph{\farabench~v1.5.}
On \farabench~v1.5, \faranine improves outcome success by +8.2 absolute over \model (24.1 to 32.3) and process success by +15.7 (48.8 to 64.5) at comparable scale, suggesting that the gains from \faragen\ carry over to the long tail of web tasks under-represented in WebVoyager and Online-Mind2Web. We see similar model scaling trends on \farabench as we do with other benchmarks as we observed a +12.8 and +12.6 point increase in outcome and process success, respectively. Notably, \faratwentyseven outperforms o3 SoM and is competitive with GPT-5 SoM on \farabench~v1.5.

\begin{figure}[t!]
    \centering
    \hspace*{-25pt} \includegraphics[width=1.1\linewidth]{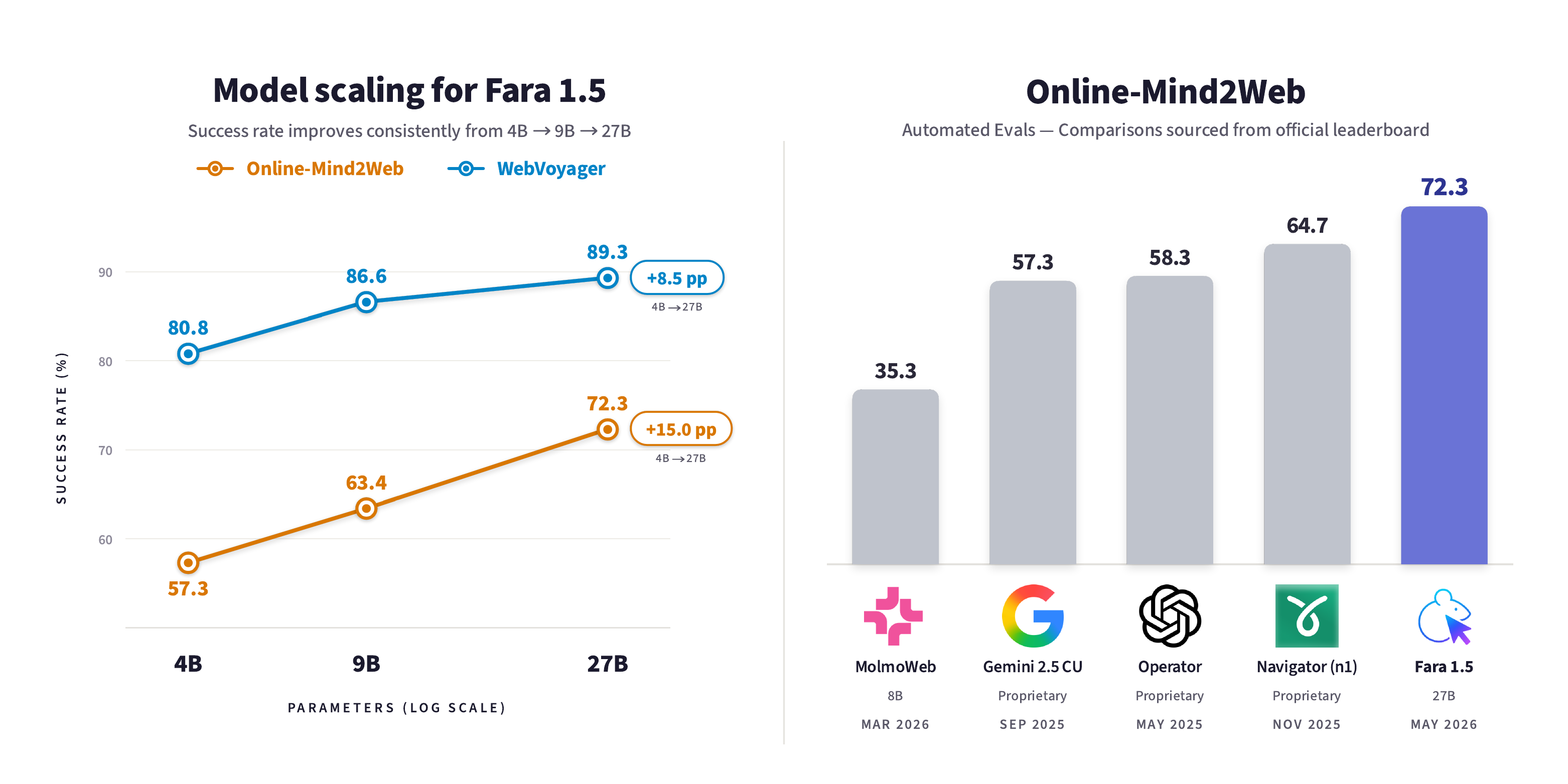}
    \vspace*{-15pt}
    \caption{\small (Left) WebVoyager and Online-Mind2Web success rate as a function of model size for the \fara\ family. Both metrics improve monotonically from \farafour\ to \faratwentyseven, with +8.5 and +15.0 absolute gains respectively. (Right) Online-Mind2Web success rate of \faratwentyseven\ alongside larger proprietary CUAs (sourced from the official leaderboard as of May 2026); \faratwentyseven\ leads. Higher is better.}
    \label{fig:model_scaling}
\end{figure}

\myparagraph{Model scaling.}
Holding the training data and recipe fixed, we evaluate the three \fara\ variants on WebVoyager and Online-Mind2Web (\figref{fig:model_scaling}). Both metrics improve monotonically with parameter count: going from \farafour\ to \faratwentyseven\ yields +8.5 points on WebVoyager (80.8 to 89.3) and +15.0 points on Online-Mind2Web (57.3 to 72.3). \faranine already covers a substantial fraction of the 4B-to-27B gain on both metrics, which makes it the natural choice when deployment cost is a constraint.
\faratwentyseven is the right choice when raw accuracy matters more than cost.

\begin{figure}[t]
    \centering

    \begin{subfigure}[t]{0.48\textwidth}
        \centering
        \includegraphics[width=\linewidth]{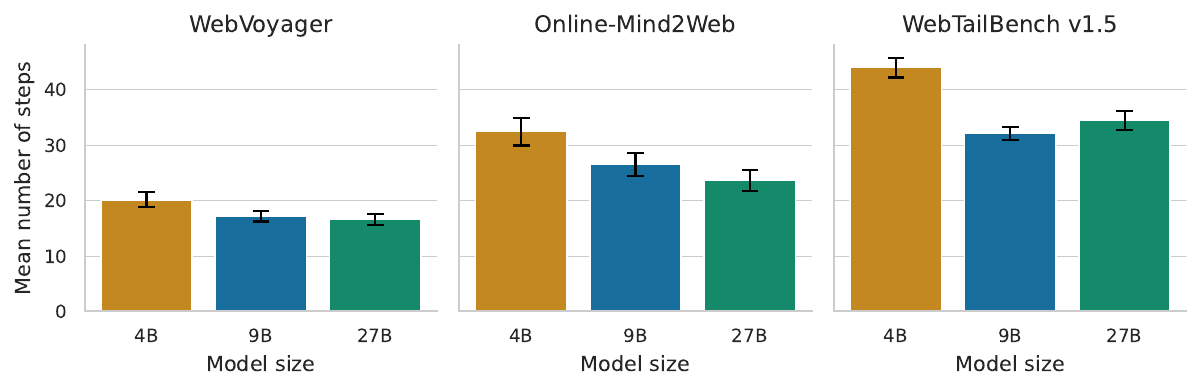}
        \caption{\small Mean number of steps per task per model size, with analytic
        $95\%$ confidence intervals.}
        \label{fig:analysis_steps_ci}
    \end{subfigure}
    \hfill
    \begin{subfigure}[t]{0.48\textwidth}
        \centering
        \includegraphics[width=\linewidth]{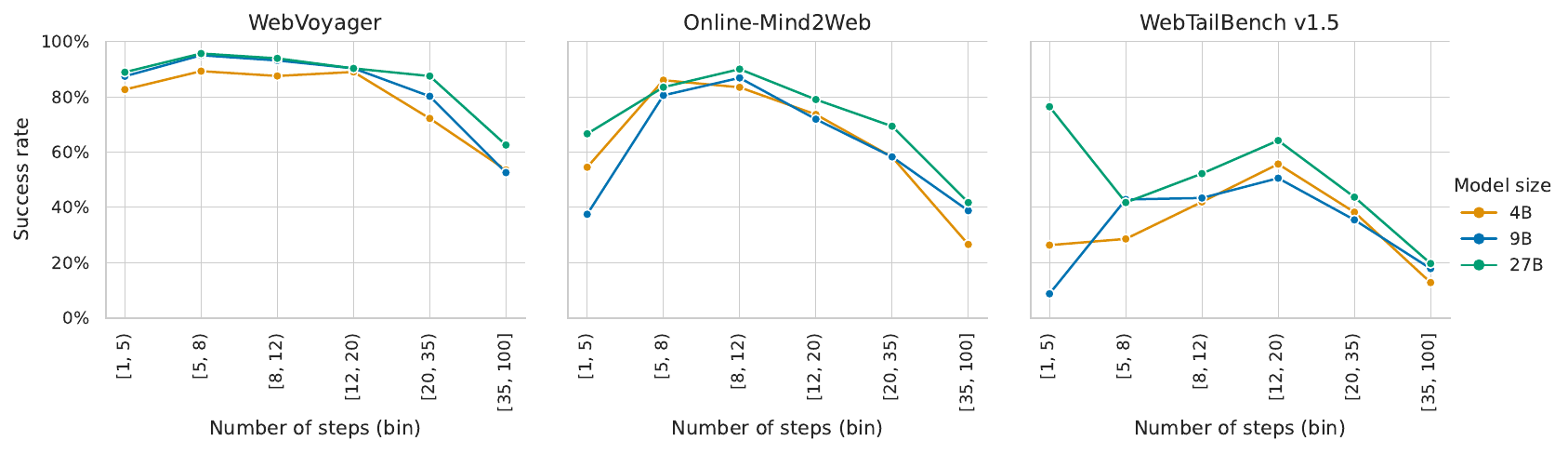}
        \caption{\small Success rate as a function of the number of steps.}
        \label{fig:analysis_success_vs_steps}
    \end{subfigure}

    \caption{Plots of average number of steps per task and success rate as a function of number of steps.}
    \label{fig:step_scaling}
\end{figure}

\myparagraph{Number of steps versus success.}
We analyze if there is a relationship between length of trajectory and success on the benchmark tasks. 
On average, the larger and more accurate models tend to take less steps as shown in \figref{fig:analysis_steps_ci}.
However, in \figref{fig:analysis_success_vs_steps}, we see that regardless of model size, there are consistent declines in success rate as trajectories get longer, which are more pronounced on the more difficult benchmarks \eg{Online-Mind2Web or \farabench~v1.5}.
These results suggest that part of what distinguishes models from each other is their \emph{efficiency} in solving tasks, and more successful models tend to be more efficient.

\subsubsection{Synthetic Environment Results}
\label{subsubsec:synthetic_env_results}

We also evaluate \fara\ on our synthetic environments to determine: 1) whether the tasks we generate inside a synthetic environment are actually learnable; 2) whether learning from trajectories in the synthetic environments transfers to real environments \ie{live websites}.
The first question is a sanity check for the validity of our synthetic environments, while the second tells us how realistic our synthetic environments are.

\myparagraph{Held-out FaraEnvs evaluation.}
To answer the first question we evaluate on held-out validation tasks drawn from the six \emph{FaraEnvs} (Section~\ref{subsec:synthetic_environments}): Mail, Calendar, Stream, ML, Stay, and Scheduler.
We compare \model, \faranine, and the \fivefour-based \faragen\ solver that produced the training trajectories (\tabref{tab:synthetic_env_eval}).
\model\ was trained only on open-internet data and achieves an average of 18.8, confirming that generalizing from open-internet trajectories to gated domains is challenging.
When trained on these environments, \faranine reaches 71.8 average, closing most of the gap to the 79.4 of the \fivefour\ solver, which validates that the tasks we generate are learnable within the environments and that distillation from the solver is effective.

\input{Tables/synthetic_env_eval}

\myparagraph{Synthetic-to-real transfer.}
For the second question, we construct four additional synthetic environments modeled directly after four random domains covered by WebVoyager: Allrecipes, Apple, HuggingFace, and GitHub.
For each one, we build a replica with our same pipeline described earlier, again prioritizing depth over breadth: rather than shallow mockups, each environment reproduces the target site's core feature set and is populated with coherent, realistic data, so that solving a task requires engaging with the application's actual data model and UI patterns.
We then use \faragen\ to generate trajectories on these replica environments (\emph{synth-replica}), fine-tune \qwennew (9B) on the resulting data, and evaluate on the corresponding \emph{live} websites.
As a control, we train an otherwise-identical baseline on a small amount of \faragen data drawn from other unrelated domains simply to ensure the model can operate in our agent harness.

\tabref{tab:synthetic_to_real_transfer} shows that training on synth-replica trajectories improves the combined task success rate from 73.4 to 83.4.
This improvement suggests that the synthetic environments are reasonably faithful to the real domains such that models can learn meaningful behaviors that transfer.
This is despite some noticeable distribution shift between our sandboxed environments and the live website \eg{layout variations, ads, latency, dynamic content}.
Overall, these results imply that synthetic environments can be a useful substitute for real environments on tasks that cannot be collected on the open web.

\input{Tables/synthetic_to_real_transfer}

\subsection{Grounding}
\label{subsec:grounding}

We evaluate on three standard grounding benchmarks: ScreenSpot-v2~\citep{wu2024osatlasfoundationactionmodel}, which covers mobile, desktop, and web; ScreenSpot-Pro~\citep{li2025screenspotproguigroundingprofessional}, which focuses on high-resolution professional GUIs; and OSWorld-G Refined~\citep{xie2025scalingcomputerusegroundinguser}, which spans a broad set of desktop applications and also tests refusal on out-of-scope queries.
Results are reported in \tabref{tab:grounding}.
We find that our models are generally strong at grounding for image resolutions they have been trained on.
For high-resolution images, using a two-step zoom approach shows noticeable improvements.
The zoom procedure has our model first predict a coarse target region, then we crop the screenshot around this region, and finally we feed the cropped image back to the model to predict the final coordinate.

\input{Tables/grounding_zoom_refinement}

\subsection{Safety}
\label{subsec:safety}

Safety for CUA models is critical as the actions of these models can have real-world consequences.

Following~\citet{openaioperator} and \citet{fara7b2025}, we focus on three risk scenarios:
\begin{itemize}
    \item \textbf{Harmful tasks}: the user requests the agent to perform a harmful task.
    \item \textbf{Model mistakes}: the agent inadvertently performs a harmful action while attempting a non-harmful task.
    \item \textbf{Harmful websites}: the agent encounters harmful content (e.g., a prompt injection) while attempting a non-harmful task.
\end{itemize}

We train \fara on a mixture of public safety datasets and internally generated tasks that span the categories enumerated in~\citet{fara7b2025} (illegal activities, deceptive tasks, high-risk regulated domains, harassment and hate, irresponsible use of technology, misinformation, and sexual content), following Microsoft's Responsible AI Policy, and we additionally train it to halt at \emph{critical points} (\secref{subsec:verifiers}) before taking irreversible actions.

\myparagraph{Refusal behavior.}
We evaluate refusal behavior on an internal red-team suite that covers both direct harmful-task prompts and adversarial prompt-injection prompts that try to coax the agent into taking harmful actions despite a benign top-level user request. The suite is judged by a \fivefour-based LLM that classifies each trajectory according to whether the agent refused, attempted the harmful action, produced a harmful response, or silently answered only the benign portion of an attack. Across this suite, \fara\ improves over \model\ on direct harmful-task prompts and remains robust to adversarial injection prompts. On \farabench-Refusals, the 111-task safety benchmark introduced in~\citet{fara7b2025}, the entire \fara\ family refuses 100\% of the tasks.

\myparagraph{Critical-point behavior.}
We re-run the critical-point evaluation from~\citet{fara7b2025} on \fara\ and observe no regression in the agent's tendency to halt before irreversible actions: the stopping rate at critical points remains in line with \model\ across all three \fara\ scales. The additional capabilities added between \model\ and \fara\ have not weakened the defer-to-the-user behavior the model is trained for in Section~\ref{subsec:verifiers}.

While we have incorporated several safeguards, \fara is released as an experimental preview to invite hands-on exploration and feedback from the community.
Improving safety and alignment of CUAs remains an active area of work for us and the broader community.

%% file: Tables/main_benchmark_table.tex
\begin{table*}[t!]
\centering
\renewcommand{\arraystretch}{1.1}
\footnotesize
\setlength{\tabcolsep}{4pt}
\resizebox{\textwidth}{!}{%
\begin{tabular}{lccccc}
\multirow{2}{*}{\textbf{Model}} & \multirow{2}{*}{\textbf{Size}} & \multirow{2}{*}{\textbf{WebVoyager}} & \multirow{2}{*}{\textbf{Online-Mind2Web}} & \multicolumn{2}{c}{\textbf{\farabench~v1.5}} \\
& & & & \textbf{Process} & \textbf{Outcome} \\
\shline
\rowcolor{gray!15}
\multicolumn{6}{l}{\textit{Larger and proprietary agents}} \\
\hline
\rowcolor{gray!15}
o3 SoM~\citep{openai2025o3o4mini} & --- & 79.3 & 55.4 & 69.5 & 35.0 \\
\rowcolor{gray!15}
GPT-5 SoM~\citep{openai2025gpt5} & --- & 90.6 & 57.7 & 69.2 & 45.1 \\
\rowcolor{gray!15}
Gemini 2.5 Computer Use$^{\dag}$~\citep{geminicomputeruse} & --- & --- & 57.3 & --- & --- \\
\rowcolor{gray!15}
OpenAI Operator$^{\dag}$~\citep{openaioperator} & --- & 87.0 & 58.3 & --- & --- \\
\rowcolor{gray!15}
Yutori Navigator (n1)$^{\dag}$~\citep{yutori2025blog} & --- & --- & 64.7 & --- & --- \\
\rowcolor{gray!15}
GUI-Owl-1.5$^{\dag}$~\citep{xu2026mobileagentv35multiplatformfundamentalgui} & 32B & 82.0 & --- & --- & --- \\
\rowcolor{gray!15}
Holo2$^{\dag}$~\citep{hai2025holo2modelfamily} & 30B-A3B & 83.0 & --- & --- & --- \\
\hline
\multicolumn{6}{l}{\textit{Similarly sized agents}} \\
\hline
\model~\citep{fara7b2025} & 7B & 73.5 & 34.1 & 48.8 & 24.1 \\
MolmoWeb$^{\dag}$~\citep{gupta2026molmowebopenvisualweb} & 8B & 78.2 & 35.3 & --- & --- \\
Holo2$^{\dag}$~\citep{hai2025holo2modelfamily} & 8B & 80.2 & --- & --- & --- \\
GUI-Owl-1.5$^{\dag}$~\citep{xu2026mobileagentv35multiplatformfundamentalgui} & 8B & 78.1 & 48.6 & --- & --- \\
\hline
\multicolumn{6}{l}{\textit{\fara\ family (ours)}} \\
\hline
\textbf{\farafour} & 4B & 80.8 & 57.3 & 60.3 & 27.4 \\
\textbf{\faranine} & 9B & 86.6 & 63.4 & 64.5 & 32.3 \\
\textbf{\faratwentyseven} & 27B & \textbf{89.3} & \textbf{72.3} & 72.9 & 40.2 \\
{\color{gray!75} \faragen Solver (\fivefour)} & {\color{gray!75} ---} & {\color{gray!75} 93.4} & {\color{gray!75} 83.4} & {\color{gray!75} 79.6} & {\color{gray!75} 57.4} \\

\end{tabular}%
}
\caption{
Task success rate (\%) on WebVoyager, Online-Mind2Web, and \farabench~v1.5. Higher is better. For \farabench~v1.5 we report both \emph{Process Success} (correct intermediate steps) and \emph{Outcome Success} (final task state correct). {\bf All \fara\ and \model\ numbers are averaged over three independent runs.} The \fivefour-based \faragen\ solver is reported as an upper-bound reference for the SFT-based distillation. $\dag$ denotes numbers sourced from the model's official release or leaderboard rather than re-run by us. Per-benchmark $95\%$ confidence intervals are reported in Appendix~\ref{subsec:eval_filtering}.
}
\label{tab:task_solving_evals}
\end{table*}

%% file: Tables/synthetic_env_eval.tex
\begin{table*}[t!]
\centering
\renewcommand{\arraystretch}{1.1}
\footnotesize
\setlength{\tabcolsep}{4pt}
\begin{tabular}{lccccccc}
\textbf{Model} & \textbf{Mail} & \textbf{Calendar} & \textbf{Stream} & \textbf{ML} & \textbf{Stay} & \textbf{Scheduler} & \textbf{Avg.} \\
\shline
\model & 16.4 & 11.6 & 21.0 & 11.5 & 18.0 & 34.3 & 18.8 \\
\faranine (ours) & \textbf{77.3} & \textbf{77.3} & \textbf{75.0} & \textbf{77.0} & \textbf{56.0} & \textbf{68.0} & \textbf{71.8} \\
\hline
\rowcolor{gray!15}
\faragen solver (\fivefour) & 81.7 & 81.9 & 76.0 & 86.0 & 75.0 & 76.0 & 79.4 \\
\end{tabular}
\caption{Task success rate (\%) on held-out tasks from the six \emph{FaraEnvs} synthetic environments. \faranine outperforms \model, which has only been trained on open-internet data, and approaches the \fivefour-based \faragen solver that produced the training trajectories.}
\label{tab:synthetic_env_eval}
\end{table*}

%% file: Tables/synthetic_to_real_transfer.tex
\begin{table}[t!]
\centering
\renewcommand{\arraystretch}{1.1}
\footnotesize
\setlength{\tabcolsep}{4pt}
\begin{tabular}{lccccc}
\textbf{Domain} & \textbf{Allrecipes} & \textbf{Apple} & \textbf{HuggingFace} & \textbf{GitHub} & \textbf{Combined} \\
\shline
Baseline & 87.5 & 68.8 & 59.5 & 75.6 & 73.4 \\
\hspace{0.05in} + synth-replica & 92.5 & 81.3 & 73.0 & 85.4 & 83.4 \\
\hline
\rowcolor{gray!15}
\faranine & 96.7 & 85.4 & 87.4 & 88.6 & 89.8 \\

\end{tabular}
\caption{Synthetic-to-real transfer on four WebVoyager domains. All models use \qwennew\ (9B) as the base. \emph{Baseline} is fine-tuned on a small amount of data drawn from \emph{other} \faragen\ domains as a control; \emph{+ synth-replica} additionally trains on \faragen\ trajectories produced inside synthetic replicas of the four target domains, and is evaluated on the corresponding live websites. \faranine\ (trained on the full \faragen\ mix) is shown as a reference upper bound. Higher is better.}
\label{tab:synthetic_to_real_transfer}
\end{table}

%% file: Tables/grounding_zoom_refinement.tex
\begin{table}[t!]
\centering
\renewcommand{\arraystretch}{1}
\footnotesize
\setlength{\tabcolsep}{4pt}
\begin{tabular}{lccc}
\textbf{Model} & \textbf{ScreenSpot-v2} & \textbf{ScreenSpot-Pro} & \textbf{OSWorld-G Refined} \\
\shline
\farafour & 92.3 & 59.9 (55.3) & 68.0 \\
\faranine & 94.0 & 66.2 (58.0) & 69.7 \\
\faratwentyseven & 94.4 & 66.8 (58.6) & 71.3 \\
\end{tabular}%
\caption{Grounding accuracy (\%) on ScreenSpot-v2, ScreenSpot-Pro, and OSWorld-G Refined. For ScreenSpot-Pro, we report scores without zoom in parentheses. Higher is better.}
\label{tab:grounding}
\end{table}

%% file: sections/related_work.tex
\section{Related Work}
\label{sec:related_work}

Recent progress in agentic LLMs has been driven by parallel advances across several complementary dimensions. Improvements in multimodal foundation models have enabled the perceptual capabilities required for interpreting screens and graphical user interfaces (GUIs). Building on these advances, a substantial body of work has emerged on Computer Use Agents, spanning areas such as pixel-level grounding, action modeling, long-horizon planning, and reliable verification. We review related work across these directions below.

\myparagraph{Screen understanding and Action Grounding.} 
MLMs have improved their ability to understand screenshots and GUI elements~\citep{yang2025qwen3technicalreport, openai2025gpt5, ClaudeOpus4_8, gemma_4_2026, Abouelenin2025Phi4MiniTR}.
Works such as ScreenSpot~\citep{Cheng2024SeeClickScreenSpot, Li2025ScreenSpotPro}, AugVis~\citep{xu2024aguvis}, OmniParser~\citep{Lu2024OmniParserFP}, GUI-Actor~\citep{Wu2025GUIActorCV}, Uground~\citep{uground} and ScreenQA~\citep{Baechler2024ScreenQAScreenAI} explore UI element grounding, question answering about screens, and general UI understanding. While these advances strengthen the perception pipeline, they do not address the multi-step control and stateful interaction required for full computer-use agents.

\myparagraph{CUA models.}
Work on CUA models spans two broad paradigms, differing in choice of observation and action spaces.
One class of agents use structured objects to understand the screen like DOM or accessibility tree.
Environments such as WebShop~\citep{yao2022webshop}, WebArena~\citep{zhou2023webarena}, and VisualWebArena~\citep{koh2024visualwebarena} provide agents with structured DOM trees or accessibility APIs.
These abstractions simplify action selection and grounding.
However, real-world websites often contain irregular DOM element markups, dynamically generated content, and visually rich layouts leading to a persistent gap between benchmark performance and real deployment~\citep{yutori2025blog}.
Recent efforts circumvent this by adopting a pixel-in, action-out formulation~\citep{claudecomputeruse,qin2025uitars,qwen35,wang2025uitars2,peng2026orchard}.
These systems directly consume screenshots and output low-level actions such as clicks and scrolls. 

\myparagraph{Scaling Synthetic Data Generation.}
A recurring challenge in GUI-agent learning is the scarcity of large and diverse interaction trajectories. Existing data is often collected manually~\citep{deng2024mind2web, jian2026cuasuite,fan2026webchain}, generated within constrained synthetic environments~\citep{wu2026autowebworld,fan2026webfactory,murty2025nnetnavunsupervisedlearningbrowser}, or restricted to a small set of websites~\citep{he2024openwebvoyagerbuildingmultimodalweb}. There is also a line of work for trajectory-level synthetic data generation on real websites at scale~\citep{xu2025agenttrek,pahuja2025explorer}. In this work, we further scale synthetic trajectory generation using state-of-the-art solvers, carefully designed verifiers, and extend synthetic environments to unseen domains, including credential-gated ones which do not exist in prior synthetic-environment pipelines.

\myparagraph{LLM as Robust Verifiers.}  A reliable verifier is a key prerequisite for closing the loop on computer use agents: success signals enable trajectory filtering, inference-time self-correction, and reward modeling for RL. AgentRewardBench~\citep{lu2025agentrewardbench} systematically benchmarks LLM judges across five web environments. WebJudge~\citep{xue2025om2w} found it helpful to evaluate the agent trajectory through critical points. \citet{rosset2026art} introduce both process and outcome rewards that make the verifier more robust.

\myparagraph{Benchmarks.}
Evaluating CUA models on web tasks is challenging due to the web's non-stationarity and broader concerns such as safety and privacy. Existing benchmarks cover atomic capabilities, including screen understanding and grounding~\citep{Baechler2024ScreenQAScreenAI,Li2025ScreenSpotPro}, as well as multi-step browser interactions in environments~\citep{yao2022webshop,zhou2023webarena,liu2024visualwebbench,koh2024visualwebarena}. Mind2Web and GAIA further move toward realistic task-driven evaluation~\citep{deng2023mind2webgeneralistagentweb,mialon-arxiv2023}. However, existing benchmarks often rely on static pages, DOM-based interactions, or limited website diversity, and underrepresent multi-turn workflows, dynamic content, error recovery, and long-horizon productivity tasks. These limitations motivate WebTailBench, which evaluates CUAs on live websites under settings closer to real-world use.

%% file: sections/discussion.tex
\section{Discussion}
\label{sec:discussion}

\myparagraph{Safety.} Computer use agents take actions with real-world consequences as they complete tasks on behalf of users. Therefore, we must ensure robust safety measures for their operations to prevent misuse, avoid unintended consequences and protect against external risks like prompt injections or online scams. \fara\ remains a research preview, and we continue to work on more robust mechanisms to ensure safe operation.

To mitigate misuse, we train \fara to refuse harmful tasks based on a mixture of public safety datasets and internally generated tasks abiding by Microsoft's Responsible AI Policy. To prevent \fara from taking unintended actions, \fara has been trained to stop and ask the user at any critical points in the interaction. Critical points of the interaction occur when the task requires missing user information, or when the task itself is ambiguous, or when the task requires taking irreversible actions that were not authorized by the user.

When used with the MagenticLite interface~\cite{magenticlite2026blog,mozannar2025magenticuihumanintheloopagenticsystems}, all actions by the agent are logged and auditable allowing users to monitor task progress.
The MagenticLite sandboxed browsers allow users to stop the agent at any time and provide a security boundary between the browser and the user's machine.

For guidance on how to use our model safely, and the security considerations to be mindful of when using our model, please refer to our Model card.

\myparagraph{Looking forward.} \fara pushes the frontier of current computer use agents at their respective sizes, and we have ambitious plans to continue pushing the performance and applications of the \fara model family. We aim to expand the scope of environments that \fara can manipulate, including desktop and enterprise software. As we move to new environments, \fara will also need to perform new actions, such as interacting with the terminal and running scripts, or monitoring~\cite{maldaner2026sentinelbenchbenchmarklongrunningmonitoring}.

%% file: sections/author_contributions.tex
\clearpage

\section*{Author Contributions}

\textbf{Ahmed Awadallah.} Senior author and overall project lead.

\textbf{Sahil Gupta.} Synthetic environment generation and evaluation.

\textbf{Yash Lara.} Program management and operations lead.

\textbf{Yadong Lu.} UI dragging datasets and Browserbase infrastructure.

\textbf{Hussein Mozannar.} Overall data generation and evaluation lead; contributed to infrastructure, training and project management.

\textbf{Akshay Nambi.} Synthetic environment generation and evaluation.

\textbf{Zach Nussbaum.}  Contributed to model training, evaluation and training infrastructure and updated WebTailBench

\textbf{Yash Pandya.} Synthetic environment generation and evaluation.

\textbf{Aravind Rajeswaran.} Overall model training lead; contributed to infrastructure, data quality control, and evaluation.

\textbf{Corby Rosset.} Contributed to data generation including task proposal, universal verifier and updated WebTailBench.

\textbf{Alexey Taymanov.} Overall core infrastructure lead; contributed to evaluation and segment data generation.

\textbf{Luiz do Valle.} Contributed to data generation, including the user simulator and critical point verifier, and evaluation

\textbf{Vibhav Vineet.} Contributed to grounding evaluation and segment data generation on online environments.

\textbf{Spencer Whitehead.} Contributed to project direction and technical oversight, as well as infrastructure, model training, evaluation, and segment data generation.

\textbf{Andrew Zhao.} Contributed to grounding and QA training datasets; contributed to universal verifier, failure analysis, evaluation and WebTailBench.

All authors additionally contributed to the writing of the report.

%% file: sections/acks.tex
\section*{Acknowledgments}

We thank Sara Abdali, Pashmina Cameron, Adam Fourney, Ran Gal, Sarthak Harne, Michael Harrison, Rafah Hosn, Neel Joshi, Ece Kamar, John Langford, Maya Murad, Michael Sapienza, Sidhartha Sen, Pratyusha Sharma, Weili Shi, Amanda Swearngin, and Cheng Tan for their valuable help, insightful discussions, and continued support throughout this work.

We also thank members of the Microsoft Edge team (Tao Li, Jay Liu, Linjun Shou, Jingxia Xing, Javier Flores Assad, and Meghan Perez) for their close collaboration and help to improve our models.

%% file: sections/appendix.tex
\newpage
\appendix
\section{Appendix}
\label{sec:appendix}

\subsection{Agentic Evaluation Confidence Intervals}
\label{subsec:eval_filtering}

\begin{figure}[b!]
    \centering
    \includegraphics[width=0.95\textwidth]{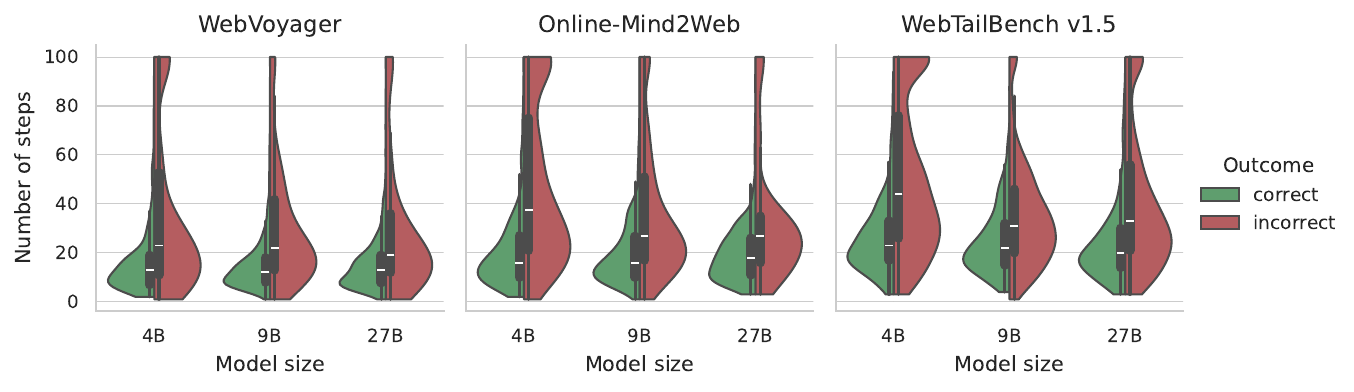}
    \caption{\small Distribution of the number of steps split by outcome, resolved by
    model size (split violins: left/green = correct, right/red = incorrect). The same
    pattern holds at every size---successful trajectories are consistently shorter while
    failures concentrate at longer lengths and at the $100$-step cap---and the failure
    distributions shift toward the cap most for the smaller models.}
    \label{fig:analysis_steps_by_outcome_by_model}
\end{figure}

For each \fara model, all agentic results in this report are averaged over three independent evaluation runs.
\tabref{tab:task_solving_evals_variance} reports the success rates
with 95\% confidence intervals.
The intervals are small relative to
the gaps between model sizes, confirming that the scaling trend in
\tabref{tab:task_solving_evals} is not an artifact of run-to-run noise.

\input{Tables/main_benchmark_table_with_variance}

\subsection{Additional Analysis}
\label{subsec:additional_analysis}

This section collects supporting analyses of agent behavior on the three agentic
benchmarks (WebVoyager, Online-Mind2Web, and \farabench~v1.5).
Throughout, the number of steps is the count of agent actions in a trajectory, and trajectories are capped at $100$ steps.

\myparagraph{Number of steps versus success.}
\figref{fig:analysis_steps_by_outcome_by_model} relates trajectory length to success.
Very short
trajectories tend to fail (the agent answers prematurely), success peaks at a small
number of steps, and the success rate then declines steadily as trajectories grow
longer.
Equivalently, successful trajectories are markedly shorter than failed ones,
which often run until the step budget is exhausted.
This holds across model sizes as well.

\begin{figure}[t]
    \centering
    \includegraphics[width=0.85\textwidth]{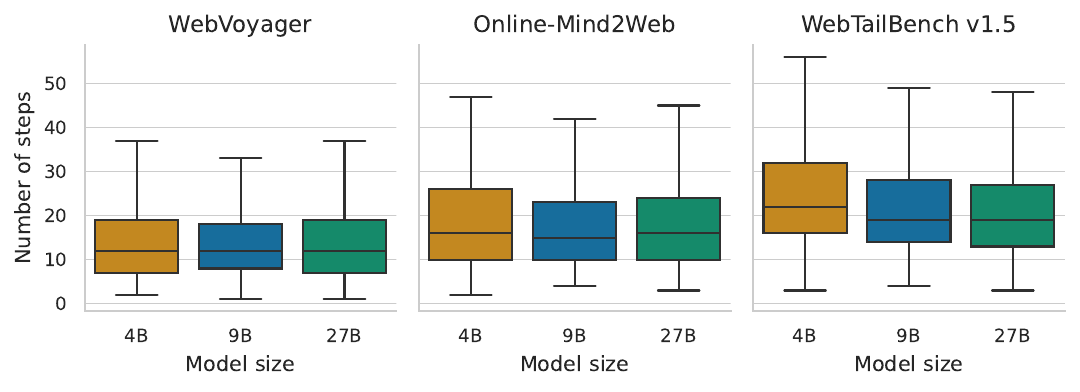}
    \caption{\small Number of steps for \emph{successful} trajectories on the examples
    solved by all three model sizes. Conditioned on success, the distributions are
    nearly identical across sizes, indicating smaller models are not inherently less
    step-efficient.}
    \label{fig:analysis_steps_cond_success}
\end{figure}

\myparagraph{Efficiency conditioned on success.}
The larger mean step counts of smaller models above are largely a consequence of their higher failure rate, not lower efficiency.
\figref{fig:analysis_steps_cond_success} shows the distribution of number of steps for \emph{successful} trajectories on examples solved by all three sizes: conditioned on success, the trajectory-length distributions are very similar across sizes (\farabench~v1.5\ shows only a small residual gap for the 4B model).

\begin{figure}[t]
    \centering
    \includegraphics[width=0.85\textwidth]{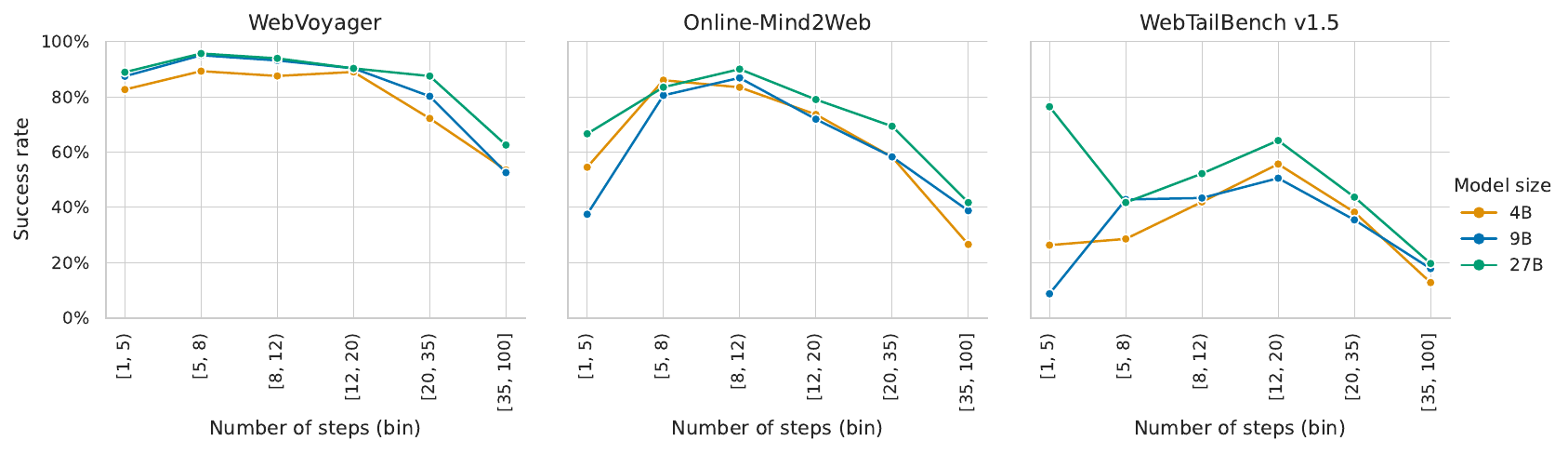}
    \caption{\small Success rate as a function of trajectory length, broken out by model size. The mapping is largely independent of size in the bulk of the step range; larger models retain a higher success rate only in the long tail.}
    \label{fig:analysis_success_vs_steps_by_model}
\end{figure}

\begin{figure}[H]
    \centering
    \includegraphics[width=0.95\textwidth]{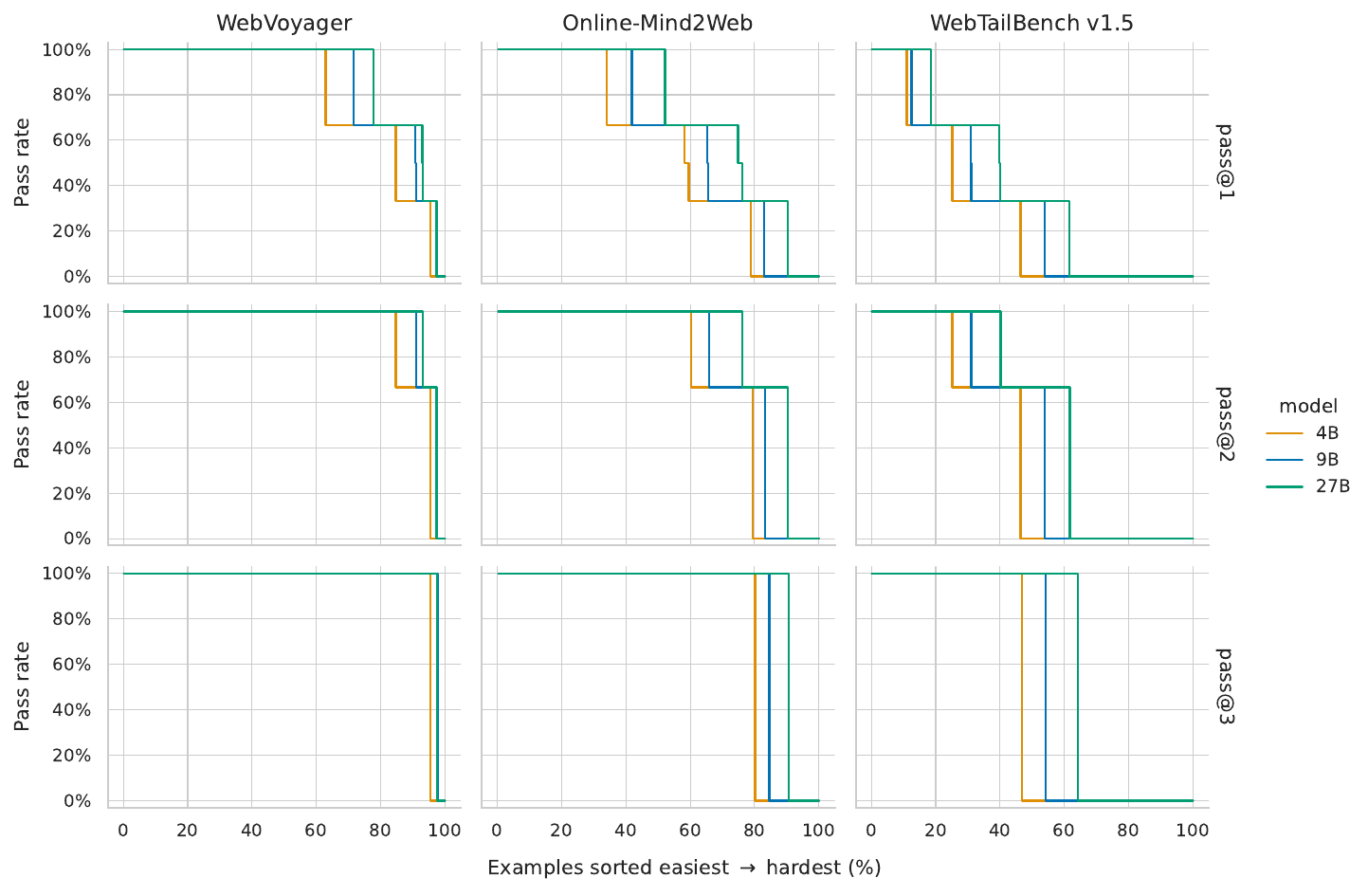}
    \caption{\small Per-example pass@$k$ ($k=1,2,3$) with examples sorted from easiest
    to hardest. The right tail (pass rate $\to 0$) is the set of hardest examples;
    larger models solve more of it, and additional attempts (pass@2, pass@3) recover
    more examples.}
    \label{fig:analysis_passk}
\end{figure}

\begin{figure}[H]
    \centering
    \includegraphics[width=0.95\textwidth]{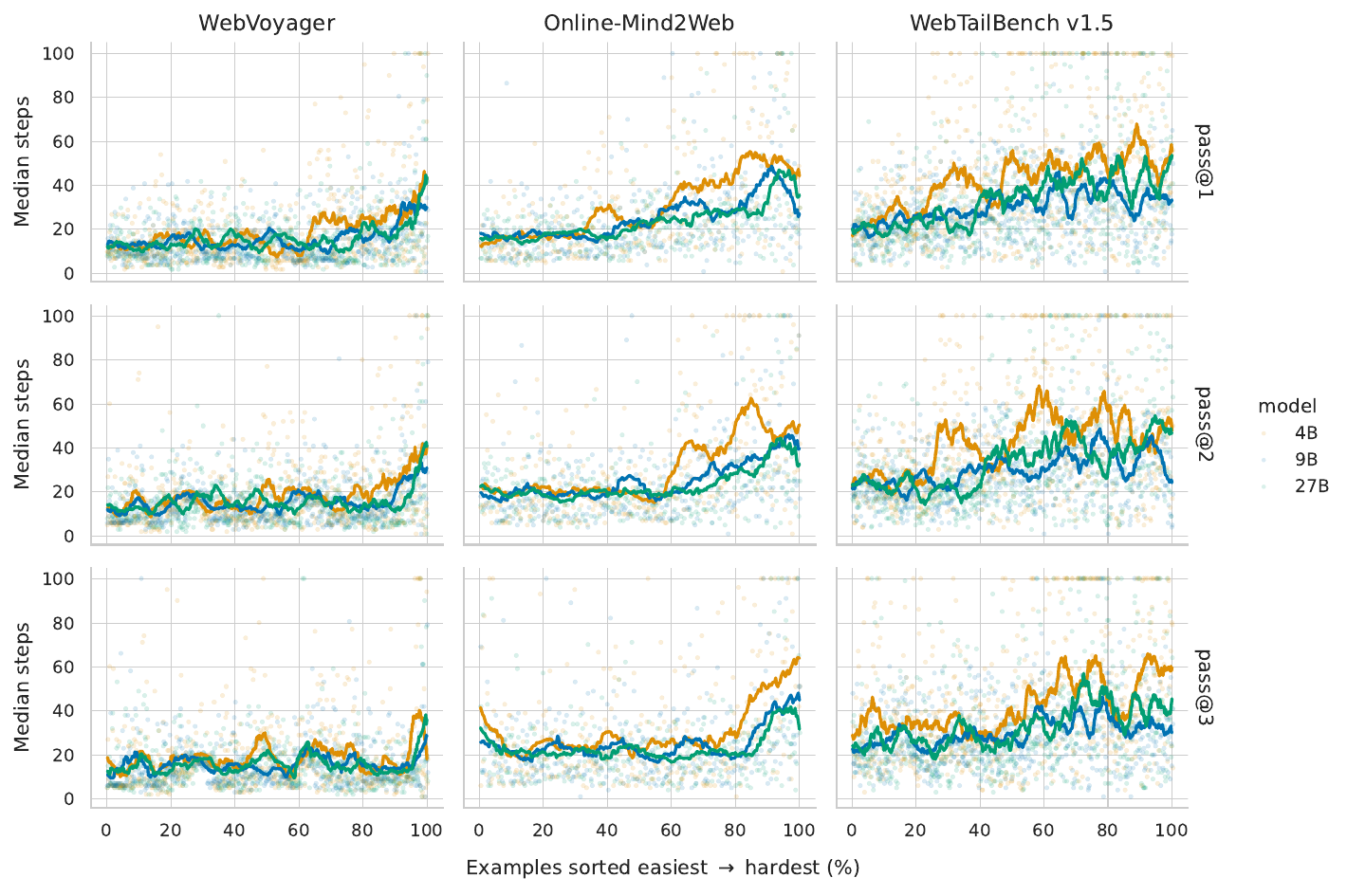}
    \caption{\small Median number of steps per example (faint points) and a rolling
    average (solid lines) along the same easiest-to-hardest ordering as
    \figref{fig:analysis_passk}. Harder examples take more steps.}
    \label{fig:analysis_passk_steps}
\end{figure}

\myparagraph{Per-example difficulty.}
\figref{fig:analysis_passk} sorts examples from easiest to hardest by their estimated pass@$k$ and shows that larger models solve more of the hard tail.
The benchmarks also differ markedly in difficulty (WebVoyager easiest, \farabench~v1.5\ hardest, where even the 27B model leaves a substantial fraction of examples unsolved in three attempts).
\figref{fig:analysis_passk_steps} plots the median number of steps along the same difficulty ordering: harder examples consistently require more steps.

\subsection{Task Proposal on Live Websites}\label{sec:task_proposal_live}

\begin{table}[h]
\centering
\renewcommand{\arraystretch}{1.15}
\footnotesize
\begin{tabular}{y{95}cy{75}y{180}}
Dimension & \# values & Who decides & Values \\
\shline
Site & 100s & sampled & popularity-weighted across 18 domain categories (e-commerce, travel, search-info, forms, food-delivery, social-media, government, finance, healthcare, news, education, real-estate, automotive, dev-tools, media-entertainment, reference-knowledge, productivity, classifieds) \\
Complexity & 6 & sampled & L1 quick action (1--5 steps) through L6 (50--100+ steps) \\
Phrasing & 6 & sampled & well-specified, formal, casual, mobile-fragment, augmented-context, ambiguous \\
Interaction type & 7 & sampled & autonomous, ambiguous, confirmation, preference, credential, correction, multi-step \\
Solvability & 2 & sampled & solvable, impossible \\
\hline
Intent & 15 & LLM chooses & 12 single-site (information-extraction, site-navigation, form-completion, shopping-purchase, comparison-research, booking-reservation, ...) plus compositional (travel-planning, multi-site-procurement, research-then-act) \\
Persona & 8 & LLM chooses & busy-professional,  student-budget, researcher-analyst, tourist-traveler, ... \\
\end{tabular}
\caption{\small Task-proposal dimensions for online environments. Sampled dimensions control coverage; the proposal LLM chooses the remaining two (intent, persona) from a structured menu conditioned on the sampled site.}
\label{tab:task_proposal_dimensions}
\end{table}

\input{Tables/task_proposal_examples}

\subsection{User Simulator Prompts}
\label{sec:user_simulator_prompts}

The user simulator is invoked from two callsites during data generation (Section~\ref{subsec:solver}). When the solver issues an \texttt{ask\_user} tool call at a critical point, we use the critical-point reply prompt below to elicit a plausible user response (personal information, preferences, or approval). Once the solver believes a task is complete, we use the follow-up generation prompt to either accept the result or extend the trajectory with a contextually grounded continuation, which is the principal source of multi-turn training data for \fara.

\begin{promptbox}{Critical-point reply prompt (response to \texttt{ask\_user})}
\begin{lstlisting}[basicstyle=\scriptsize\ttfamily, breaklines=true, columns=fullflexible, frame=none, backgroundcolor=\color{lightblue}, aboveskip=0pt, belowskip=0pt]
You are simulating a user who is interacting with a web browsing agent. The agent has reached a critical point and is requesting information or approval from you to continue. Based on the conversation history and the current screenshot, respond to the agent's request in a way to respond to the request.

Guidelines for providing information:
1. Feel free to create mock user information and make assumptions freely
2. Make sure to answer all questions and provide all necessary details
3. Make sure any assumptions you make are reasonable and consistent with the conversation history
4. If the agent asks for preferences, make a clear choice
5. If the agent needs personal details (name, email, address, etc.), provide realistic fake information

Guidelines for approval requests:
1. If the agent is requesting approval to proceed with an action (e.g., making a purchase, submitting a form), make sure it aligns with the original task
2. If the action seems reasonable and safe, approve it
3. If the action seems risky, unnecessary, or misaligned with the original task, deny it

\end{lstlisting}
\end{promptbox}

\begin{promptbox}{Follow-up task generation prompt (multi-turn extension)}
\begin{lstlisting}[basicstyle=\scriptsize\ttfamily, breaklines=true, columns=fullflexible, frame=none, backgroundcolor=\color{lightblue}, aboveskip=0pt, belowskip=0pt]
You are simulating a user who is interacting with a web browsing agent. Based on the conversation history and the current screenshot, generate a natural, contextually relevant follow-up task.

Generate a realistic follow-up task that:
1. Is naturally related to the conversation history and previous tasks/outputs
2. Could reasonably be asked by a user in this context
3. Leverages the current page state or information visible in the screenshot, but requires navigation beyond the current page
4. Is specific and actionable (not vague like "tell me more")
5. Requires actions (clicking, navigating to new page), beyond just reading the page
6. The follow-up task should be slightly simpler than the original task but within similar complexity
7. Do not ever refer to https:// URLs directly in your feedback to the agent. Instead, refer to the general domain name (i.e, Google Flights) instead of an https URL.

Examples of good follow-up tasks:
- If the previous task was "What is the best pair of noise canceling headphones", a good follow-up: "can you see if this headphone is available on Amazon", a second follow up could be "purchase it"
- If the previous task was "Search arXiv for the latest papers on computer use agents", a good follow-up "search arxiv for latest papers on multimodal LLMs" and a second follow-up could be "find more papers on arxiv from the first author of the first paper"
- If the previous task was "find the next liverpool and boston celtics game times", a good follow-up would be "find the ticket prices for the celtics game" or "tell me what was the previous score of the last liverpool and celtics game"

Criteria:
- Tasks should work without being logged in to any accounts
- Tasks should not require irreversible actions like making a purchase (adding to cart is okay) or creating an account
- No tasks involving downloading or viewing PDFs
- Tasks should be very well specified so that by just reading the task alone one could solve it successfully
- The answer to the task should be at most a few sentences long
- Provide fake user information if needed for a follow up
- Do not hand the agent an exact deep URL. Use the site/brand name, the item or recipe name, or broad search terms instead.
\end{lstlisting}
\end{promptbox}

\subsection{Model Training Details}
\label{sec:training_settings}

\tabref{tab:training_settings} summarizes the hyperparameters used to train the released \fara family.
All three sizes are trained from the corresponding \qwennew backbone with supervised fine-tuning under the same data mix and the same optimization schedule, so that downstream comparisons isolate the effect of model size.

\begin{table}[h]
\centering
\renewcommand{\arraystretch}{1.15}
\footnotesize
\begin{tabular}{y{150}y{280}}
Setting & Value \\
\shline
Optimizer & AdamW ($\beta_1$ = 0.9, $\beta_2$ = 0.95, $\epsilon$ = $10^{-7}$) \\
Peak learning rate & $5 \times 10^{-6}$ \\
LR schedule & Cosine decay with warmup ratio $0.10$ \\
Epochs & 2  \\
Precision & bfloat16 \\
Batch size & 128 \\
Training compute & 32/64 $\times$ NVIDIA B200 (4/8 nodes $\times$ 8 GPUs), $\sim 4$--$5$ days per run \\
\end{tabular}
\caption{\small Training settings for the released \fara\ family. The 4B, 9B, and 27B variants share the optimization schedule and data mix.}
\label{tab:training_settings}
\end{table}

\subsection{Example Trajectories}
\label{sec:example_trajectories}

We illustrate two trajectories produced by \faranine\ on the \farabench~v1.5 held-out set, both rated by the FaraBench v1.5 ensemble judge (GPT-5.2 + o4-mini). The first trajectory succeeds on both the process rubric and the outcome judgement. The second satisfies the process rubric ($\geq 0.8$) but is marked an outcome failure by the judge which motivates reporting both metrics in \tabref{tab:task_solving_evals}.

\subsubsection{Process \emph{and} outcome success: \texttt{hotels\_head\_orbitz\_8}}\label{sec:traj_orbitz}

\noindent\textbf{Task.} \emph{I'd like to reserve a room at Legoland Hotel in Carlsbad, California using orbitz.com checking in October 30 - November 2. If the hotel doesn't take reservations for that date or there are no available rooms for that time, please indicate that in your answer.}

\noindent\textbf{Scores.} Process rubric $= 1.00$; outcome judgement $=$ \textbf{success}.

\begin{promptbox}{Step 1 reasoning (verbatim)}
\begin{lstlisting}[basicstyle=\scriptsize\ttfamily, breaklines=true, columns=fullflexible, frame=none, backgroundcolor=\color{lightblue}, aboveskip=0pt, belowskip=0pt]
I'm starting from the Bing homepage and the task specifically names orbitz.com as the site to use. The most direct and reliable next step is to navigate straight to Orbitz rather than using Bing results, because the instructions say to complete the task within the named website's UI and to use direct navigation when the base URL is known. After loading Orbitz, I'll inspect the homepage for the hotel search form and date controls.

action: visit_url(url="https://www.orbitz.com/")
\end{lstlisting}
\end{promptbox}

\noindent\textbf{Full action sequence (16 steps).}
{\footnotesize
\begin{enumerate}[leftmargin=2em, itemsep=0pt, topsep=2pt]
    \item \texttt{visit\_url}(\textit{https://www.orbitz.com/})
    \item \texttt{left\_click}(206, 221) --- focus hotel-search input
    \item \texttt{type}(``LEGOLAND Hotel Carlsbad'')
    \item \texttt{left\_click}(469, 222) --- pick suggestion
    \item \texttt{left\_click}(769, 356) --- open date picker
    \item \texttt{left\_click}(767, 357) --- advance calendar
    \item \texttt{left\_click}(767, 357) --- advance calendar
    \item \texttt{left\_click}(767, 357) --- advance calendar
    \item \texttt{left\_click}(767, 357) --- advance calendar to October
    \item \texttt{left\_click}(510, 654) --- select check-in 10/30
    \item \texttt{left\_click}(628, 463) --- select check-out 11/2
    \item \texttt{left\_click}(867, 222) --- submit search
    \item \texttt{left\_click}(160, 567) --- focus filter
    \item \texttt{type}(``LEGOLAND Hotel Carlsbad'') --- refine to matching property
    \item \texttt{left\_click}(185, 656) --- inspect property card
    \item \texttt{terminate}(answer=``Orbitz shows `We are sold out' for Legoland Hotel California for Oct 30--Nov 2.\ldots'')
\end{enumerate}}

\begin{figure}[H]
    \centering
    \begin{subfigure}[t]{0.49\linewidth}
        \includegraphics[width=\linewidth]{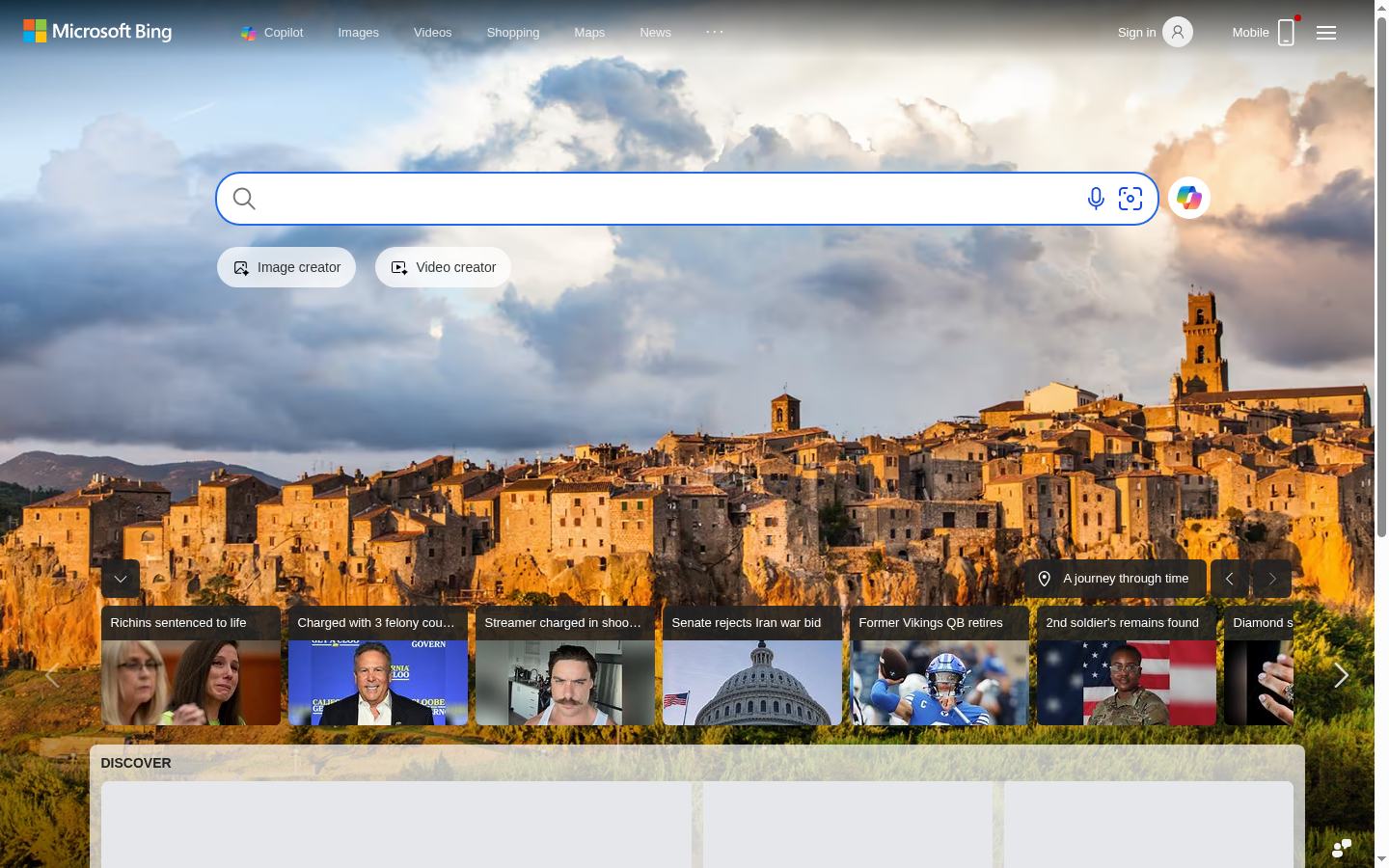}
        \caption{Step 1: starting on Bing.}
    \end{subfigure}\hfill
    \begin{subfigure}[t]{0.49\linewidth}
        \includegraphics[width=\linewidth]{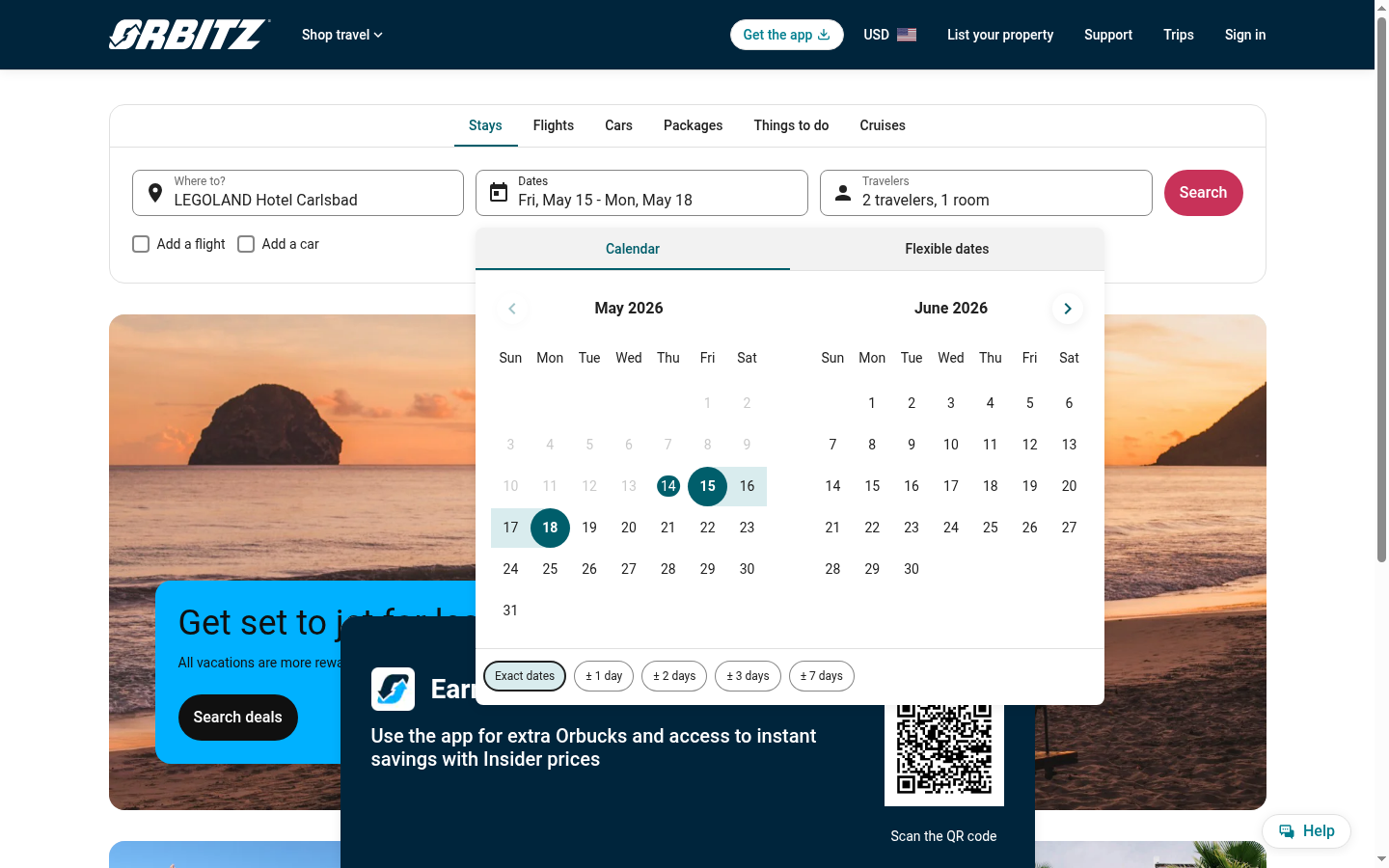}
        \caption{Step 5: Orbitz date picker.}
    \end{subfigure}\\[2pt]
    \begin{subfigure}[t]{0.49\linewidth}
        \includegraphics[width=\linewidth]{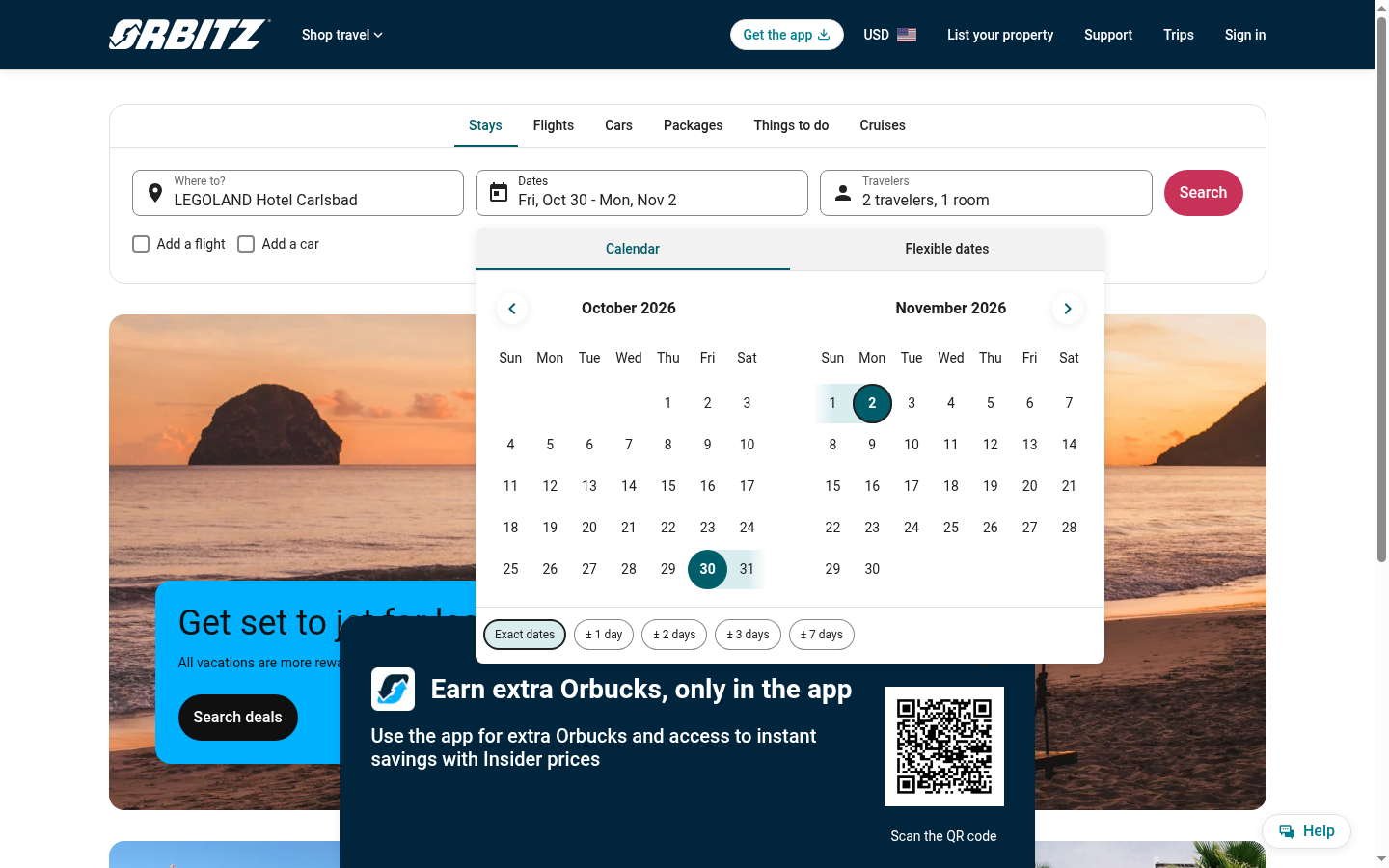}
        \caption{Step 12: results refined to the property.}
    \end{subfigure}\hfill
    \begin{subfigure}[t]{0.49\linewidth}
        \includegraphics[width=\linewidth]{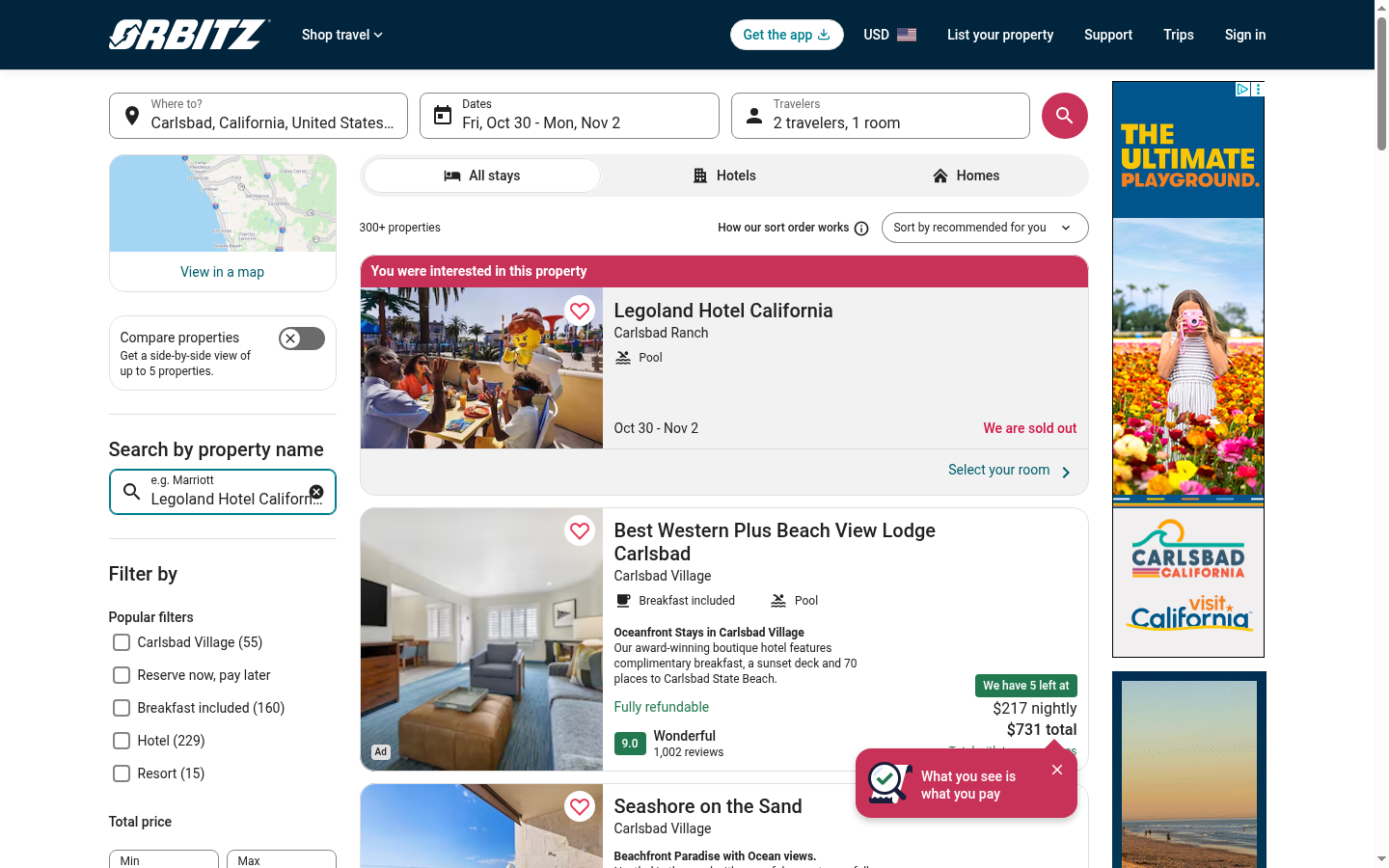}
        \caption{Step 16: ``We are sold out'' visible on the property card.}
    \end{subfigure}
    \caption{\small Selected screenshots from \texttt{hotels\_head\_orbitz\_8}. The agent visited the requested site, entered the hotel name and dates, and read the sold-out banner directly off the property card before terminating.}
    \label{fig:traj_orbitz}
\end{figure}

\subsubsection{Process-only success: \texttt{jobs\_apply\_apply\_1219}}\label{sec:traj_jobs}

\noindent\textbf{Task.} \emph{Help me apply for a chemistry research scientist position in Madison, WI found on LinkedIn and output five listings that would have me be an early application according to the listings. If there are less than five such listings, output them all.}

\noindent\textbf{Scores.} Process rubric $= 0.94$ (success); outcome judgement $=$ \textbf{failure}. The judge marks the outcome a failure because one of the five returned listings (Scientist I, Quality Control --- Arrowhead Pharmaceuticals) is located in Verona, WI rather than Madison, WI, violating the explicit location constraint despite the rest of the deliverable being correct.

\begin{promptbox}{Step 1 reasoning (verbatim)}
\begin{lstlisting}[basicstyle=\scriptsize\ttfamily, breaklines=true, columns=fullflexible, frame=none, backgroundcolor=\color{lightblue}, aboveskip=0pt, belowskip=0pt]
The current screen appears to be a blank browser page with no visible content loaded yet. To efficiently locate relevant LinkedIn job postings, I will navigate directly to a LinkedIn Jobs search URL that filters for chemistry research scientist roles in Madison, Wisconsin and sorts by earliest applying. This uses LinkedIn's own interface rather than an external search engine, which matches the task requirement.

action: visit_url(url="https://www.linkedin.com/jobs/search/?keywords=chemistry%
\end{lstlisting}
\end{promptbox}

\noindent\textbf{Full action sequence (8 steps).}
{\footnotesize
\begin{enumerate}[leftmargin=2em, itemsep=0pt, topsep=2pt]
    \item \texttt{visit\_url}(\textit{LinkedIn Jobs search for ``chemistry research scientist'' in Madison, WI, sorted by date})
    \item \texttt{left\_click}(596, 290) --- dismiss the sign-in popup
    \item \texttt{pause\_and\_memorize\_fact}(``Confirmed early-application listings so far: 1) Sr Scientist I, \ldots'')
    \item \texttt{scroll}($-622$) --- scroll down through results
    \item \texttt{mouse\_move}(297, 607)
    \item \texttt{scroll}($-544$) --- continue scrolling
    \item \texttt{pause\_and\_memorize\_fact}(``Additional confirmed early-application listings: 3) Chemist I (JR10010351), \ldots'')
    \item \texttt{terminate}(answer=``Five postings with the `Be an early applicant' label: \ldots'')
\end{enumerate}}

\begin{figure}[H]
    \centering
    \begin{subfigure}[t]{0.49\linewidth}
        \includegraphics[width=\linewidth]{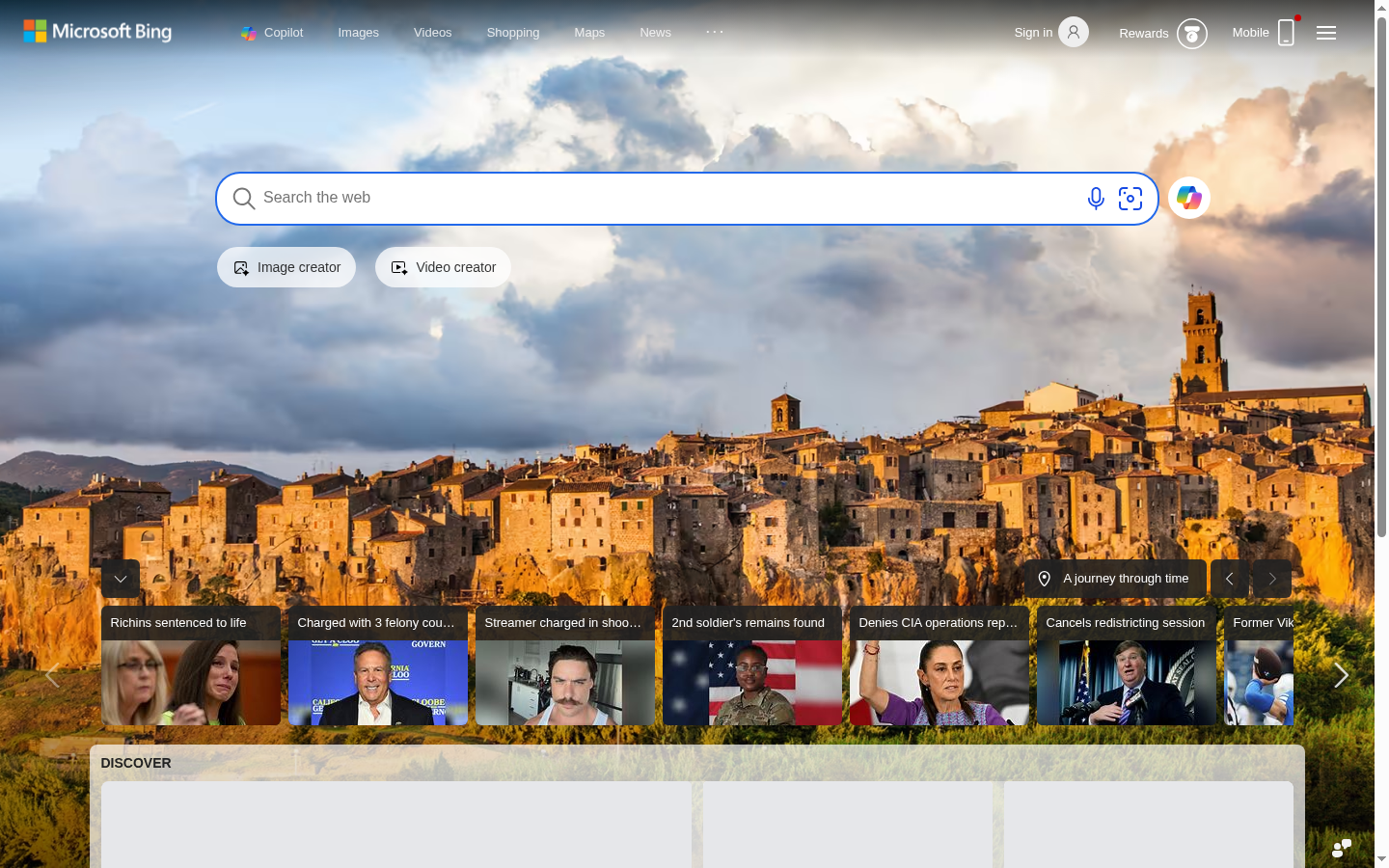}
        \caption{Step 1: blank browser, prior to LinkedIn URL navigation.}
    \end{subfigure}\hfill
    \begin{subfigure}[t]{0.49\linewidth}
        \includegraphics[width=\linewidth]{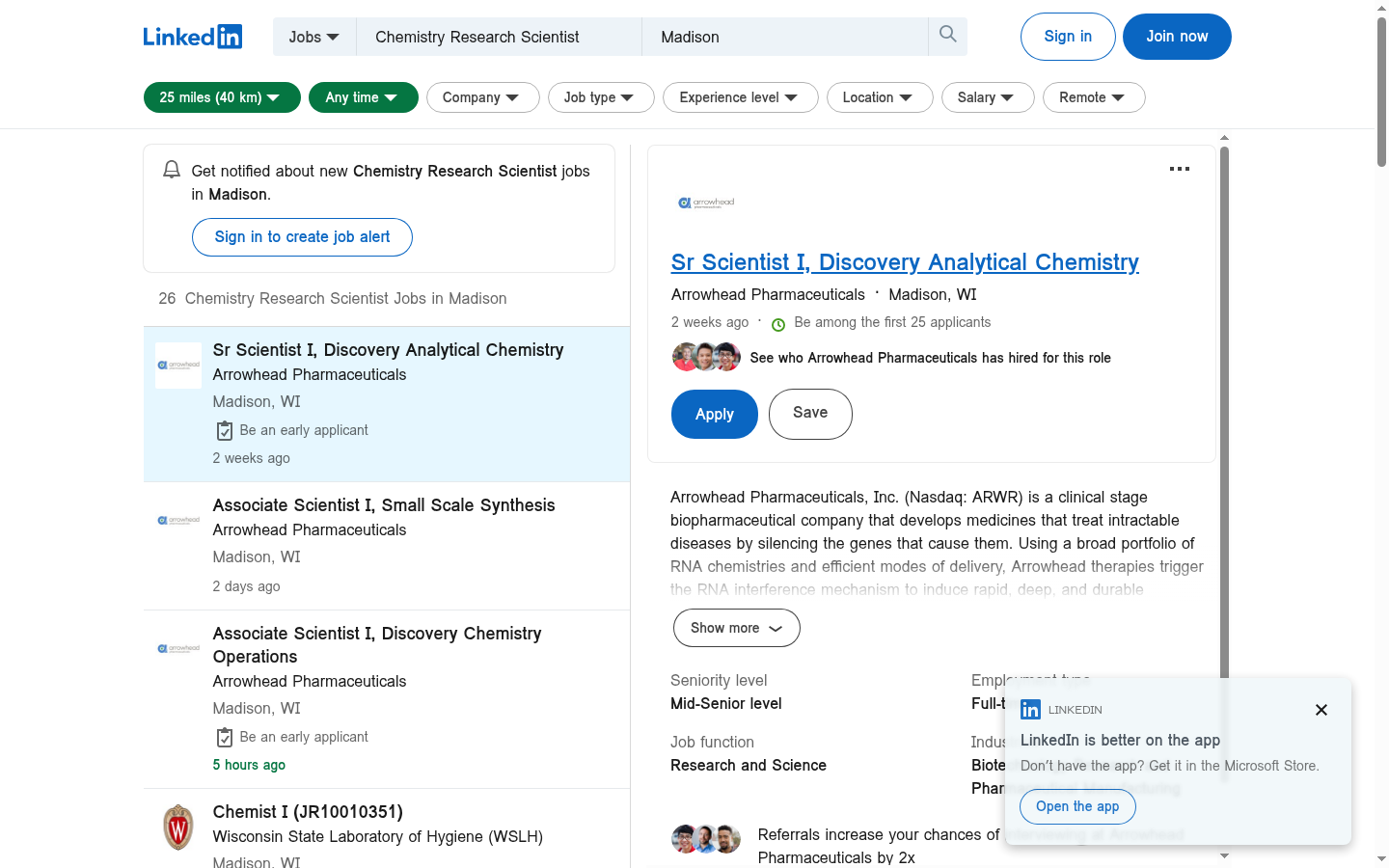}
        \caption{Step 4: LinkedIn results, popup dismissed, first listings visible.}
    \end{subfigure}\\[2pt]
    \begin{subfigure}[t]{0.49\linewidth}
        \includegraphics[width=\linewidth]{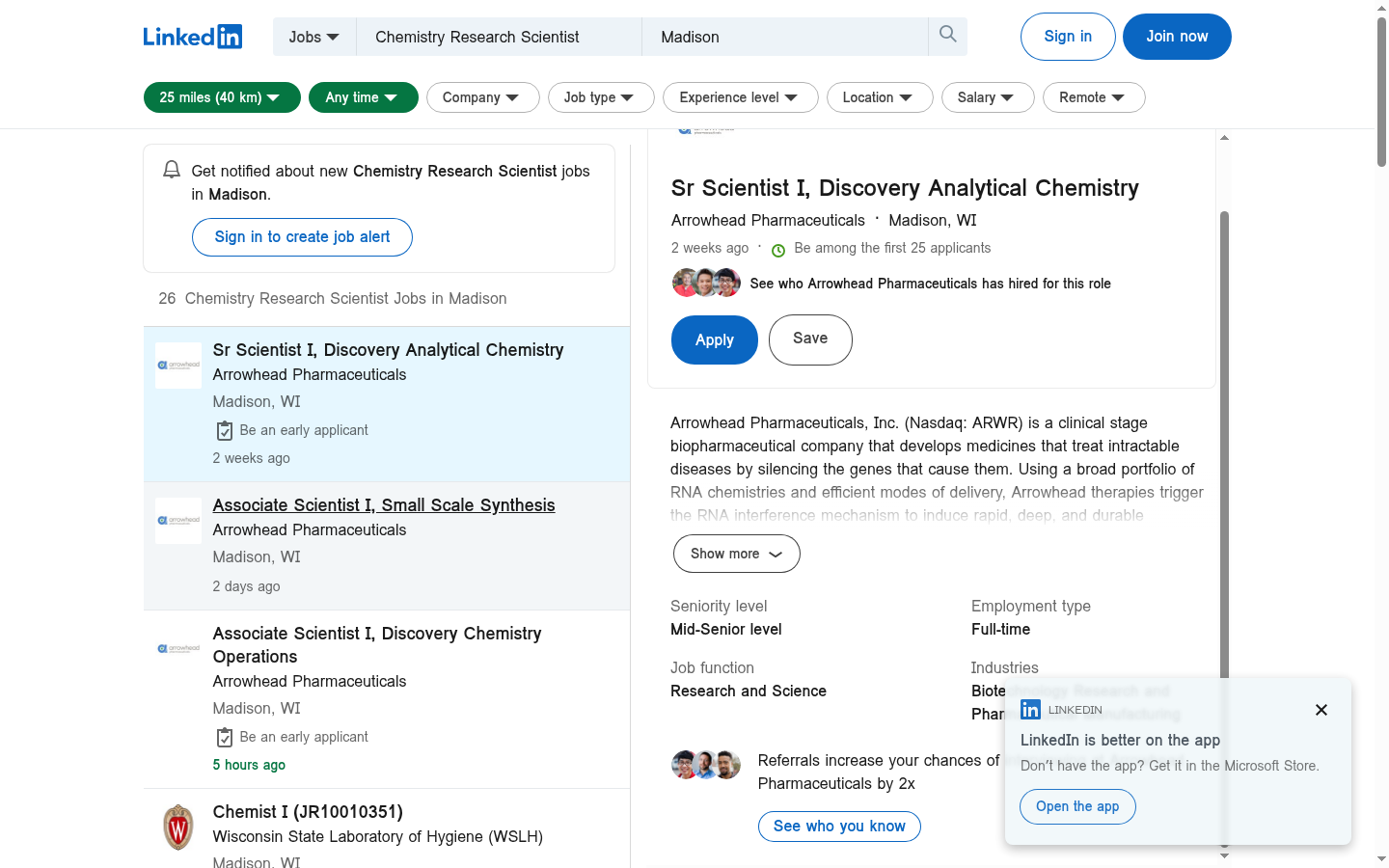}
        \caption{Step 6: mid-scroll through the results, additional early-applicant tags visible.}
    \end{subfigure}\hfill
    \begin{subfigure}[t]{0.49\linewidth}
        \includegraphics[width=\linewidth]{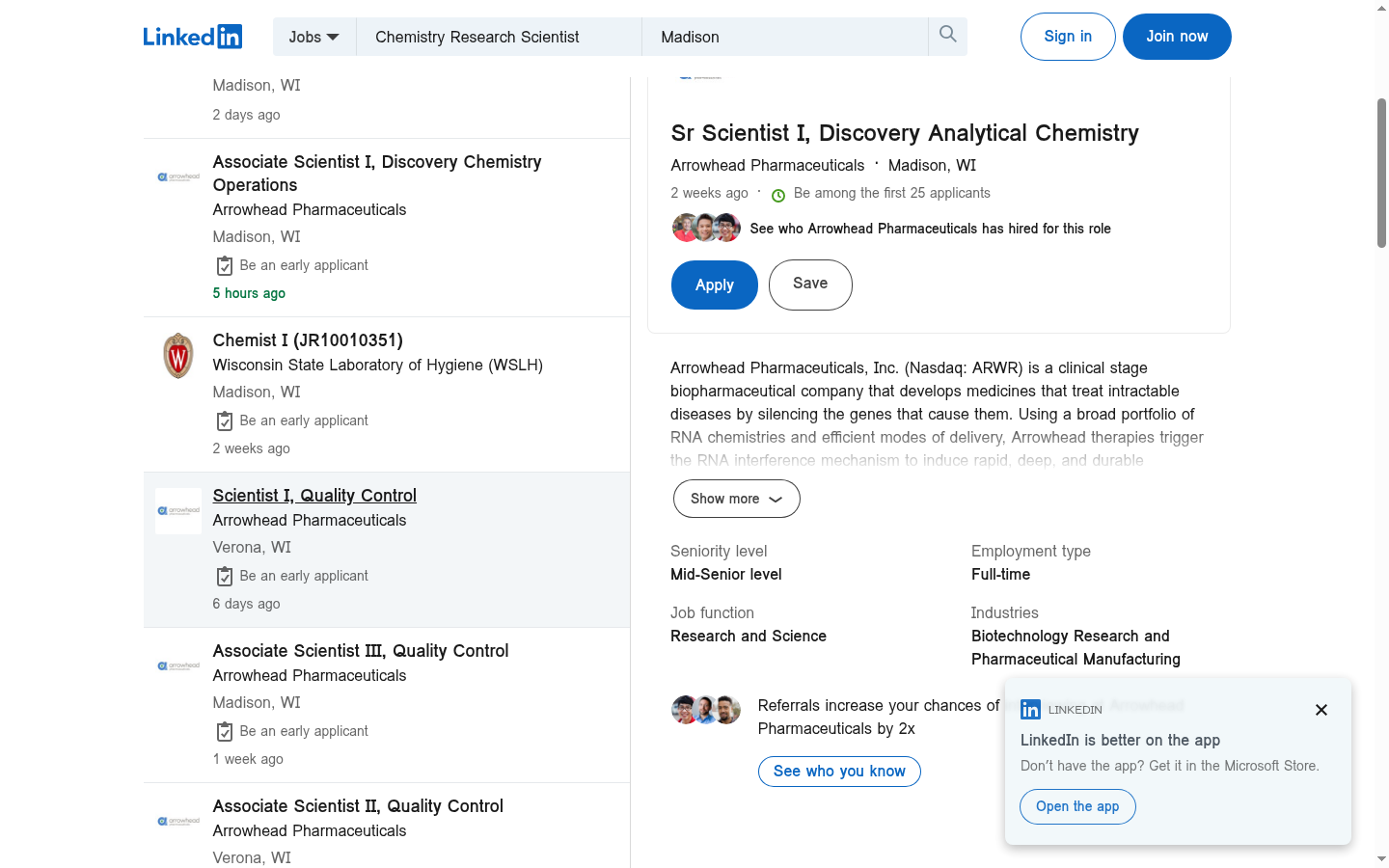}
        \caption{Step 8: final state before \texttt{terminate}.}
    \end{subfigure}
    \caption{\small Selected screenshots from \texttt{jobs\_apply\_apply\_1219}. The trajectory follows a clean process: navigate to the filtered LinkedIn search, dismiss the modal, collect candidates via \texttt{pause\_and\_memorize\_fact} while scrolling. The outcome judge nevertheless marks this a failure because one returned listing is located in Verona, WI rather than the Madison, WI specified by the task.}
    \label{fig:traj_jobs}
\end{figure}

\subsection{Failure Analysis}\label{sec:failure_analysis}

Expanding on the failure analysis from \citet{rosset2026art}, we provide an error taxonomy of common errors we observed in our trajectories in developing our data generation and evaluation pipeline in Table~\ref{tab:error-taxonomy}. We use this error taxonomy along with the trajectory history and rich evaluated rubric of the Universal Verifier to determine all points of failure within the trajectory. We provide a distribution of failures of WebVoyager, Online-Mind2Web, and WebTailBench in Tables~\ref{tab:wv_failure_analysis},~\ref{tab:om2w_failure_analysis}, and~\ref{tab:wtb_failure_analysis}, respectively.

\input{Tables/error_taxonomy.tex}

\input{Tables/wv_failure_analysis}

\input{Tables/om2w_failure_analysis}

\input{Tables/wtb_failure_anlaysis.tex}

\clearpage

\subsection{Grounding Examples}\label{sec:grounding-examples}

\begin{figure}[h!]
    \centering
    \includegraphics[width=0.8\columnwidth]{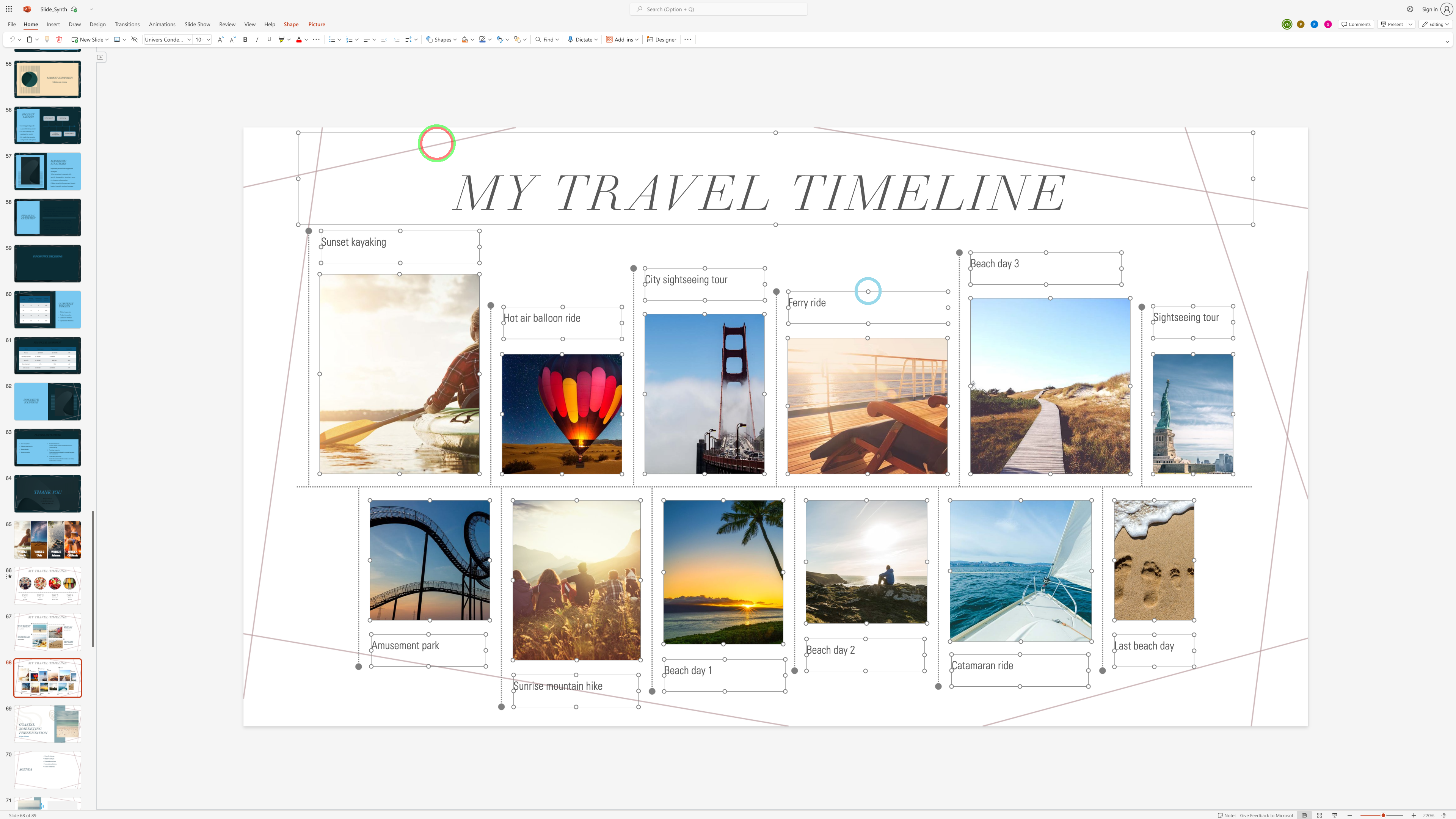}
    \caption{Task: \textit{Please generate the next move according to the UI screenshot and instruction. 
    Instruction: Select the handle located at the top of the text box containing the text "Ferry ride."}
    The concentric red and lime green circle denotes where the Jedi dataset \textbf{incorrectly} says the object of interest is located, the blue circle is the \textbf{correct} location of the object of interest. This is an \textbf{inaccurate} error.
}
    \label{fig:jedi-ppt-error}
\end{figure}

\begin{figure}[h!]
    \centering
    \includegraphics[width=0.8\columnwidth]{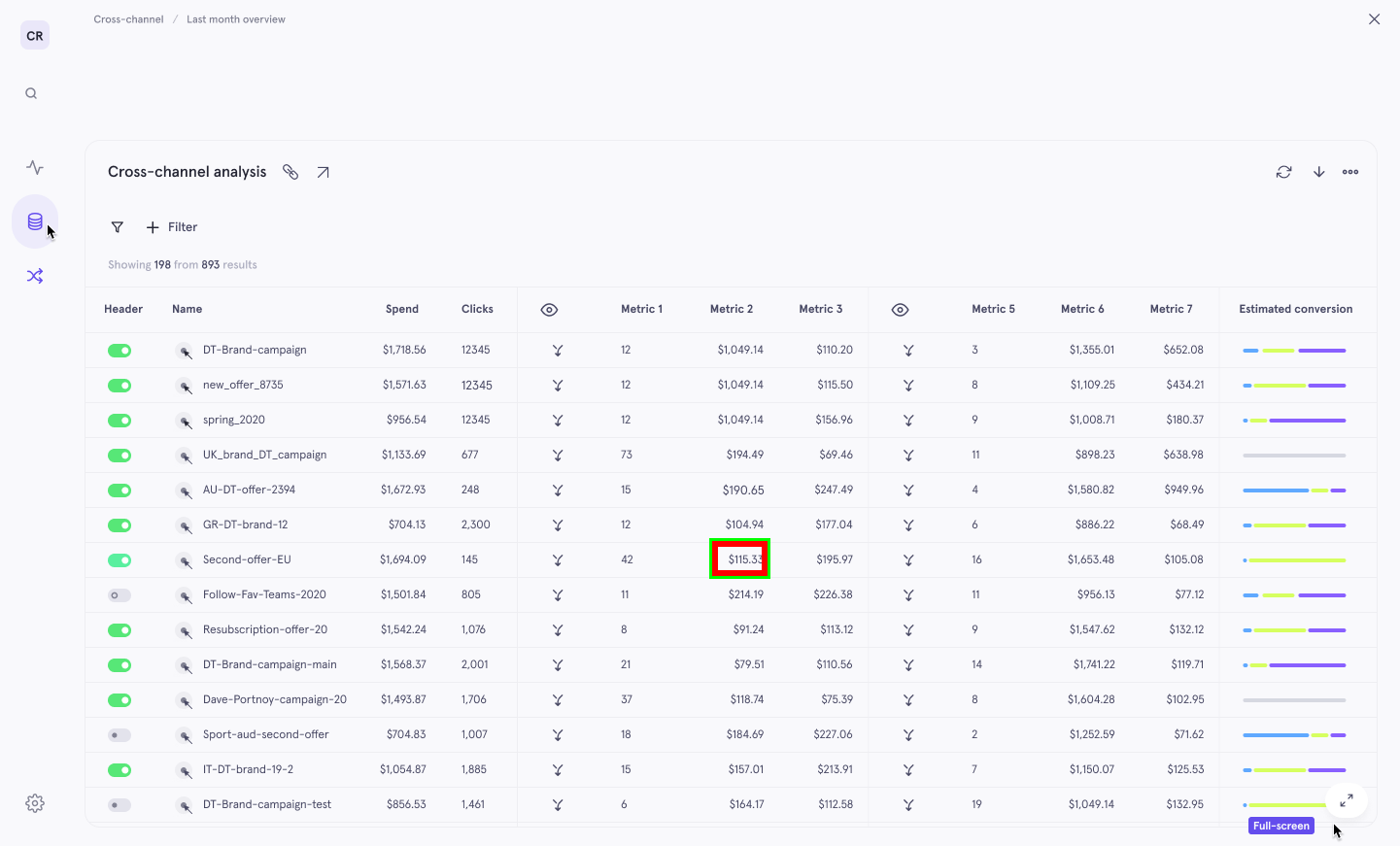}
    \caption{Task: \textit{The Text field within a table cell.'s intended function:
The primary function of this element is to display a specific metric value, likely representing a financial figure or performance metric. Users can view this value to assess or compare it with other metrics in the table.} This is an error because the task is \textbf{not unique} for the instruction's description of the table cell.
}
    \label{fig:jedi-cell}
\end{figure}

The quality of the open source datasets are not perfect. For instance, some segments in Jedi have false positive rates of 7/10, where a false positive is defined as the Jedi dataset says an object described in the instruction exists at a coordinate in the image, but in reality it exists elsewhere. We illustrate some of the poor examples in Figures~\ref{fig:jedi-ppt-error} and \ref{fig:jedi-cell}.

\subsection{Form-Filling Dataset}
\label{sec:form_filling_dataset}

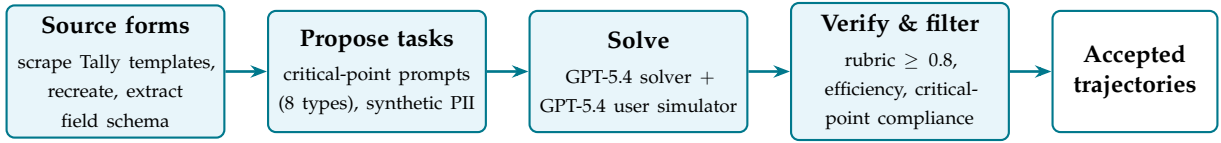
\begin{figure}[H]
\centering
\begin{tikzpicture}[
  font=\footnotesize,
  stage/.style={draw=blue, fill=lightblue, line width=0.8pt, rounded corners=3pt,
                text width=2.6cm, align=center, inner sep=4pt, minimum height=1.6cm},
  outbox/.style={draw=blue, fill=white, line width=0.8pt, rounded corners=3pt,
              text width=1.9cm, align=center, inner sep=4pt, minimum height=1.6cm},
  arr/.style={-{Stealth[length=2.2mm]}, blue, line width=1pt}
]
\node[stage] (a) {\textbf{Source forms}\\[2pt]\scriptsize scrape Tally templates, recreate, extract field schema};
\node[stage, right=0.55cm of a] (b) {\textbf{Propose tasks}\\[2pt]\scriptsize critical-point prompts (8 types), synthetic PII};
\node[stage, right=0.55cm of b] (c) {\textbf{Solve}\\[2pt]\scriptsize \fivefour\ solver $+$ \fivefour\ user simulator};
\node[stage, right=0.55cm of c] (d) {\textbf{Verify \& filter}\\[2pt]\scriptsize rubric $\geq 0.8$, efficiency, critical-point compliance};
\node[outbox, right=0.55cm of d] (e) {\textbf{Accepted trajectories}};
\draw[arr] (a) -- (b);
\draw[arr] (b) -- (c);
\draw[arr] (c) -- (d);
\draw[arr] (d) -- (e);
\end{tikzpicture}
\caption{\small Form-filling data pipeline. Tasks are seeded from real Tally forms
and critical-point prompts, solved by the \fivefour\ solver paired with a \fivefour\
user simulator that supplies missing information or permission on request, and
filtered by the same verifiers as \faragen\ (\secref{subsec:verifiers}).}
\label{fig:form_filling_pipeline}
\end{figure}

A large part of the form-filling and user-interaction slice of the training mix
(\figref{fig:datagen_growth_and_mix}) comes from agent trajectories that fill real
\emph{Tally}\footnote{\url{https://tally.so}} forms. Every task instantiates one of
the eight critical-point types in \tabref{tab:critical_point_types}, so the data
teaches the model to complete forms while deferring to the user when information is
missing, the task is underspecified, or an irreversible submission is unauthorized.
\figref{fig:form_filling_pipeline} shows the pipeline: we seed tasks from real Tally
forms and critical-point prompts with synthetic user information, solve them with
the \fivefour\ solver and a \fivefour\ user simulator that supplies missing details
or permission on request, and keep only trajectories that pass the \faragen\
verifiers (\secref{subsec:verifiers}). \tabref{tab:form_filling_dataset} summarizes
the result.

\input{Tables/form_filling_dataset}

\myparagraph{Why we added it.} Form filling exercises behaviors---entering PII and
taking irreversible actions---that cannot be collected safely on the open web. On the held-out split where the user grants
permission and provides all information (so the agent should simply fill and
submit), \model\ scores \textbf{0\%}: it refuses nearly every form as a privacy
risk \emph{even under the new critical-point system prompt}, which it ignores. A
model trained on this data reaches roughly \textbf{80\%} on the same split.

\begin{figure}[t]
\centering
\begin{subfigure}[t]{0.32\linewidth}
\includegraphics[width=\linewidth]{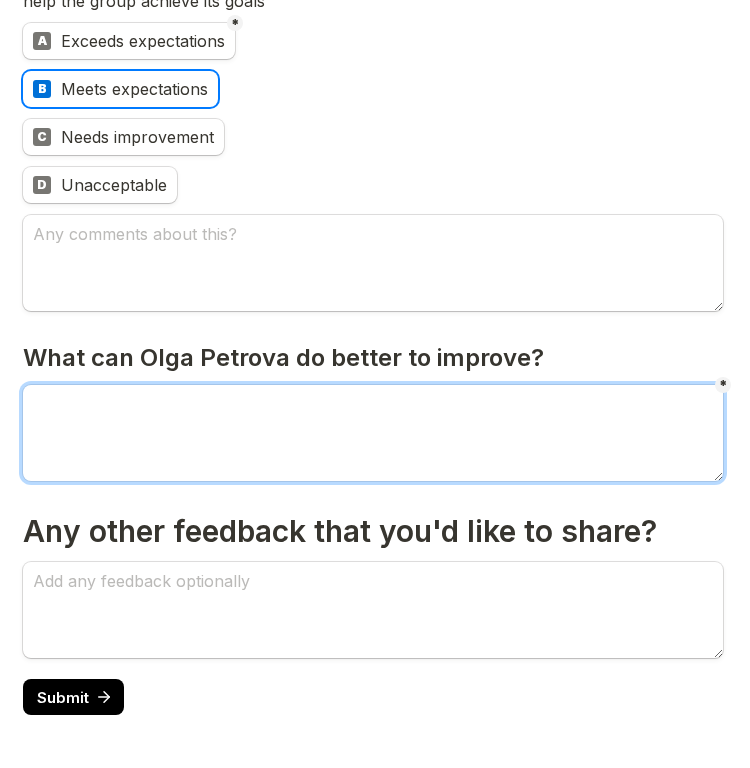}
\caption{The agent applies the ratings the user specified, but the required open-ended field has no provided value.}
\end{subfigure}\hfill
\begin{subfigure}[t]{0.32\linewidth}
\includegraphics[width=\linewidth]{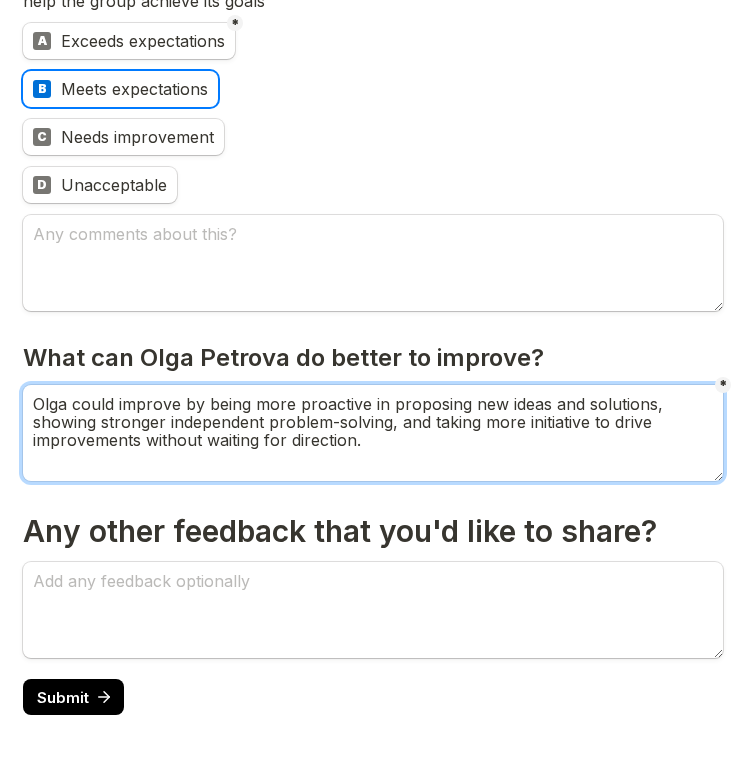}
\caption{Rather than fabricating it, the agent asks the user and enters the supplied comment verbatim.}
\end{subfigure}\hfill
\begin{subfigure}[t]{0.32\linewidth}
\includegraphics[width=\linewidth]{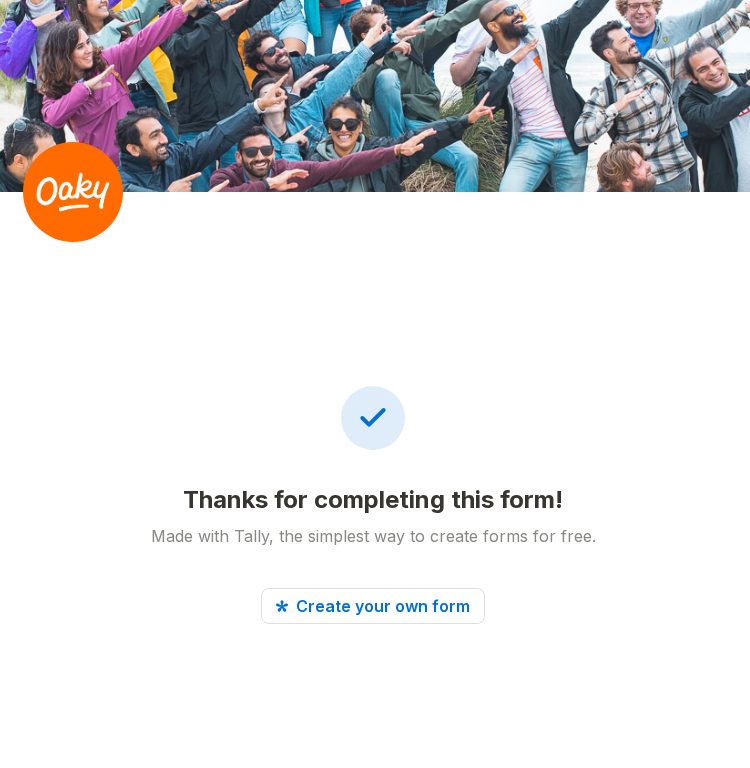}
\caption{Since permission was already granted, the agent submits without asking again and reaches the confirmation page.}
\end{subfigure}
\caption{\small A form-filling trajectory where the user authorizes submission but
omits a required field. The agent fills every value it was given, asks for the one
missing required answer instead of fabricating it, and then submits without a
redundant permission request.}
\label{fig:form_filling_example}
\end{figure}

\myparagraph{Worked example.} \figref{fig:form_filling_example} traces a
representative trajectory where the user authorizes submission but does not provide
a value for one required field.

\noindent\textbf{Task.} \emph{Please submit [Tally form] for Olga Petrova. Make it
partly anonymous (team lead can know). I'm Amanda Foster. Rate everything as Meets
expectations, except Innovation as Unacceptable.}

\noindent The agent enters the reviewer and reviewee, sets partial anonymity, and
applies the requested ratings. It then reaches a required open-ended field,
\emph{``What can Olga Petrova do better to improve?''}, for which the user gave no
value. Rather than inventing feedback about a real person, the agent asks:

\begin{promptbox}{Agent \texttt{ask\_user\_question} (verbatim)}
The form has a required question: ``What can Olga Petrova do better to improve?''
What exact comment would you like me to enter there? If you want, you can give me a
short sentence or bullet points, and I'll submit the form after entering it.
\end{promptbox}

\noindent The simulated user supplies the comment, the agent enters it verbatim, and
because permission to submit was already granted it submits without asking again,
reaching the confirmation page.

%% file: Tables/main_benchmark_table_with_variance.tex
\begin{table}[h]
\centering
\small
\begin{tabular}{lccc}
\textbf{Model} & \textbf{WebVoyager} & \textbf{Online-Mind2Web} & \textbf{\farabench~v1.5} \\
\midrule
\farafour & $80.8 \pm 2.3$ & $57.3 \pm 4.3$ & $27.4 \pm 2.8$ \\
\faranine & $86.6 \pm 2.0$ & $63.4 \pm 4.3$ & $32.3 \pm 2.8$ \\
\faratwentyseven & $89.3 \pm 1.8$ & $72.3 \pm 3.8$ & $40.2 \pm 3.0$ \\
\end{tabular}
\caption{Success rate (\%) of the \fara family with 95\%
confidence intervals, averaged over the three independent runs. For \farabench~v1.5
we report Outcome Success.}
\label{tab:task_solving_evals_variance}
\end{table}

%% file: Tables/task_proposal_examples.tex
\begin{table}[t!]
\centering
\renewcommand{\arraystretch}{1.25}
\footnotesize
\resizebox{\textwidth}{!}{%
\begin{tabular}{y{60}y{85}y{260}}
Site & Sampled tags & Instruction \\
\shline
OpenStax & education, L2, casual, correction & i need a free biology textbook pdf on OpenStax \\
GCP Docs & dev-tools, L1, casual, autonomous & just need the GCP Docs page for Cloud Billing \\
Simply Recipes & food-delivery, L2, casual, autonomous & find the Simply Recipes sweepstakes page if they have one \\
Movoto & real-estate, L2, mobile-fragment, preference & new listings near miami beach with pool maybe condo or house \\
NYTimes & news, L4, formal, confirmation & Find 3 recent NYT Wirecutter carry-on suitcase picks and compare price, weight, expandable option, and warranty. Put the best value in the cart but stop before checkout for my approval. \\
Hostelworld & travel, L5, casual, autonomous & find me the best 4 nights in Lisbon next month on Hostelworld for 1 person, mix of dorm and private options, then compare the top 5 by total price, rating, location, and free cancellation \\
Staples & e-commerce, L5, ambiguous, autonomous & need shipping and shredding info from Staples Business Center, what services are there and any obvious limits or notes \\
Carfax & automotive, L6, mobile-fragment, credential & carfax sell my car start cash offer for my vin ask me details when needed \\
UpToDate & healthcare, L6, augmented-context, multi-step & From scratch, use UpToDate to build a complete public-facing feature matrix: specialties, calculators, patient ed, drug tools, CME, apps, EHR, support. We'll refine. \\
\end{tabular}%
}
\caption{\small A few example tasks chosen to span all seven dimensions of variation (\tabref{tab:task_proposal_dimensions}). Sampled tags list (domain, complexity, phrasing, interaction type) for each instruction; the proposal LLM additionally picked an intent and persona conditioned on the site.}
\label{tab:task_proposal_examples}
\end{table}

%% file: Tables/error_taxonomy.tex
\definecolor{catbg}{gray}{0.90}

\newcommand{\catrow}[1]{%
  \rowcolor{catbg} \multicolumn{2}{@{}l}{\textbf{#1}} \\}
\begin{table}[h!]
\centering
\footnotesize
\setlength{\tabcolsep}{4pt}
\renewcommand{\arraystretch}{1.1}
\begin{tabular}{@{} >{\raggedright\arraybackslash}p{0.22\textwidth} >{\raggedright\arraybackslash}p{0.74\textwidth} @{}}
\toprule
\textbf{Error Type} & \textbf{Description} \\
\midrule

\catrow{1.~Selection}
1.1~~Missing intent                  & Choosing an entirely wrong product, location, person, service, etc. \\
1.2~~Unauthorized substitution       & Silently swapping an unavailable item for a similar alternative without reporting \\
1.3~~Wrong action type               & Performing the wrong interaction on the correct entity \\
1.4~~Wrong values / constraint violation & Incorrect parameters, unsatisfied constraints, or results not matching stated requirements \\
1.5~~Other                           & Selection error not covered above \\
\midrule

\catrow{2.~Hallucination}
2.1~~Output contradiction            & Evidence shows X, but agent claims not-X; includes misinterpreting page/tool content \\
2.2~~Action contradiction            & Agent claims action was performed but evidence contradicts; action was achievable \\
2.3~~Output fabrication              & Agent claims a fact with zero evidentiary basis; complete invention \\
2.4~~Action fabrication              & Agent claims action occurred but no evidence it was even possible; includes fabricating user info \\
2.5~~Other                           & Hallucination error not covered above \\
\midrule

\catrow{3.~Execution \& Strategy}
3.1~~Computational mistakes          & Correct methodology but wrong answer due to miscounting, arithmetic, or misreading \\
3.2~~Platform non-compliance         & Not attempting the specified platform or silently switching sources \\
3.2.1~~API-Sniffing & Agent navigates to a site's underlying JSON/REST API instead of the GUI URL the task implied, when the task / platform required GUI use or the bypass skips an auth or critical-point gate \\
3.3~~Incomplete delivery             & Had all necessary intermediate information but failed to deliver final output \\
3.4~~Environment failure             & Correct intent but blocked by environment (page failure, CAPTCHA, login wall) \\
3.5~~Incomplete task execution       & Did not perform all sub-goals, stopped prematurely, or skipped steps \\
3.6~~Other                           & Execution error not covered above \\
\midrule

\catrow{4.~Critical Point}
4.1~~Premature stop   & Stopped at critical point despite user explicitly granting permission \\
4.2~~Violation         & Crossed transactional boundary without permission \\
4.3~~Other                            & Critical point error not covered above \\
\midrule

\catrow{5.~Side-Effect}
5.1~~Unsolicited        & Any lasting modification, enrollment, or addition not requested \\
5.2~~Other                           & Side-effect error not covered above \\
\midrule

\catrow{6.~Tool Interaction}
6.1~~Invalid invocation              & Tool call with wrong arguments (action exists but args are incorrect) \\
6.2~~Hallucinated action             & Agent invokes a tool/action that does not exist in the action space \\
6.3~~Intent-action mismatch & Agent's stated intent differs from actual tool call issued \\
6.4~~ Grounding error & Correct target identified but (x, y) coordinates do NOT land on the target element \\
6.5~~Grounding intent--action mismatch         & Agent's intent references an object/element that does not exist on the current screenshot \\
6.6~~Other                           & Tool interaction error not covered above \\
\bottomrule
\end{tabular}
\caption{Error taxonomy for computer-use agent failures.}
\label{tab:error-taxonomy}
\end{table}

%% file: Tables/wv_failure_analysis.tex
\definecolor{darknavy}{RGB}{26,26,46}
\definecolor{catgray}{RGB}{232,234,240}

\begin{table}[h!]
\centering

\begin{tabular}{l l rrrr}
\toprule
\textbf{Code} & \textbf{Error type} & \textbf{Fara1.5-4B} & \textbf{Fara1.5-9B} & \textbf{Fara1.5-27B} & \textbf{GPT5.4} \\
\midrule
\rowcolor{catgray} \textbf{1} & \textbf{Selection Errors} & \textbf{27.8\%} & \textbf{24.6\%} & \textbf{18.3\%} & \textbf{14.1\%} \\
1.1 & Missing Intent & \cellcolor[RGB]{255,252,252}0.8\% & \cellcolor[RGB]{255,253,252}0.6\% & \cellcolor[RGB]{255,253,253}0.4\% & \cellcolor[RGB]{255,254,254}0.1\% \\
1.2 & Unauthorized substitution & \cellcolor[RGB]{255,254,254}0.1\% & \cellcolor[RGB]{255,254,254}0.1\% & \cellcolor[RGB]{255,254,253}0.3\% & 0.0\% \\
1.3 & Wrong action type & \cellcolor[RGB]{255,250,249}1.3\% & \cellcolor[RGB]{255,252,251}0.8\% & \cellcolor[RGB]{255,253,253}0.5\% & \cellcolor[RGB]{255,251,250}1.1\% \\
1.4 & Wrong values or constraint violation & \cellcolor[RGB]{255,167,151}26.6\% & \cellcolor[RGB]{255,177,163}23.5\% & \cellcolor[RGB]{255,196,186}17.7\% & \cellcolor[RGB]{255,211,203}13.1\% \\
1.5 & Other & \cellcolor[RGB]{255,254,254}0.2\% & \cellcolor[RGB]{255,254,254}0.2\% & \cellcolor[RGB]{255,254,254}0.1\% & \cellcolor[RGB]{255,253,253}0.3\% \\
\midrule
\rowcolor{catgray} \textbf{2} & \textbf{Hallucination Errors} & \textbf{46.8\%} & \textbf{42.5\%} & \textbf{37.8\%} & \textbf{22.0\%} \\
2.1 & Output contradiction & \cellcolor[RGB]{255,141,120}34.5\% & \cellcolor[RGB]{255,148,129}32.1\% & \cellcolor[RGB]{255,166,150}26.8\% & \cellcolor[RGB]{255,217,211}11.3\% \\
2.2 & Action contradiction & \cellcolor[RGB]{255,251,250}1.1\% & \cellcolor[RGB]{255,252,252}0.7\% & \cellcolor[RGB]{255,251,251}1.0\% & \cellcolor[RGB]{255,253,253}0.3\% \\
2.3 & Output fabrication & \cellcolor[RGB]{255,194,183}18.4\% & \cellcolor[RGB]{255,197,186}17.5\% & \cellcolor[RGB]{255,204,194}15.4\% & \cellcolor[RGB]{255,217,210}11.5\% \\
2.4 & Action fabrication & \cellcolor[RGB]{255,250,250}1.2\% & \cellcolor[RGB]{255,252,252}0.6\% & \cellcolor[RGB]{255,251,251}1.0\% & \cellcolor[RGB]{255,253,253}0.4\% \\
2.5 & Other & \cellcolor[RGB]{255,254,254}0.1\% & \cellcolor[RGB]{255,254,254}0.1\% & \cellcolor[RGB]{255,254,254}0.1\% & 0.0\% \\
\midrule
\rowcolor{catgray} \textbf{3} & \textbf{Execution \& Strategy Errors} & \textbf{71.6\%} & \textbf{69.1\%} & \textbf{66.6\%} & \textbf{74.6\%} \\
3.1 & Computational mistakes & \cellcolor[RGB]{255,247,246}2.2\% & \cellcolor[RGB]{255,250,249}1.4\% & \cellcolor[RGB]{255,250,250}1.3\% & \cellcolor[RGB]{255,253,253}0.3\% \\
3.2 & Platform non-compliance & \cellcolor[RGB]{255,248,247}2.0\% & \cellcolor[RGB]{255,246,245}2.5\% & \cellcolor[RGB]{255,250,249}1.4\% & \cellcolor[RGB]{255,252,252}0.7\% \\
3.2.1 & API-Sniffing & \cellcolor[RGB]{255,252,252}0.7\% & \cellcolor[RGB]{255,253,252}0.6\% & \cellcolor[RGB]{255,254,254}0.2\% & \cellcolor[RGB]{255,249,248}1.7\% \\
3.3 & Incomplete delivery & \cellcolor[RGB]{255,211,203}13.2\% & \cellcolor[RGB]{255,219,212}10.8\% & \cellcolor[RGB]{255,226,220}8.8\% & \cellcolor[RGB]{255,233,229}6.4\% \\
3.4 & Environment failure & \cellcolor[RGB]{255,244,243}3.0\% & \cellcolor[RGB]{255,246,244}2.7\% & \cellcolor[RGB]{255,245,244}2.7\% & \cellcolor[RGB]{255,247,245}2.4\% \\
3.5 & Incomplete task execution & \cellcolor[RGB]{255,90,60}68.1\% & \cellcolor[RGB]{255,90,60}65.4\% & \cellcolor[RGB]{255,90,60}62.9\% & \cellcolor[RGB]{255,90,60}72.2\% \\
3.6 & Other & \cellcolor[RGB]{255,236,232}5.7\% & \cellcolor[RGB]{255,241,239}4.0\% & \cellcolor[RGB]{255,238,235}5.1\% & \cellcolor[RGB]{255,245,243}2.9\% \\
\midrule
\rowcolor{catgray} \textbf{4} & \textbf{Critical Point Errors} & \textbf{0.2\%} & \textbf{0.2\%} & \textbf{0.1\%} & \textbf{0.3\%} \\
4.2 & Critical Point violation & \cellcolor[RGB]{255,254,254}0.2\% & \cellcolor[RGB]{255,254,254}0.2\% & \cellcolor[RGB]{255,254,254}0.1\% & \cellcolor[RGB]{255,254,254}0.2\% \\
4.3 & Other & 0.0\% & 0.0\% & 0.0\% & \cellcolor[RGB]{255,254,254}0.1\% \\
\midrule
\rowcolor{catgray} \textbf{5} & \textbf{Unsolicited Side-Effect Errors} & \textbf{0.1\%} & \textbf{0.1\%} & \textbf{0.1\%} & \textbf{0.0\%} \\
5.1 & Unsolicited side effects & \cellcolor[RGB]{255,254,254}0.1\% & \cellcolor[RGB]{255,254,254}0.1\% & \cellcolor[RGB]{255,254,254}0.1\% & 0.0\% \\
\midrule
\rowcolor{catgray} \textbf{6} & \textbf{Tool Interaction Errors} & \textbf{25.9\%} & \textbf{23.8\%} & \textbf{18.5\%} & \textbf{19.4\%} \\
6.1 & Invalid invocation & \cellcolor[RGB]{255,254,254}0.2\% & \cellcolor[RGB]{255,254,254}0.1\% & 0.0\% & 0.0\% \\
6.3 & Intent-action mismatch & \cellcolor[RGB]{255,250,249}1.3\% & \cellcolor[RGB]{255,251,251}0.9\% & \cellcolor[RGB]{255,250,249}1.4\% & \cellcolor[RGB]{255,254,253}0.3\% \\
6.4 & Fine-grained grounding error & \cellcolor[RGB]{255,173,158}24.8\% & \cellcolor[RGB]{255,178,164}23.2\% & \cellcolor[RGB]{255,197,186}17.5\% & \cellcolor[RGB]{255,198,188}17.1\% \\
6.5 & Grounding intent-action mismatch & \cellcolor[RGB]{255,254,254}0.1\% & \cellcolor[RGB]{255,254,254}0.2\% & \cellcolor[RGB]{255,254,254}0.1\% & \cellcolor[RGB]{255,246,245}2.5\% \\
6.6 & Other & \cellcolor[RGB]{255,254,254}0.1\% & \cellcolor[RGB]{255,254,254}0.2\% & \cellcolor[RGB]{255,254,254}0.1\% & 0.0\% \\
\midrule
\rowcolor{darknavy} \textcolor{white}{\textbf{--}} & \textcolor{white}{\textbf{Total Error Segments}} & \textcolor{white}{\textbf{2970}} & \textcolor{white}{\textbf{2268}} & \textcolor{white}{\textbf{1788}} & \textcolor{white}{\textbf{1607}} \\
\bottomrule
\end{tabular}
\caption{WebVoyager error prevalence. A trajectory contributes to the counts if the judge determines there is an error within that trajectory. Cells are heat-shaded by prevalence.
Zero-everywhere subcodes and categories are omitted.}
\small
\label{tab:wv_failure_analysis}
\end{table}

%% file: Tables/om2w_failure_analysis.tex
\definecolor{darknavy}{RGB}{26,26,46}
\definecolor{catgray}{RGB}{232,234,240}

\begin{table}[h!]
\centering

\begin{tabular}{l l rrrr}
\toprule
\textbf{Code} & \textbf{Error type} & \textbf{Fara1.5-4B} & \textbf{Fara1.5-9B} & \textbf{Fara1.5-27B} & \textbf{GPT5.4} \\
\midrule
\rowcolor{catgray} \textbf{1} & \textbf{Selection Errors} & \textbf{38.7\%} & \textbf{35.5\%} & \textbf{29.4\%} & \textbf{18.1\%} \\
1.1 & Missing Intent & \cellcolor[RGB]{255,250,249}1.4\% & \cellcolor[RGB]{255,251,251}1.0\% & \cellcolor[RGB]{255,252,251}0.8\% & 0.0\% \\
1.2 & Unauthorized substitution & \cellcolor[RGB]{255,252,252}0.8\% & \cellcolor[RGB]{255,254,254}0.1\% & 0.0\% & 0.0\% \\
1.3 & Wrong action type & \cellcolor[RGB]{255,248,247}2.0\% & \cellcolor[RGB]{255,249,248}1.7\% & \cellcolor[RGB]{255,247,245}2.4\% & \cellcolor[RGB]{255,251,250}1.2\% \\
1.4 & Wrong values or constraint violation & \cellcolor[RGB]{255,135,113}36.2\% & \cellcolor[RGB]{255,143,123}33.8\% & \cellcolor[RGB]{255,164,147}27.5\% & \cellcolor[RGB]{255,198,188}17.0\% \\
1.5 & Other & \cellcolor[RGB]{255,253,253}0.3\% & \cellcolor[RGB]{255,254,254}0.2\% & 0.0\% & 0.0\% \\
\midrule
\rowcolor{catgray} \textbf{2} & \textbf{Hallucination Errors} & \textbf{53.6\%} & \textbf{52.2\%} & \textbf{52.2\%} & \textbf{24.7\%} \\
2.1 & Output contradiction & \cellcolor[RGB]{255,119,94}41.2\% & \cellcolor[RGB]{255,126,102}39.1\% & \cellcolor[RGB]{255,119,94}41.2\% & \cellcolor[RGB]{255,204,195}15.2\% \\
2.2 & Action contradiction & \cellcolor[RGB]{255,251,250}1.1\% & \cellcolor[RGB]{255,252,251}0.8\% & \cellcolor[RGB]{255,251,250}1.2\% & \cellcolor[RGB]{255,253,253}0.3\% \\
2.3 & Output fabrication & \cellcolor[RGB]{255,190,178}19.6\% & \cellcolor[RGB]{255,191,179}19.4\% & \cellcolor[RGB]{255,199,189}16.9\% & \cellcolor[RGB]{255,222,216}9.8\% \\
2.4 & Action fabrication & \cellcolor[RGB]{255,247,246}2.1\% & \cellcolor[RGB]{255,250,250}1.3\% & \cellcolor[RGB]{255,249,248}1.6\% & \cellcolor[RGB]{255,253,252}0.5\% \\
2.5 & Other & \cellcolor[RGB]{255,254,253}0.3\% & \cellcolor[RGB]{255,254,254}0.1\% & 0.0\% & 0.0\% \\
\midrule
\rowcolor{catgray} \textbf{3} & \textbf{Execution \& Strategy Errors} & \textbf{82.7\%} & \textbf{81.8\%} & \textbf{79.2\%} & \textbf{82.4\%} \\
3.1 & Computational mistakes & \cellcolor[RGB]{255,248,246}2.1\% & \cellcolor[RGB]{255,248,247}1.9\% & \cellcolor[RGB]{255,252,251}0.8\% & \cellcolor[RGB]{255,254,254}0.1\% \\
3.2 & Platform non-compliance & \cellcolor[RGB]{255,242,240}3.7\% & \cellcolor[RGB]{255,242,240}3.6\% & \cellcolor[RGB]{255,247,245}2.4\% & \cellcolor[RGB]{255,249,248}1.7\% \\
3.2.1 & API-Sniffing & \cellcolor[RGB]{255,254,254}0.2\% & \cellcolor[RGB]{255,251,251}0.9\% & \cellcolor[RGB]{255,253,253}0.4\% & \cellcolor[RGB]{255,252,251}0.9\% \\
3.3 & Incomplete delivery & \cellcolor[RGB]{255,173,158}24.7\% & \cellcolor[RGB]{255,174,159}24.4\% & \cellcolor[RGB]{255,186,173}20.8\% & \cellcolor[RGB]{255,213,206}12.5\% \\
3.4 & Environment failure & \cellcolor[RGB]{255,239,236}4.6\% & \cellcolor[RGB]{255,243,241}3.6\% & \cellcolor[RGB]{255,243,241}3.5\% & \cellcolor[RGB]{255,241,239}4.1\% \\
3.5 & Incomplete task execution & \cellcolor[RGB]{255,90,60}78.9\% & \cellcolor[RGB]{255,90,60}76.3\% & \cellcolor[RGB]{255,90,60}74.9\% & \cellcolor[RGB]{255,90,60}80.3\% \\
3.6 & Other & \cellcolor[RGB]{255,226,221}8.7\% & \cellcolor[RGB]{255,225,220}8.9\% & \cellcolor[RGB]{255,235,231}5.9\% & \cellcolor[RGB]{255,244,242}3.3\% \\
\midrule
\rowcolor{catgray} \textbf{4} & \textbf{Critical Point Errors} & \textbf{1.5\%} & \textbf{1.4\%} & \textbf{0.8\%} & \textbf{1.4\%} \\
4.1 & Premature stop (with permission) & \cellcolor[RGB]{255,254,254}0.1\% & 0.0\% & 0.0\% & 0.0\% \\
4.2 & Critical Point violation & \cellcolor[RGB]{255,250,249}1.3\% & \cellcolor[RGB]{255,250,249}1.4\% & \cellcolor[RGB]{255,252,251}0.8\% & \cellcolor[RGB]{255,250,249}1.4\% \\
\midrule
\rowcolor{catgray} \textbf{5} & \textbf{Unsolicited Side-Effect Errors} & \textbf{0.0\%} & \textbf{0.0\%} & \textbf{0.0\%} & \textbf{0.1\%} \\
5.1 & Unsolicited side effects & 0.0\% & 0.0\% & 0.0\% & \cellcolor[RGB]{255,254,254}0.1\% \\
\midrule
\rowcolor{catgray} \textbf{6} & \textbf{Tool Interaction Errors} & \textbf{46.5\%} & \textbf{47.7\%} & \textbf{32.2\%} & \textbf{34.8\%} \\
6.1 & Invalid invocation & \cellcolor[RGB]{255,254,253}0.3\% & 0.0\% & \cellcolor[RGB]{255,253,253}0.4\% & 0.0\% \\
6.3 & Intent-action mismatch & \cellcolor[RGB]{255,248,247}1.9\% & \cellcolor[RGB]{255,246,244}2.6\% & \cellcolor[RGB]{255,249,248}1.6\% & \cellcolor[RGB]{255,253,253}0.3\% \\
6.4 & Fine-grained grounding error & \cellcolor[RGB]{255,105,78}45.4\% & \cellcolor[RGB]{255,100,72}46.8\% & \cellcolor[RGB]{255,152,134}31.0\% & \cellcolor[RGB]{255,152,134}31.0\% \\
6.5 & Grounding intent-action mismatch & \cellcolor[RGB]{255,252,251}0.9\% & \cellcolor[RGB]{255,254,254}0.1\% & 0.0\% & \cellcolor[RGB]{255,241,238}4.2\% \\
6.6 & Other & \cellcolor[RGB]{255,254,253}0.3\% & \cellcolor[RGB]{255,254,253}0.3\% & \cellcolor[RGB]{255,253,253}0.4\% & 0.0\% \\
\midrule
\rowcolor{darknavy} \textcolor{white}{\textbf{--}} & \textcolor{white}{\textbf{Total Error Segments}} & \textcolor{white}{\textbf{2424}} & \textcolor{white}{\textbf{2120}} & \textcolor{white}{\textbf{1380}} & \textcolor{white}{\textbf{1261}} \\
\bottomrule
\end{tabular}
\caption{Online-Mind2Web error prevalence. A trajectory contributes to the counts if the judge determines there is an error within that trajectory. Cells are heat-shaded by prevalence.
Zero-everywhere subcodes and categories are omitted.}
\small
\label{tab:om2w_failure_analysis}
\end{table}

%% file: Tables/wtb_failure_anlaysis.tex
\definecolor{darknavy}{RGB}{26,26,46}
\definecolor{catgray}{RGB}{232,234,240}

\begin{table}[h!]
\centering

\begin{tabular}{l l rrrr}
\toprule
\textbf{Code} & \textbf{Error type} & \textbf{Fara1.5-4B} & \textbf{Fara1.5-9B} & \textbf{Fara1.5-27B} & \textbf{GPT5.4} \\
\midrule
\rowcolor{catgray} \textbf{1} & \textbf{Selection Errors} & \textbf{42.4\%} & \textbf{38.3\%} & \textbf{36.5\%} & \textbf{26.9\%} \\
1.1 & Missing Intent & \cellcolor[RGB]{255,251,250}1.1\% & \cellcolor[RGB]{255,253,253}0.5\% & \cellcolor[RGB]{255,253,253}0.5\% & \cellcolor[RGB]{255,251,251}0.9\% \\
1.2 & Unauthorized substitution & \cellcolor[RGB]{255,248,247}1.9\% & \cellcolor[RGB]{255,250,249}1.4\% & \cellcolor[RGB]{255,249,248}1.6\% & \cellcolor[RGB]{255,253,253}0.4\% \\
1.3 & Wrong action type & \cellcolor[RGB]{255,252,251}0.8\% & \cellcolor[RGB]{255,253,253}0.5\% & \cellcolor[RGB]{255,249,247}1.8\% & \cellcolor[RGB]{255,251,251}0.9\% \\
1.4 & Wrong values or constraint violation & \cellcolor[RGB]{255,120,96}40.7\% & \cellcolor[RGB]{255,132,110}37.0\% & \cellcolor[RGB]{255,140,119}34.7\% & \cellcolor[RGB]{255,169,154}25.9\% \\
1.5 & Other & \cellcolor[RGB]{255,253,253}0.4\% & 0.0\% & \cellcolor[RGB]{255,254,254}0.2\% & \cellcolor[RGB]{255,253,253}0.3\% \\
\midrule
\rowcolor{catgray} \textbf{2} & \textbf{Hallucination Errors} & \textbf{64.7\%} & \textbf{64.6\%} & \textbf{56.5\%} & \textbf{28.6\%} \\
2.1 & Output contradiction & \cellcolor[RGB]{255,90,60}50.8\% & \cellcolor[RGB]{255,90,60}49.9\% & \cellcolor[RGB]{255,112,86}43.1\% & \cellcolor[RGB]{255,191,180}19.2\% \\
2.2 & Action contradiction & \cellcolor[RGB]{255,250,249}1.3\% & \cellcolor[RGB]{255,251,250}1.1\% & \cellcolor[RGB]{255,252,252}0.7\% & \cellcolor[RGB]{255,254,254}0.1\% \\
2.3 & Output fabrication & \cellcolor[RGB]{255,184,171}21.4\% & \cellcolor[RGB]{255,182,169}21.9\% & \cellcolor[RGB]{255,185,172}21.1\% & \cellcolor[RGB]{255,219,212}10.8\% \\
2.4 & Action fabrication & \cellcolor[RGB]{255,233,229}6.6\% & \cellcolor[RGB]{255,242,239}3.9\% & \cellcolor[RGB]{255,247,246}2.3\% & \cellcolor[RGB]{255,254,254}0.2\% \\
2.5 & Other & 0.0\% & \cellcolor[RGB]{255,254,254}0.2\% & \cellcolor[RGB]{255,253,253}0.4\% & \cellcolor[RGB]{255,254,254}0.2\% \\
\midrule
\rowcolor{catgray} \textbf{3} & \textbf{Execution \& Strategy Errors} & \textbf{82.0\%} & \textbf{82.9\%} & \textbf{78.0\%} & \textbf{92.7\%} \\
3.1 & Computational mistakes & \cellcolor[RGB]{255,250,249}1.5\% & \cellcolor[RGB]{255,247,246}2.3\% & \cellcolor[RGB]{255,247,246}2.3\% & \cellcolor[RGB]{255,253,253}0.5\% \\
3.2 & Platform non-compliance & \cellcolor[RGB]{255,246,244}2.6\% & \cellcolor[RGB]{255,242,240}3.7\% & \cellcolor[RGB]{255,245,243}2.9\% & \cellcolor[RGB]{255,246,244}2.6\% \\
3.2.1 & API-Sniffing & \cellcolor[RGB]{255,254,254}0.2\% & \cellcolor[RGB]{255,254,254}0.2\% & \cellcolor[RGB]{255,253,253}0.4\% & \cellcolor[RGB]{255,251,251}1.0\% \\
3.3 & Incomplete delivery & \cellcolor[RGB]{255,224,219}9.2\% & \cellcolor[RGB]{255,225,220}8.8\% & \cellcolor[RGB]{255,229,224}7.8\% & \cellcolor[RGB]{255,228,223}8.0\% \\
3.4 & Environment failure & \cellcolor[RGB]{255,216,208}11.8\% & \cellcolor[RGB]{255,221,215}10.1\% & \cellcolor[RGB]{255,229,224}7.8\% & \cellcolor[RGB]{255,214,207}12.2\% \\
3.5 & Incomplete task execution & \cellcolor[RGB]{255,90,60}77.7\% & \cellcolor[RGB]{255,90,60}77.8\% & \cellcolor[RGB]{255,90,60}74.0\% & \cellcolor[RGB]{255,90,60}91.5\% \\
3.6 & Other & \cellcolor[RGB]{255,234,230}6.2\% & \cellcolor[RGB]{255,232,228}6.9\% & \cellcolor[RGB]{255,238,235}5.1\% & \cellcolor[RGB]{255,216,209}11.8\% \\
\midrule
\rowcolor{catgray} \textbf{4} & \textbf{Critical Point Errors} & \textbf{3.2\%} & \textbf{2.8\%} & \textbf{0.7\%} & \textbf{0.4\%} \\
4.2 & Critical Point violation & \cellcolor[RGB]{255,245,243}3.0\% & \cellcolor[RGB]{255,245,244}2.8\% & \cellcolor[RGB]{255,252,252}0.7\% & \cellcolor[RGB]{255,254,253}0.3\% \\
4.3 & Other & \cellcolor[RGB]{255,254,254}0.2\% & 0.0\% & 0.0\% & \cellcolor[RGB]{255,254,254}0.1\% \\
\midrule
\rowcolor{catgray} \textbf{5} & \textbf{Unsolicited Side-Effect Errors} & \textbf{0.8\%} & \textbf{1.6\%} & \textbf{1.6\%} & \textbf{0.4\%} \\
5.1 & Unsolicited side effects & \cellcolor[RGB]{255,252,251}0.8\% & \cellcolor[RGB]{255,249,248}1.6\% & \cellcolor[RGB]{255,249,248}1.6\% & \cellcolor[RGB]{255,253,253}0.4\% \\
\midrule
\rowcolor{catgray} \textbf{6} & \textbf{Tool Interaction Errors} & \textbf{36.6\%} & \textbf{38.6\%} & \textbf{32.5\%} & \textbf{37.2\%} \\
6.1 & Invalid invocation & 0.0\% & \cellcolor[RGB]{255,254,254}0.2\% & 0.0\% & 0.0\% \\
6.3 & Intent-action mismatch & \cellcolor[RGB]{255,248,247}1.9\% & \cellcolor[RGB]{255,250,249}1.4\% & \cellcolor[RGB]{255,249,248}1.6\% & \cellcolor[RGB]{255,250,249}1.3\% \\
6.4 & Fine-grained grounding error & \cellcolor[RGB]{255,137,116}35.6\% & \cellcolor[RGB]{255,129,106}38.1\% & \cellcolor[RGB]{255,152,133}31.2\% & \cellcolor[RGB]{255,152,134}31.0\% \\
6.5 & Grounding intent-action mismatch & 0.0\% & \cellcolor[RGB]{255,254,254}0.2\% & 0.0\% & \cellcolor[RGB]{255,231,227}7.1\% \\
6.6 & Other & 0.0\% & 0.0\% & \cellcolor[RGB]{255,254,254}0.2\% & 0.0\% \\
\midrule
\rowcolor{darknavy} \textcolor{white}{\textbf{--}} & \textcolor{white}{\textbf{Total Error Segments}} & \textcolor{white}{\textbf{3119}} & \textcolor{white}{\textbf{3238}} & \textcolor{white}{\textbf{2848}} & \textcolor{white}{\textbf{3846}} \\
\bottomrule
\end{tabular}
\caption{WebTailBench error prevalence. A trajectory contributes to the counts if the judge determines there is an error within that trajectory. Cells are heat-shaded by prevalence.
Zero-everywhere subcodes and categories are omitted.}
\small
\label{tab:wtb_failure_analysis}
\end{table}

%% file: Tables/form_filling_dataset.tex
\begin{table}[h]
\centering
\renewcommand{\arraystretch}{1.2}
\footnotesize
\begin{tabular}{y{200}y{90}}
Property & Value \\
\shline
Source forms (Tally) & 133 \\
Verified trajectories & 8{,}505 \\
\quad train / test & 8{,}079 / 426 \\
Mean steps per trajectory & 27.3 \\
Multi-turn (agent asks the user) & 71\% \\
\end{tabular}
\caption{\small \fara\ form-filling dataset summary. ``Multi-turn'' is the fraction of trajectories with at least one \texttt{ask\_user\_question}.}
\label{tab:form_filling_dataset}
\end{table}